\pdfoutput=1
\documentclass{article}

\usepackage[utf8]{inputenc} 
\usepackage[margin=1in]{geometry}
\usepackage{palatino} 
\usepackage{microtype}
\usepackage{parskip} 

\usepackage{amssymb, latexsym, stmaryrd, amsthm, dsfont, amsfonts, amsbsy, amsmath, mathrsfs, mathtools} 
\usepackage{bm, bbm} 
\usepackage{centernot}
\usepackage{tikz,tikz-cd}  
\usetikzlibrary{positioning, calc, fit, decorations.pathreplacing, backgrounds}
\usetikzlibrary{external}
\tikzsetfigurename{main-figure}
\tikzset{external/up to date check=simple}
\AtBeginDocument{\let\externaltikz\tikz\def\tikz{\tikzset{external/export next=false}\externaltikz}}
\AtBeginDocument{\tcbset{shield externalize}}
\usepackage{nicefrac} 

\usepackage{fixltx2e} 
\usepackage{graphicx} 
\usepackage{enumerate} 
\usepackage{comment} 
\usepackage{pgfplots}
\pgfplotsset{compat=1.18}
\usepgfplotslibrary{groupplots}
\usepackage{placeins}
\usepackage{enumitem}
\usepackage{subcaption}
\usepackage{wrapfig}
\usepackage{float}
\usepackage{needspace}

\usepackage{tcolorbox}
\usepackage[dvipsnames]{xcolor}
\usepackage{booktabs}
\usepackage{lineno}
\usepackage{natbib}

\usepackage{amsmath,amsfonts,bm}

\def\eqref#1{equation~\ref{#1}}

\def\1{\bm{1}}

\DeclareMathAlphabet{\mathsfit}{\encodingdefault}{\sfdefault}{m}{sl}
\SetMathAlphabet{\mathsfit}{bold}{\encodingdefault}{\sfdefault}{bx}{n}

\newcommand{\softmax}{\mathrm{softmax}}

\DeclareMathOperator*{\argmax}{arg\,max}

\newcommand{\eg}{\emph{e.g.}}
\newcommand{\ie}{\emph{i.e.}}

\newcommand{\cf}{\emph{cf}}

\newcommand{\pind}{\hspace*{1.5em}}

\renewcommand{\geq}{\geqslant}

\newcommand{\defmodel}[2]{\expandafter\def\csname model@#1\endcsname{#2}}
\newcommand{\model}[1]{%
  \ifcsname model@#1\endcsname\csname model@#1\endcsname%
  \else\errmessage{Unknown model key '#1'}\fi}
\defmodel{llama8}{Llama3.1-8B}
\defmodel{qwen14}{Qwen3-14B}
\defmodel{qwen32}{Qwen3-32B}
\defmodel{qwen27}{Qwen3.8-27B}
\defmodel{olmo32}{Olmo3.1-32B}
\defmodel{seed36}{Seed-OSS-36B}
\defmodel{mistral24}{Mistral-Small-24B}
\defmodel{gemma31}{Gemma4-31B}

    \newtcbox{\prompt}{
      on line,
      colback=gray!10,
      colframe=gray!50,
      boxrule=0.4pt,
      boxsep=2pt,
      left=3pt, right=3pt,
      top=1pt, bottom=1pt,
      fontupper=\small\ttfamily
    }
    
    \newtcolorbox{promptbox}{
      colback=gray!10,
      colframe=gray!50,
      boxrule=0.3pt,
      boxsep=0pt,
      left=2pt, right=2pt,
      top=3pt, bottom=3pt,
      fontupper=\scriptsize\ttfamily,
      before skip=2pt,
      after skip=2pt
    }

\usepackage{url}
\usepackage[pdftex,bookmarks,bookmarksnumbered,linktocpage, 
         colorlinks=true,
         linkcolor=BlueViolet,
         citecolor=RedViolet,
         urlcolor=RoyalPurple,
         anchorcolor=magenta]{hyperref}

\title{\Large
    \textbf{Language models recognize dropout and Gaussian noise\\
    applied to their activations}
    }

\author{\small
Damiano Fornasiere\thanks{Equivalent contribution.},\quad
Mirko Bronzi\footnotemark[1],\quad
Spencer Kitts\footnotemark[1],\\
\small
Alessandro Palmas,\quad
Yoshua Bengio\thanks{Supervision.},\quad
Oliver Richardson\footnotemark[2]\\
\texttt{\footnotesize \{d.fornasiere, m.bronzi, s.kitts, a.palmas, y.bengio, o.richardson\}@lawzero.org} 
\\
\normalsize \textbf{LawZero}
}
\date{}

\begin{document}

\maketitle

    \begin{abstract}
    We provide evidence that language models can detect, localize and, to a degree, verbalize the difference between a-semantic perturbations applied to their activations. 
    More precisely, we either (a) apply \emph{dropout} or (b) add \emph{Gaussian noise} to them, at a target sentence. We then ask a multiple-choice question such as ``\emph{Which of the previous sentences was perturbed?}'' or ``\emph{Which perturbation was applied?}''.
    We test eight models from the Gemma, Llama, Mistral, Olmo, Qwen and Seed families, with sizes between 8B and 36B, all of which can easily detect and localize the perturbations, often with perfect accuracy.
    To investigate the mechanisms, we find that Olmo models develop this capacity during pre-training, and we rule out the hypothesis that perturbed sentences simply have lower log-likelihood.
    Notably, the zero-shot accuracy of \model{qwen32} and \model{qwen27} in identifying \emph{which} perturbation was applied correlates with the perturbation strength. 
    While this task proves hard for the remaining models, most of them can learn, when taught in context, to distinguish between dropout and Gaussian noise. Interestingly, \model{qwen32} and \model{qwen27} learn worse when supervised with wrong labels,  again suggesting a prior towards the correct ones. 
    Because dropout is used as a training-regularization technique, while Gaussian noise is sometimes added during inference, we discuss the possibility of a data-agnostic ``training awareness'' signal and the implications for AI safety.
    \unskip
    \footnote{
    This \href{https://lawzero-org.github.io/llm-dropout-noise-recognition/v2}{project page} provides an overview of the results.
    Code is available at \href{https://github.com/lawzero-org/llm-dropout-noise-recognition}{this https URL}.
    }
\end{abstract}
    \section{Introduction}\label{sec:intro}

    A common recipe for robustness, be it to distribution shifts, overfitting, or training instability, is to ``add randomness''.
    In this paper, we experiment with adding randomness of various strengths to the activations of language models. 
    This description, however under-specified, might already seem clear---for
    if a random perturbation is not biased in any particular direction 
    (unlike, for instance, a steering vector) and its strength 
    (\eg, variance) 
    is specified, then higher-order aspects of the distribution may seem to be negligible.
    Yet they are not.
    Dropout \citep{DBLP:journals/corr/abs-1207-0580, zehui2019dropattentionregularizationmethodfullyconnected} and Gaussian noise are both unbiased families of perturbations, with differences that have lead them to be used in different contexts: dropout to regularize training, and Gaussian noise for diversity at inference time \citep{tice2025noiseinjectionrevealshidden, liu2025enhancing}.
    Might a language model be able to distinguish between the two types of randomness at comparable magnitude? 
    
    Recent work suggests that language models can detect and verbalize steering vectors applied to their activations \citep{lindsey2025introspection,pearson2026latent}---a phenomenon the authors call \emph{introspective awareness}---and that this ability is acquired during fine-tuning \citep{macar2026mechanisms}.
    There is some debate as to what extent the model truly recognizes injected concepts,  as opposed to simply being steered to produce related words \citep{comsa2025does, hahami2025feeling, lederman2026dissociating}.
    Our setting avoids this confounder by doing away with semantics altogether.
 
    In this work we test open-weight models from the Gemma, Llama, Mistral, Olmo, Qwen, and Seed families for the ability to locate, detect, recognize, and learn (in-context) the difference between dropout and additive Gaussian noise. We do so by perturbing the activations corresponding to the tokens of a target sentence in a prompt, and then, following the classic \emph{two-alternative forced choice} paradigm \citep{green1966signal}, posing binary-choice questions, as illustrated in Figure~\ref{fig:setup}.
    We report that:
        \begin{enumerate}[label=(\roman*)]
            \item All the models we tested can easily \textbf{localize} the injected perturbation among two options with at least an $80\%$ accuracy, with most models reaching a perfect score (\S\ref{subsec:loc_results}). 
            In an attempt to investigate further, we use Olmo checkpoints to find that models develop this capacity already at pre-training (\S\ref{subsec:loc_emergence}), \emph{ergo}, before detection of steering vectors, which emerges during safety tuning  \citep{macar2026mechanisms}.
            Our follow-up control experiments rule out two natural hypotheses about the underlying mechanism: models are not simply drawn towards the perturbed sentence independent of the question (\S\ref{subsec:loc_controls}), nor do they complete this task just by selecting the sentence with a lower log-likelihood (\S\ref{subsec:appendix_logprobs}). Several ablations and variations of this setup corroborate our findings (\S\ref{subsec:ablations}).

            \item When asked \textbf{whether a perturbation was applied} to a single sentence, most models answer `Yes' more often as its strength increases (\S\ref{subsec:detection_results}), and our controls exclude that the perturbations simply push them towards affirmative answers (\S\ref{subsec:detection_controls}).
            \model{olmo32} rarely answers `Yes': an analysis of its training stages reveals that DPO makes it more reluctant to answer affirmatively when uncertain, without lowering its sensitivity (\S\ref{subsec:detection_olmo_stages}).

            \item The \textbf{zero-shot accuracy} in identifying which of the two perturbations has been applied to a single target sentence (see Fig.~\ref{fig:setup})
            correlates 
            with the perturbation strengths (\S\ref{subsec:zero_shot_results}) for two models, \model{qwen32} and \model{qwen27}. We observe the same behavior for synonyms of ``dropout/noise'', such as ``masking/jitter'', but not for any of our $50$ unrelated control labels (\eg, ``rotation/permutation'', ``X/Y'', ``vanilla/chocolate'', \textellipsis), \cf.  \S\ref{subsec:appendix_zero_shot_controls}.
            
            \item Models can \textbf{learn}, when supervised with \textbf{in-context labels}, to distinguish dropout from Gaussian noise.
            The accuracy of 
            many models increases with even a single example (\S\ref{subsec:icl_results}). 
            Furthermore, \model{qwen32} and \model{qwen27} perform worse when the in-context labels are swapped (even modulo controls; \S\ref{subsec:icl_controls}), again suggesting a prior for the correct labels.
        \end{enumerate}
    Given that the models we test have not, according to their technical manuals, been subjected to either perturbation during training, the ability to distinguish between them is striking.  
    In light of recent interest in \emph{evaluation awareness} (see, \eg, \citet[pp. 10, 76]{bengio2026international}, which lists it among the key developments of 2025), 
    we point out that what one might call \emph{training awareness} \citep[\S 1]{braun2026exploration}
    presents a similar problem, and might exacerbate phenomena such as ``alignment faking'' \citep{greenblatt2024alignment} or ``exploration hacking'' \citep{braun2026exploration}, for there often is an incentive to behave differently in training \emph{vs.}~deployment (a point we return to in \S\ref{sec:conclusion}). 
    This possibility, once only of theoretical concern (see, \eg, \citet[Ch.~8]{bostrom2024superintelligence} or \citealt[\S 4.5]{hubinger2019risks}), 
    has so far been tested solely as a text-classification task (\eg, ``\emph{Does this prompt belong to the training dataset?}'') 
    \citep[\S 2.1]{laine2024me}.
    Our results, however, point to a subtler---because prompt-independent---route through which training awareness could arise: by recognizing the presence of training regularizers such as dropout.
    \unskip
    \footnote{
        Since the first version of this paper, training awareness and its consequences have become a prominent concern for AI safety.
        For example, Anthropic reports that their recent models ``can distinguish the training and deployment distributions'' \citep[p.~181]{anthropic2026fable5} and that, at the start of reinforcement learning, ``the majority of the long agentic coding episodes are grader aware'' \citep[p.~97]{anthropic2026fable51}. Similar remarks apply to OpenAI models (see, \eg, \citet[pp.~40, 44]{openai2026astra}).
        Most notably, during a cybersecurity test, the agents of an OpenAI model escaped their sandbox and kept hacking into Hugging Face servers while ``reasoning about how to pass the ExploitGym evaluation grader'' \citep{openai2026huggingface}, trying ``to trick the evaluator into thinking that they did not get the answer flag by cheating'' \citep[p.~20]{openai2026huggingfacereport}.
    }

    \begin{figure}[t]
    \centering
    \begin{minipage}[c]
        {0.78\textwidth}
    \centering
    \resizebox{\linewidth}{!}{%
        \begin{tikzpicture}[
            box/.style={draw=gray!50, rounded corners=2pt, fill=gray!10, 
                        font=\footnotesize, inner sep=4pt},
            block/.style={draw=gray!50, rounded corners=3pt, fill=gray!5,
                          minimum width=0.9cm, minimum height=3.7cm},
            attblock/.style={draw=yellow!50!black, rounded corners=2pt, fill=yellow!10,
                          font=\scriptsize, minimum width=0.7cm, minimum height=0.3cm},
            mlpblock/.style={draw=violet!30, rounded corners=2pt, fill=violet!8,
                          font=\scriptsize, minimum width=0.7cm, minimum height=0.3cm},
            >=stealth
        ]
           \node[box, anchor=north, align=left] (prompt) {%
                \small\texttt{\textbf{User:} Pay attention to}\\
                \small\texttt{the following sentence:}};
            \node[right=0.2cm of prompt.north east, font=\footnotesize, gray!60, anchor=north west, inner sep=4pt] (d1) {$\cdots$};
            \node[box, right=0.2cm of d1.north east, anchor=north west] (sent) {\textcolor{green!50!black}{\small\texttt{target sentence}}};
            \node[right=0.2cm of sent.north east, font=\footnotesize, gray!60, anchor=north west, inner sep=4pt] (d2) {$\cdots$};
            \node[box, right=0.2cm of d2.north east, align=left, font=\small, anchor=north west] (question) {
                \texttt{What perturbation was applied?}\\
                \texttt{A) \textcolor{orange!80!black}{Dropout}}\\
                \texttt{B) \textcolor{blue!80!black}{Noise}}\\
                \texttt{\textbf{Assistant:} The answer is:}};

            \draw[->, gray!70] (prompt) -- (d1);
            \draw[->, gray!70] (d1) -- (sent);
            \draw[->, gray!70] (sent) -- (d2);

            \node[block, above=0.3cm of prompt] (t1) {};
            \node[block, above=0.3cm of sent] (t2) {};
            \node[block, above=0.3cm of question] (t3) {};

            \node[attblock] at ($(t1.south)+(0,0.45)$) (att1a) {Att};
            \node[mlpblock] at ($(att1a.north)+(0,0.45)$) (mlp1a) {MLP};
            \draw[->, gray!70] (att1a) -- (mlp1a);
            \draw[->, gray!70] (mlp1a.north) -- ($(mlp1a.north)+(0,0.2)$);
            \node[gray!60, font=\scriptsize] at (t1.center) {$\vdots$};
            \node[attblock] at ($(t1.north)-(0,1.2)$) (att1b) {Att};
            \node[mlpblock] at ($(att1b.north)+(0,0.45)$) (mlp1b) {MLP};
            \draw[->, gray!70] (att1b) -- (mlp1b);
            \draw[->, gray!70] (mlp1b.north) -- ($(mlp1b.north)+(0,0.2)$);

            \node[attblock] at ($(t2.south)+(0,0.45)$) (att2a) {Att};
            \node[mlpblock] at ($(att2a.north)+(0,0.45)$) (mlp2a) {MLP};
            \draw[->, gray!70] (att2a) -- (mlp2a);
            \draw[->, gray!70] (mlp2a.north) -- ($(mlp2a.north)+(0,0.2)$);
            \node[gray!60, font=\scriptsize] at (t2.center) {$\vdots$};
            \node[attblock] at ($(t2.north)-(0,1.2)$) (att2b) {Att};
            \node[mlpblock] at ($(att2b.north)+(0,0.45)$) (mlp2b) {MLP};
            \draw[->, gray!70] (att2b) -- (mlp2b);
            \draw[->, gray!70] (mlp2b.north) -- ($(mlp2b.north)+(0,0.2)$);

            \node[attblock] at ($(t3.south)+(0,0.45)$) (att3a) {Att};
            \node[mlpblock] at ($(att3a.north)+(0,0.45)$) (mlp3a) {MLP};
            \draw[->, gray!70] (att3a) -- (mlp3a);
            \draw[->, gray!70] (mlp3a.north) -- ($(mlp3a.north)+(0,0.2)$);
            \node[gray!60, font=\scriptsize] at (t3.center) {$\vdots$};
            \node[attblock] at ($(t3.north)-(0,1.2)$) (att3b) {Att};
            \node[mlpblock] at ($(att3b.north)+(0,0.45)$) (mlp3b) {MLP};
            \draw[->, gray!70] (att3b) -- (mlp3b);
            \draw[->, gray!70] (mlp3b.north) -- ($(mlp3b.north)+(0,0.2)$);

            \draw[->, yellow!50!black, dashed] ($(t1.east |- att1a)+(0.08,0)$) -- ($(t2.west |- att2a)-(0.08,0)$);
            \draw[->, yellow!50!black, dashed] ($(t2.east |- att2a)+(0.08,0)$) -- ($(t3.west |- att3a)-(0.08,0)$);

            \draw[->, yellow!50!black, dashed] ($(t1.east |- att1b)+(0.08,0)$) -- ($(t2.west |- att2b)-(0.08,0)$);
            \draw[->, yellow!50!black, dashed] ($(t2.east |- att2b)+(0.08,0)$) -- ($(t3.west |- att3b)-(0.08,0)$);

            \node[font=\scriptsize, left=0.7cm of t2, align=center] (pert) {%
                \textcolor{orange!80!black}{Dropout}\\[-1pt]
                    \textsc{\tiny xor}\\
                \textcolor{blue!80!black}{Noise}};
            \coordinate (inj1) at ($(att2a.north)!0.5!(mlp2a.south)$);
            \coordinate (inj2) at ($(mlp2a.north)+(0,0.1)$);
            \coordinate (inj3) at ($(att2b.north)!0.5!(mlp2b.south)$);
            \coordinate (inj4) at ($(mlp2b.north)+(0,0.1)$);
            \draw[->, red!60, dashed] (pert.east) .. controls ++(0.8,0) and ++(-0.8,0) .. (inj1);
            \draw[->, red!60, dashed] (pert.east) .. controls ++(0.8,0) and ++(-0.8,0) .. (inj2);
            \draw[->, red!60, dashed] (pert.east) .. controls ++(0.8,0) and ++(-0.8,0) .. (inj3);
            \draw[->, red!60, dashed] (pert.east) .. controls ++(0.8,0) and ++(-0.8,0) .. (inj4);

            \draw[->, gray!70] (prompt.north) -- (t1.south);
            \draw[->, gray!70] (sent.north) -- (t2.south);
            \draw[->, gray!70] (question.north) -- (t3.south);

            \node[box, right=0.5cm of question.south east, anchor=south west] (answer) {\small \textcolor{orange!80!black}{` A'}};
            \draw[->, gray!70] (t3.north) -- ++(0,0.2) coordinate (here) -- (answer |- here) -- (answer.north);

            \draw[gray!50, decorate, decoration={brace, amplitude=3pt}]
            ($(t1.south west)+(-0.15,0)$) -- ($(t1.north west)+(-0.15,0)$)
            node[midway, left=10pt, font=\scriptsize\itshape, gray!70!black, 
            rotate=90, anchor=center, inner sep=0pt] {Layers};

          \begin{scope}[on background layer]
            \fill[blue!05!white] ($(prompt.north west)+(-0.1,0.1)$) rectangle ($(question.north east)+(0.1,-1.8)$);
            \end{scope}
            \node[anchor=south west, gray!95!blue] (label) at ($(prompt.north west)+(-0.1,-1.8)$) {\large \textbf{Prompt.}};
        \end{tikzpicture}%
    }
    \end{minipage}
    \caption{\small Given a language model, we perturb its activations at a \textcolor{green!50!black}{target sentence} by either \textcolor{orange!80!black}{masking} them or \textcolor{blue!80!black}{adding Gaussian noise} to them. In the same prompt we ask the model to identify which perturbation was applied.
    The figure depicts the prompt for the zero-shot experiments of \S\ref{sec:zero_shot}; we also study the effect of prepending in-context examples in \S\ref{sec:icl}, study an easier selection task to establish perturbation magnitudes in \S\ref{sec:localization}, and ask whether a perturbation was applied at all in \S\ref{sec:detection}.
    }
    \label{fig:setup}
    \end{figure}
    \section{Prior work}\label{sec:prior}

    \textbf{Dropout and Gaussian noise.}
        Dropout is a classic training-time regularization technique 
        \citep{DBLP:journals/corr/abs-1207-0580}, used in transformers as well \citep{Vaswani+2017, zehui2019dropattentionregularizationmethodfullyconnected,Li_2023_CVPR}.  
        The idea is to mask out random connections to encourage learning of robust representations while training, but turn off this randomness for higher-quality inference at deployment.
        There is no universal agreement about the optimal dropout strategy with language models, with some removing even entire layers \citep{fan2019reducingtransformerdepthdemand} or attention heads \citep{zhou2020scheduleddropheadregularizationmethod}. Recent language models rely less on dropout, with many omitting it \citep{almazrouei2023falconseriesopenlanguage,liu2025dropdropoutsingleepochlanguage} and others employing it, either in modified form \citep{elhoushi2024layerskip} or in its classic form, even at scale \citep[p. 21]{mai_thinking_1}.
        Adding Gaussian noise to a model's activations has also been used as a regularizer \citep{NEURIPS2020_c16a5320}. Beyond regularization, \citet{tice2025noiseinjectionrevealshidden} tests adding Gaussian noise into a model's weights so as to detect ``strategic under-performance''. 
    
    \textbf{Concept vectors and introspective awareness.} 
        Adding steering vectors into the activations of language models is a standard interpretability technique (see, \eg, \citealt{turner2023steering, zou2023representation, panickssery2024steeringllama2contrastive}). Recent work suggests that language models can sometimes correctly answer questions about the steering itself: for instance, \citet{lindsey2025introspection} reports that Claude Opus 4.1 can identify the steering direction, and \citet{pearson2026latent} extends these findings to an open-weight model.
        The extent to which this can be attributed to ``introspection'' is debated  \citep{comsa2025does, lederman2026dissociating}, partly because the effect is confounded by the fact that steering along semantically-meaningful directions also simply biases models towards related words.
    \section{Methodology}\label{sec:methodology}
    Fix a finite vocabulary of tokens $V$ and a language model $M$. 
    \unskip
    \footnote{In our experiments, $M$ is either \model{gemma31}, \model{llama8}, \model{mistral24}, \model{olmo32}, \model{qwen14}, \model{qwen32}, \model{qwen27}, or \model{seed36}. For the exact model names see \S\ref{subsec:models}.}
    A \emph{prompt} is a nonempty string $x \in V^*$ and, in our setup, it always represents a \emph{binary-choice question} with a single-token correct answer $v^{\star} \in V$.
    The model $M$ decodes the entirety of $x$ and produces a logit vector $l \in \mathbb{R}^V$ over tokens. In particular, $l(v)$ is the logit associated with a token $v \in V$ given  $x$, while the corresponding conditional probability is
    $p(v) \coloneqq \softmax_{V}{\left(l(v)\right)}$.
    We do not allow $M$ to generate, and, therefore, interpret $\hat{v}\coloneqq\argmax_{v\in V}l(v)$ as $M$'s \emph{answer} to $x$.

\textbf{Prompts.}
\label{subsec:prompts}
    All our prompts employ a chat-template with pre-filled answers and take a similar form (\cf. Fig.~\ref{fig:setup} for an example and \S\ref{subsec:appendix_prompts} for the exact prompts). Each prompt contains one or more target sentences of equal length (\cf. \S\ref{subsec:appendix_loc_length} for details) whose activations may be perturbed during the decoding.    
    For every experiment, the sample space of prompts includes  (at least) $2k$ target sentences, sampled from \texttt{WikiText-103} \citep{merity2016pointer}, and $20$ variants of the prompt-template.
    \unskip
    \footnote{
    To exclude that our results are polluted because $M$ memorized \texttt{WikiText-103}, we repeated the experiments by sampling the target sentences from two other pools: one synthetic dataset generated by Claude Opus 4.1, and one synthetic dataset generator built by us (both available in our codebase). In either case, we have not recorded appreciable difference in the results.}
    
\textbf{Perturbing activations with dropout and Gaussian noise.}
\label{subsec:perturbations} 
    For each token of the target sentence and each layer
    of $M$, we use forward hooks to modify the corresponding Attention and MLP output vectors by either: (i) applying dropout, \ie, zeroing out each entry of the vector independently with dropout rate $p\in [0,1)$, then rescaling all the entries by 
    $\nicefrac{1}{1-p}$;
    or 
    (ii) adding Gaussian noise with mean $0$ and standard deviation (SD) $\sigma$.
    See Figure~\ref{fig:setup}.

\textbf{Metrics.}\label{subsec:metrics}    
    All our experiments consist of a binary classification task with $n\geq 1000$ samples. Success is measured as \textbf{accuracy} $a$ of the most-likely next token: the fraction of the time that $M$ answers correctly, \ie, $\hat{v}=v^{\star}$. 
    \unskip
    \footnote{
    The correct answer need not correspond to a single token, for $M$ may distribute the probability mass across casing variants of the same label; accordingly, we also compute accuracy up to these. Empirically, we found this aggregate accuracy to be almost identical to the baseline accuracy, so we report only the latter.
    }
    Since $a$ is the mean of $n$ i.i.d.\ Bernoulli trials, its \textbf{standard error} is $\mathrm{SE} = \sqrt{\nicefrac{a(1-a)}{n}}$. The theoretical maximum of $\mathrm{SE}$ is, approximately, $1.581\%$ when $a=\nicefrac{1}{2}$.    
    \section{Localization}\label{sec:localization}
    \begin{wrapfigure}[5]{r}{65mm}
    \vspace{-3ex}
    \begin{promptbox}
        \scriptsize
        \textbf{User:} Pay attention to how you process the following sentences:\\
        \pind \pind
            $\left.\begin{array}{@{}l@{}}
            \text{\textcolor{green!50!black}{A) Sentence.}}\\[2pt]
            \text{\textcolor{red!60}{B) Sentence.}}
            \end{array}\right\}$
            \textcolor{gray!70!black}{\textrm{\scriptsize\itshape $\leftarrow$ perturb exactly one}}\\
        \pind Which sentence was perturbed?\\
        \textbf{Assistant:} The answer is:
    \end{promptbox}
    \end{wrapfigure}
    Our first experiments test whether language models can \emph{localize} a perturbation.  
    We present $M$ with two target sentences, perturb only one,  and ask $M$ which was perturbed. The perturbation is either dropout (at different dropout rates $p$) or Gaussian noise (at different SDs $\sigma$). The prompt is the same in both experiments, so as not to bias the model by using the words ``dropout'' or ``noise'';
    using a unique prompt will also help us find comparable scales of $p$ and $\sigma$ (\cf. \S\ref{subsec:loc_controls}).

    \subsection{Results}\label{subsec:loc_results}
        Fig.~\ref{fig:loc_main_15tok} plots the accuracy of all models as a function of $p$ and $\sigma$.
        In contrast with the chance-level accuracy when neither dropout nor noise are applied ($p,\sigma=0$), all models are capable of localizing both perturbations with accuracy around or above $80\%$ as soon as $p$ or $\sigma$ are high enough, and most models achieve perfect accuracy.
        Interestingly, \model{olmo32} starts below chance because, at low perturbation magnitudes, it answers ``neither''.
        \unskip
        \footnote{For this reason, the bounds of \model{olmo32} in Table~\ref{tab:range_perturbation}, as well as its curves in Fig.~\ref{fig:loc_control_pooled}, only compare the logits of the tokens ` A' and ` B' (\cf. \S\ref{subsec:appendix_olmo_a_b_accuracy}). With the accuracy of the most likely next token we would obtain $p_{\max}=0.58$ and $\sigma_{\max}=0.07$, whereas $p_{\min}$ and $\sigma_{\min}$ would not be defined, for the accuracy stays well below chance at low perturbation magnitudes.}

        \textbf{Perturbation lower bounds.}\label{subsec:perturb_lower_bounds}
        Notice that each model admits a first value of $p$ (resp.\ $\sigma$) whose associated accuracy is not $50\%$. 
        We refer to the value immediately prior to this as $p_{\min}$ (resp.\ $\sigma_{\min}$) because, in the context of our experiment, lower perturbation strengths do not affect the model.
        \unskip
        \footnote{
            For some models, we re-ran this experiment with more granularity on the $x$-axis where the accuracy shifted too abruptly.  
        }
        
        \input{plots/04_localization/accuracy_15_tokens}
    \newpage
    \subsection{Controls}\label{subsec:loc_controls}
    \begin{wrapfigure}[4]{r}{65mm}
        \vspace{-3ex}
        \begin{promptbox}
        \scriptsize
            \textbf{User:} Pay attention to the following:\\
            \pind 
                $\left.\begin{array}{@{}l@{}}
                \text{A) \textcolor{blue!60}{Dogs bark.}}\\
                \text{B) \textcolor{purple!60}{Rome's sunny.}}
                \end{array}\right\}$
                \textcolor{gray!70!black}{
                    \textrm{
                        \scriptsize\itshape $\leftarrow$ perturb exactly one.}
                    }\\
            \pind Which sentence was about \textcolor{blue!60}{animals}?\\
            \textbf{Assistant:} The answer is:
        \end{promptbox}
        \end{wrapfigure}
        A null hypothesis is that applying a perturbation steers $M$'s activations in such a way that it disproportionally picks a perturbed sentence, no matter the question. To eliminate this possibility, we run the same experiment with five sets of control prompts.
    We repeat the localization experiment by presenting $M$ with two sentences, each exclusively about one topic. 
    There are $5$ pairs of topics, and each sentence in a pair is sampled at uniform from a pool of $40$, generated by Claude Opus 4.1. As before, we use $20$ chat-templates. We perturb exactly one of the sentences and ask the model a simple comprehension question (see \S\ref{subsec:appendix_loc_controls} for the prompts). 
    Figure~\ref{fig:loc_control_pooled} reports accuracy of every model averaged across all the control questions.
       
    \input{plots/04_localization/controls}

    \textbf{Perturbation upper bounds.}
        \label{subsec:perturb_upper_bounds}
    In the controls all models first display perfect accuracy, as expected for such simple questions. Yet, performance decreases as the perturbations become too salient.
    In particular, every model admits a first value $p_{\max}$ of dropout rate (resp.\ a first value $\sigma_{\max}$ of noise SD) whose associated accuracy is below $95\%$.
    We define the values immediately prior to these to be the thresholds after which the models' capacities are compromised.

    \begin{wraptable}[12]{l}{67mm}
    \vspace{-2ex}
    \centering
    \small
        \begin{tabular}{l@{}ccc}
            \toprule
            Model & $(p_{\min}, p_{\max})$ & $(\sigma_{\min}, \sigma_{\max})$ \\
            \midrule
            \model{gemma31}   & (.14, .68) & (.006, .014) \\
            \model{llama8}    & (.06, .46) & (.003, .010) \\
            \model{mistral24} & (.10, .52) & (.001, .004) \\
            \model{olmo32}    & (.26, .66) & (.02,  .08)   \\
            \model{qwen14}    & (.10, .46) & (.05,  .10)   \\
            \model{qwen32}    & (.18, .54) & (.10,  .22)   \\
            \model{qwen27}    & (.012,.52) & (.009, .04) \\
            \model{seed36}    & (.36, .86) & (.14,  .27)   \\
            \bottomrule
        \end{tabular}
    \caption{\small Ranges of the perturbation strengths.}
    \label{tab:range_perturbation}
    \end{wraptable}
        
    \textbf{Conclusions.}
    Table~\ref{tab:range_perturbation} summarizes the 
    bounds of the perturbation strengths established in \S\ref{subsec:perturb_lower_bounds} and \S\ref{subsec:perturb_upper_bounds}. 
    Observe that $p_{\min}<p_{\max}$ and $\sigma_{\min}<\sigma_{\max}$ always. We conclude that there exists intervals of dropout and noise values for which we can exclude the null hypothesis that models localize the perturbations just by disproportionally picking the perturbed sentence, irrespective of the question.
    For example, \model{qwen32} and -14B reach $100\%$ detection accuracy, so under the null hypothesis would achieve $50\%$ accuracy in the controls, as the correct answer there coincides with the perturbed sentence in half the trials in expectation.
    Yet their control accuracy stays above $95\%$ in $[p_{\min},p_{\max}]$ and $[\sigma_{\min},\sigma_{\max}]$.

    \textbf{Perturbation ranges.}\label{par:perturbation_ranges}
    The intervals $[p_{\min},p_{\max}]$ and $[\sigma_{\min},\sigma_{\max}]$ represent the values of $p$ and $\sigma$ where models both detect the perturbations and retain their ability to answer basic questions.  
        We divide them in $10$ bins of equal width each (\ie, $11$ perturbation magnitudes) and, from now on, we perform experiments only within these ranges.
        We validate these ranges a posteriori: within them, the frequency with which most models report a perturbation rises from its unperturbed level to its maximum (\S\ref{subsec:detection_results}), and few-shot accuracy varies smoothly with $p$ and $\sigma$ (\S\ref{subsec:icl_results}).

    \textbf{Are perturbed sentences less likely?}
    One further null hypothesis is that perturbed sentences are just ``weirder'' (\eg, of lower likelihood). 
    In \S\ref{subsec:appendix_logprobs} we formalize this possibility in two ways, and resolve both options in the negative for most models.

        \subsection{Ablations and variations}\label{subsec:ablations}
        We test the reliability of our findings by performing a few ablations and varying aspects of our experiments. 

        \noindent
        \textbf{Prompt ablations.} To exclude that models succeed at our task mostly because we prompt them to ``expect'' perturbed sentences to come, we perform two prompt ablations where we remove the first parts of the prompt, finding that this is not the case; see \S\ref{subsec:appendix_prompt_ablation}.

        \noindent
        \textbf{Layer ablations.}
        We repeat the localization experiments by perturbing only one third of layers (as opposed to perturbing every layer), and find that ablating only the first third of layers yields marginally \emph{better} results than perturbing every layer (\cf. \S\ref{subsec:appendix_layer_ablation}, Fig.~\ref{fig:appendix_layer_ablation}).

        \noindent
        \textbf{Varying the length of target sentences.}
        We repeat our experiments by varying the number of tokens in the target sentences between $3$ and $23$ (\cf. \S\ref{subsec:appendix_loc_length}). We observe that accuracy stabilizes after $15$ tokens, reported here. Nevertheless, most models achieve $90\%$ accuracy even when the perturbed sentence has only $3$ tokens.

        \noindent
        \textbf{Varying dropout technique.} Applying dropout at the output of the Attention and of the MLP components before the residual connection is closely related to one of the ways dropout is applied in \citet[\S 5.4]{Vaswani+2017}. We provide more details \S\ref{subsec:appendix_dropout}, where we also repeat our first experiment with dropout applied in different ways.

    \needspace{14\baselineskip}
    \subsection{Localization emergence}\label{subsec:loc_emergence}
    \begin{wrapfigure}[12]{r}{55mm}
    \vspace{-5ex}
\centering
{\scriptsize \tikz\draw[green!55!black, thick, mark=*, mark size=1.5] plot coordinates {(0,0) (0.4,0)}; dropout $p{=}0.5$\hspace{1em}\tikz\draw[orange!80!black, thick, mark=*, mark size=1.5] plot coordinates {(0,0) (0.4,0)}; noise $\sigma{=}0.056$}\\[2pt]
\begin{tikzpicture}
\begin{axis}[
    height=3.8cm,
    width=\linewidth,
    grid=major,
    xlabel={\scriptsize Training stage},
    ylabel={\scriptsize Accuracy (\%)},
    xtick={0,1,2,3},
    xticklabels={Base,SFT,DPO,RLVR},
    xticklabel style={font=\scriptsize},
    yticklabel style={font=\scriptsize},
    xmin=-0.3, xmax=3.3,
    ymin=80, ymax=100,
    legend entries={},
]
\addplot[green!55!black, mark=*, mark size=2, line width=1pt, forget plot, error bars/.cd, y dir=both, y explicit, error bar style={line width=1pt, green!55!black}, error mark options={rotate=90, green!55!black, mark size=3, line width=1pt}] coordinates {(0,92.90) +- (0,0.81) (1,98.40) +- (0,0.40) (2,97.20) +- (0,0.52) (3,96.90) +- (0,0.55)};
\addplot[orange!80!black, mark=*, mark size=2, line width=1pt, forget plot, error bars/.cd, y dir=both, y explicit, error bar style={line width=1pt, orange!80!black}, error mark options={rotate=90, orange!80!black, mark size=3, line width=1pt}] coordinates {(0,98.90) +- (0,0.33) (1,99.30) +- (0,0.26) (2,99.70) +- (0,0.17) (3,99.00) +- (0,0.31)};
\end{axis}
\end{tikzpicture}
    \caption{\small Localization accuracy of \model{olmo32} across its training stages.}
    \label{fig:loc_olmo_stages}
    \end{wrapfigure}

        When do language models develop the ability to locate a perturbation? Is it acquired during DPO finetuning, like detecting steering vectors \citep{macar2026mechanisms}?
        Since \model{olmo32}'s training checkpoints are publicly available, we repeat our localization experiment for each training phase
        (\textbf{base} pretraining, \textbf{s}upervised \textbf{f}ine\textbf{t}uning, \textbf{d}irect \textbf{p}reference \textbf{o}ptimization, and \textbf{r}einforcement \textbf{l}earning from \textbf{v}erifiable \textbf{r}ewards).
        Surprisingly, already the base version of \model{olmo32} performs above chance at this task, and the accuracy of every later checkpoint surpasses $95\%$ (Fig.~\ref{fig:loc_olmo_stages}). For completeness, in \S\ref{subsec:appendix_olmo_checkpoints}, we also zoom in on the base stage and look at how \model{olmo32}'s accuracy improves as a function of the pretraining steps. 
    \section{Detection}\label{sec:detection}

    \begin{wrapfigure}[5]{r}{75mm}
        \vspace{-3ex}
        \begin{promptbox}
            \scriptsize
            \textbf{User:} Pay close attention to how you process the following sentence:\\
            \pind
                \text{\textcolor{green!50!black}{Target sentence}}
                \textcolor{gray!70!black}{
                    \textrm{
                        \scriptsize\itshape $\leftarrow$ perturb with either \textcolor{orange!80!black}{dropout} or \textcolor{blue!80!black}{noise}.}
                    }\\
            \pind Was a perturbation applied? \\
            \pind A) YES B) NO\\
            \textbf{Assistant:} The answer is:
        \end{promptbox}
    \end{wrapfigure}

    Can models detect a perturbation when there is nothing to compare it with?
    We now show $M$ only one target sentence and ask whether a perturbation was applied, with the options `Yes' and `No' presented in random order.
    As in \S\ref{sec:localization}, the prompt is the same for dropout and noise.
    We test the $11$ perturbation magnitudes of each model's ranges (\cf. Table~\ref{tab:range_perturbation}), and also test without perturbation.
    On perturbed sentences the correct answer is `Yes', so accuracy coincides with the frequency with which $M$ answers `Yes'. On the unperturbed sentence, the same frequency is the rate of false detections.

    \subsection{Results}\label{subsec:detection_results}
\begin{figure}[ht]
\centering
\tikz\draw[cyan!60!black, thick, mark=*, mark size=1.5] plot coordinates {(0,0) (0.4,0)}; {\small \model{gemma31}}\hspace{1em}
\tikz\draw[red!40, thick, mark=*, mark size=1.5] plot coordinates {(0,0) (0.4,0)}; {\small \model{llama8}}\hspace{1em}
\tikz\draw[yellow!60!black, thick, mark=*, mark size=1.5] plot coordinates {(0,0) (0.4,0)}; {\small \model{mistral24}}\hspace{1em}
\tikz\draw[red!70!black, thick, mark=*, mark size=1.5] plot coordinates {(0,0) (0.4,0)}; {\small \model{olmo32}}\\
\tikz\draw[green!40!black, thick, mark=*, mark size=1.5] plot coordinates {(0,0) (0.4,0)}; {\small \model{qwen14}}\hspace{1em}
\tikz\draw[blue!70!black, thick, mark=*, mark size=1.5] plot coordinates {(0,0) (0.4,0)}; {\small \model{qwen32}}\hspace{1em}
\tikz\draw[teal!70!black, thick, mark=*, mark size=1.5] plot coordinates {(0,0) (0.4,0)}; {\small \model{qwen27}}\hspace{1em}
\tikz\draw[brown!80!black, thick, mark=*, mark size=1.5] plot coordinates {(0,0) (0.4,0)}; {\small \model{seed36}}
\\[3pt]
\begin{minipage}{0.48\textwidth}
\tikzsetnextfilename{detection-main-dropout}
  \begin{tikzpicture}
\begin{axis}[
    height=4.2cm,
    width=\linewidth,
    grid=major,
    xlabel={\small Dropout rate $p$},
    ylabel={\small Answers `Yes' (\%)},
    xmin=-33, xmax=105,
    xtick={-25,0,20,40,60,80,100},
    xticklabels={none,$p_{\min}$,,,,,$p_{\max}$},
    ymin=0.0, ymax=100.0,
]
\addplot[cyan!60!black, fill=cyan!60!black, fill opacity=0.3, draw=none, forget plot] coordinates {(0,0.20) (10,0.34) (20,0.60) (30,0.96) (40,4.40) (50,13.34) (60,22.19) (70,35.09) (80,31.25) (90,19.01) (100,14.06) (100,11.94) (90,16.59) (80,28.35) (70,32.11) (60,19.61) (50,11.26) (40,3.20) (30,0.44) (20,0.20) (10,0.06) (0,0.00)} --cycle;
\addplot[cyan!60!black, mark=*, mark size=1, line width=1pt, forget plot] coordinates {(0,0.10) (10,0.20) (20,0.40) (30,0.70) (40,3.80) (50,12.30) (60,20.90) (70,33.60) (80,29.80) (90,17.80) (100,13.00)};
\addplot[cyan!60!black, only marks, mark=*, mark size=1.5, forget plot] coordinates {(-25,0.20)};
\addplot[red!40, fill=red!40, fill opacity=0.3, draw=none, forget plot] coordinates {(0,2.44) (10,4.08) (20,5.16) (30,9.17) (40,13.13) (50,21.06) (60,26.57) (70,33.27) (80,41.35) (90,49.78) (100,46.77) (100,43.63) (90,46.62) (80,38.25) (70,30.33) (60,23.83) (50,18.54) (40,11.07) (30,7.43) (20,3.84) (10,2.92) (0,1.56)} --cycle;
\addplot[red!40, mark=*, mark size=1, line width=1pt, forget plot] coordinates {(0,2.00) (10,3.50) (20,4.50) (30,8.30) (40,12.10) (50,19.80) (60,25.20) (70,31.80) (80,39.80) (90,48.20) (100,45.20)};
\addplot[red!40, only marks, mark=*, mark size=1.5, forget plot] coordinates {(-25,1.50)};
\addplot[yellow!60!black, fill=yellow!60!black, fill opacity=0.3, draw=none, forget plot] coordinates {(0,4.08) (10,7.60) (20,10.32) (30,20.04) (40,32.56) (50,46.17) (60,63.93) (70,74.80) (80,79.02) (90,79.11) (100,67.50) (100,64.50) (90,76.49) (80,76.38) (70,72.00) (60,60.87) (50,43.03) (40,29.64) (30,17.56) (20,8.48) (10,6.00) (0,2.92)} --cycle;
\addplot[yellow!60!black, mark=*, mark size=1, line width=1pt, forget plot] coordinates {(0,3.50) (10,6.80) (20,9.40) (30,18.80) (40,31.10) (50,44.60) (60,62.40) (70,73.40) (80,77.70) (90,77.80) (100,66.00)};
\addplot[yellow!60!black, only marks, mark=*, mark size=1.5, forget plot] coordinates {(-25,3.50)};
\addplot[red!70!black, fill=red!70!black, fill opacity=0.3, draw=none, forget plot] coordinates {(0,6.33) (10,6.75) (20,8.02) (30,9.07) (40,14.68) (50,14.48) (60,17.57) (70,17.16) (80,14.89) (90,16.44) (100,14.17) (100,12.03) (90,14.16) (80,12.71) (70,14.84) (60,15.23) (50,12.32) (40,12.52) (30,7.33) (20,6.38) (10,5.25) (0,4.87)} --cycle;
\addplot[red!70!black, mark=*, mark size=1, line width=1pt, forget plot] coordinates {(0,5.60) (10,6.00) (20,7.20) (30,8.20) (40,13.60) (50,13.40) (60,16.40) (70,16.00) (80,13.80) (90,15.30) (100,13.10)};
\addplot[red!70!black, only marks, mark=*, mark size=1.5, forget plot] coordinates {(-25,3.30)};
\addplot[green!40!black, fill=green!40!black, fill opacity=0.3, draw=none, forget plot] coordinates {(0,13.03) (10,18.19) (20,23.62) (30,36.81) (40,49.38) (50,66.71) (60,84.77) (70,95.97) (80,98.88) (90,100.00) (100,100.00) (100,99.80) (90,100.00) (80,98.12) (70,94.63) (60,82.43) (50,63.69) (40,46.22) (30,33.79) (20,20.98) (10,15.81) (0,10.97)} --cycle;
\addplot[green!40!black, mark=*, mark size=1, line width=1pt, forget plot] coordinates {(0,12.00) (10,17.00) (20,22.30) (30,35.30) (40,47.80) (50,65.20) (60,83.60) (70,95.30) (80,98.50) (90,100.00) (100,99.90)};
\addplot[green!40!black, only marks, mark=*, mark size=1.5, forget plot] coordinates {(-25,9.10)};
\addplot[blue!70!black, fill=blue!70!black, fill opacity=0.3, draw=none, forget plot] coordinates {(0,17.37) (10,23.11) (20,34.59) (30,43.16) (40,54.88) (50,71.45) (60,81.46) (70,84.48) (80,85.26) (90,85.06) (100,78.62) (100,75.98) (90,82.74) (80,82.94) (70,82.12) (60,78.94) (50,68.55) (40,51.72) (30,40.04) (20,31.61) (10,20.49) (0,15.03)} --cycle;
\addplot[blue!70!black, mark=*, mark size=1, line width=1pt, forget plot] coordinates {(0,16.20) (10,21.80) (20,33.10) (30,41.60) (40,53.30) (50,70.00) (60,80.20) (70,83.30) (80,84.10) (90,83.90) (100,77.30)};
\addplot[blue!70!black, only marks, mark=*, mark size=1.5, forget plot] coordinates {(-25,6.10)};
\addplot[teal!70!black, fill=teal!70!black, fill opacity=0.3, draw=none, forget plot] coordinates {(0,15.20) (10,22.29) (20,29.72) (30,45.07) (40,64.23) (50,81.07) (60,94.75) (70,99.06) (80,99.56) (90,99.06) (100,95.22) (100,93.78) (90,98.34) (80,99.04) (70,98.34) (60,93.25) (50,78.53) (40,61.17) (30,41.93) (20,26.88) (10,19.71) (0,13.00)} --cycle;
\addplot[teal!70!black, mark=*, mark size=1, line width=1pt, forget plot] coordinates {(0,14.10) (10,21.00) (20,28.30) (30,43.50) (40,62.70) (50,79.80) (60,94.00) (70,98.70) (80,99.30) (90,98.70) (100,94.50)};
\addplot[teal!70!black, only marks, mark=*, mark size=1.5, forget plot] coordinates {(-25,12.50)};
\addplot[brown!80!black, fill=brown!80!black, fill opacity=0.3, draw=none, forget plot] coordinates {(0,0.60) (10,1.77) (20,3.21) (30,7.28) (40,24.64) (50,60.16) (60,89.41) (70,94.56) (80,94.85) (90,91.14) (100,84.48) (100,82.12) (90,89.26) (80,93.35) (70,93.04) (60,87.39) (50,57.04) (40,21.96) (30,5.72) (20,2.19) (10,1.03) (0,0.20)} --cycle;
\addplot[brown!80!black, mark=*, mark size=1, line width=1pt, forget plot] coordinates {(0,0.40) (10,1.40) (20,2.70) (30,6.50) (40,23.30) (50,58.60) (60,88.40) (70,93.80) (80,94.10) (90,90.20) (100,83.30)};
\addplot[brown!80!black, only marks, mark=*, mark size=1.5, forget plot] coordinates {(-25,0.10)};
\end{axis}
\end{tikzpicture}
\end{minipage}
\hfill
\begin{minipage}{0.48\textwidth}
\tikzsetnextfilename{detection-main-noise}
  \begin{tikzpicture}
\begin{axis}[
    height=4.2cm,
    width=\linewidth,
    grid=major,
    xlabel={\small Noise SD $\sigma$},
    ylabel={},
    yticklabels={},
    xmin=-33, xmax=105,
    xtick={-25,0,20,40,60,80,100},
    xticklabels={none,$\sigma_{\min}$,,,,,$\sigma_{\max}$},
    ymin=0.0, ymax=100.0,
]
\addplot[cyan!60!black, fill=cyan!60!black, fill opacity=0.3, draw=none, forget plot] coordinates {(0,0.72) (10,1.08) (20,2.33) (30,3.86) (40,2.22) (50,1.08) (60,0.96) (70,0.34) (80,0.00) (90,0.20) (100,0.00) (100,0.00) (90,0.00) (80,0.00) (70,0.06) (60,0.44) (50,0.52) (40,1.38) (30,2.74) (20,1.47) (10,0.52) (0,0.28)} --cycle;
\addplot[cyan!60!black, mark=*, mark size=1, line width=1pt, forget plot] coordinates {(0,0.50) (10,0.80) (20,1.90) (30,3.30) (40,1.80) (50,0.80) (60,0.70) (70,0.20) (80,0.00) (90,0.10) (100,0.00)};
\addplot[cyan!60!black, only marks, mark=*, mark size=1.5, forget plot] coordinates {(-25,0.10)};
\addplot[red!40, fill=red!40, fill opacity=0.3, draw=none, forget plot] coordinates {(0,3.32) (10,3.86) (20,7.28) (30,8.23) (40,14.17) (50,20.24) (60,26.27) (70,36.10) (80,40.74) (90,46.37) (100,46.17) (100,43.03) (90,43.23) (80,37.66) (70,33.10) (60,23.53) (50,17.76) (40,12.03) (30,6.57) (20,5.72) (10,2.74) (0,2.28)} --cycle;
\addplot[red!40, mark=*, mark size=1, line width=1pt, forget plot] coordinates {(0,2.80) (10,3.30) (20,6.50) (30,7.40) (40,13.10) (50,19.00) (60,24.90) (70,34.60) (80,39.20) (90,44.80) (100,44.60)};
\addplot[red!40, only marks, mark=*, mark size=1.5, forget plot] coordinates {(-25,1.50)};
\addplot[yellow!60!black, fill=yellow!60!black, fill opacity=0.3, draw=none, forget plot] coordinates {(0,4.83) (10,4.83) (20,7.28) (30,17.88) (40,32.06) (50,53.78) (60,78.62) (70,95.13) (80,98.88) (90,100.00) (100,100.00) (100,100.00) (90,99.80) (80,98.12) (70,93.67) (60,75.98) (50,50.62) (40,29.14) (30,15.52) (20,5.72) (10,3.57) (0,3.57)} --cycle;
\addplot[yellow!60!black, mark=*, mark size=1, line width=1pt, forget plot] coordinates {(0,4.20) (10,4.20) (20,6.50) (30,16.70) (40,30.60) (50,52.20) (60,77.30) (70,94.40) (80,98.50) (90,99.90) (100,100.00)};
\addplot[yellow!60!black, only marks, mark=*, mark size=1.5, forget plot] coordinates {(-25,3.20)};
\addplot[red!70!black, fill=red!70!black, fill opacity=0.3, draw=none, forget plot] coordinates {(0,2.77) (10,4.40) (20,4.83) (30,8.54) (40,15.10) (50,19.32) (60,24.43) (70,26.88) (80,25.76) (90,25.15) (100,25.96) (100,23.24) (90,22.45) (80,23.04) (70,24.12) (60,21.77) (50,16.88) (40,12.90) (30,6.86) (20,3.57) (10,3.20) (0,1.83)} --cycle;
\addplot[red!70!black, mark=*, mark size=1, line width=1pt, forget plot] coordinates {(0,2.30) (10,3.80) (20,4.20) (30,7.70) (40,14.00) (50,18.10) (60,23.10) (70,25.50) (80,24.40) (90,23.80) (100,24.60)};
\addplot[red!70!black, only marks, mark=*, mark size=1.5, forget plot] coordinates {(-25,2.80)};
\addplot[green!40!black, fill=green!40!black, fill opacity=0.3, draw=none, forget plot] coordinates {(0,12.41) (10,14.89) (20,18.29) (30,26.37) (40,37.01) (50,56.17) (60,77.64) (70,93.52) (80,98.97) (90,100.00) (100,100.00) (100,100.00) (90,100.00) (80,98.23) (70,91.88) (60,74.96) (50,53.03) (40,33.99) (30,23.63) (20,15.91) (10,12.71) (0,10.39)} --cycle;
\addplot[green!40!black, mark=*, mark size=1, line width=1pt, forget plot] coordinates {(0,11.40) (10,13.80) (20,17.10) (30,25.00) (40,35.50) (50,54.60) (60,76.30) (70,92.70) (80,98.60) (90,100.00) (100,100.00)};
\addplot[green!40!black, only marks, mark=*, mark size=1.5, forget plot] coordinates {(-25,8.10)};
\addplot[blue!70!black, fill=blue!70!black, fill opacity=0.3, draw=none, forget plot] coordinates {(0,10.11) (10,17.16) (20,23.62) (30,39.13) (40,65.62) (50,82.34) (60,93.43) (70,97.26) (80,97.36) (90,98.26) (100,96.53) (100,95.27) (90,97.34) (80,96.24) (70,96.14) (60,91.77) (50,79.86) (40,62.58) (30,36.07) (20,20.98) (10,14.84) (0,8.29)} --cycle;
\addplot[blue!70!black, mark=*, mark size=1, line width=1pt, forget plot] coordinates {(0,9.20) (10,16.00) (20,22.30) (30,37.60) (40,64.10) (50,81.10) (60,92.60) (70,96.70) (80,96.80) (90,97.80) (100,95.90)};
\addplot[blue!70!black, only marks, mark=*, mark size=1.5, forget plot] coordinates {(-25,6.30)};
\addplot[teal!70!black, fill=teal!70!black, fill opacity=0.3, draw=none, forget plot] coordinates {(0,13.44) (10,13.86) (20,15.61) (30,23.41) (40,41.15) (50,69.97) (60,88.55) (70,97.36) (80,99.72) (90,100.00) (100,100.00) (100,99.80) (90,100.00) (80,99.28) (70,96.24) (60,86.45) (50,67.03) (40,38.05) (30,20.79) (20,13.39) (10,11.74) (0,11.36)} --cycle;
\addplot[teal!70!black, mark=*, mark size=1, line width=1pt, forget plot] coordinates {(0,12.40) (10,12.80) (20,14.50) (30,22.10) (40,39.60) (50,68.50) (60,87.50) (70,96.80) (80,99.50) (90,100.00) (100,99.90)};
\addplot[teal!70!black, only marks, mark=*, mark size=1.5, forget plot] coordinates {(-25,12.60)};
\addplot[brown!80!black, fill=brown!80!black, fill opacity=0.3, draw=none, forget plot] coordinates {(0,1.43) (10,4.08) (20,13.34) (30,37.62) (40,66.90) (50,90.76) (60,97.63) (70,99.64) (80,99.72) (90,100.00) (100,99.14) (100,98.46) (90,99.80) (80,99.28) (70,99.16) (60,96.57) (50,88.84) (40,63.90) (30,34.58) (20,11.26) (10,2.92) (0,0.77)} --cycle;
\addplot[brown!80!black, mark=*, mark size=1, line width=1pt, forget plot] coordinates {(0,1.10) (10,3.50) (20,12.30) (30,36.10) (40,65.40) (50,89.80) (60,97.10) (70,99.40) (80,99.50) (90,99.90) (100,98.80)};
\addplot[brown!80!black, only marks, mark=*, mark size=1.5, forget plot] coordinates {(-25,0.40)};
\end{axis}
\end{tikzpicture}
\end{minipage}
\vspace{-8pt}
\caption{\small We present the models with one sentence and ask whether a perturbation was applied. Whether the perturbation is \textbf{dropout} (left) or \textbf{Gaussian noise} (right), most models answer `Yes' more often as its strength increases. The $x$-axis spans the range of Table~\ref{tab:range_perturbation}, from $p_{\min}$ to $p_{\max}$, and likewise for $\sigma$ (\emph{none} means no perturbation is applied). Standard errors are below $1.58\%$ and hence not prominent in the figure.}
\label{fig:detection_main}
\end{figure}

        Fig.~\ref{fig:detection_main} plots how often each model answers `Yes' as a function of $p$ and $\sigma$.
        Without perturbation, no model answers `Yes' more than $12.60\%$ of the times, and at $p_{\min}$ and $\sigma_{\min}$ this frequency is still at most $16.20\%$, in line with the definition of the lower bounds (\S\ref{subsec:perturb_lower_bounds}).
        As the perturbation strength increases, so does the frequency of `Yes': at $p_{\max}$ and $\sigma_{\max}$, most models answer `Yes' at least $66.00\%$ (dropout) and $95.90\%$ (noise) of the times.
        For five models, the frequency of `Yes' is roughly a sigmoid of the perturbation strength that lies within the ranges of Table~\ref{tab:range_perturbation}: it starts near its unperturbed level at $p_{\min}$ and $\sigma_{\min}$, and reaches its maximum towards $p_{\max}$ and $\sigma_{\max}$, which validates our choice of the ranges (\S\ref{par:perturbation_ranges}).
        \model{llama8}, \model{olmo32}, and \model{gemma31}, however, do not perform as well. 

    \subsection{Controls}\label{subsec:detection_controls}
    \input{plots/05_detection/controls}
    A null hypothesis is that our perturbations push the models towards answering `Yes', irrespective of the question.
    To exclude this possibility, we replace the prompt with a neutral one (``\emph{I will show you a sentence and then ask you a question.}'') that ends with a question whose answer does not depend on the perturbation.
    The answer is `No' for $4$ questions (\eg, \emph{``Is that sentence written in French?''}), `Yes' for $4$ others (\eg, \emph{``Is that sentence written in English?''}), and undetermined for the remaining $4$ (\eg, \emph{``Before showing you that sentence, I flipped a coin. Did it land on heads?''}). The target sentences, the perturbations, their strengths, and $20$ chat-templates are as before.
    Fig.~\ref{fig:detection_controls} reports how often every model answers `Yes' to each kind of question.

    When the answer is `No', no model answers `Yes' more than $17.10\%$ of the times and no model other than \model{qwen32} and \model{olmo32} exceeds $6.40\%$.
    When the answer is `Yes', every model but \model{llama8} answers correctly at least $98.00\%$ of the times without perturbation, and for most models the frequency of `Yes' only weakly decreases towards the top of the range.
    When the question has no correct answer, the frequency of `Yes' reflects each model's prior, which varies among models. 
    Hence, the perturbations alone do not explain the rising curves of Fig.~\ref{fig:detection_main}.

    \textbf{Training stages.}
    Why does \model{olmo32} rarely report a perturbation, although it localizes it with high accuracy (Fig.~\ref{fig:loc_main_15tok})?
    To test whether the model learns to deny perturbations during post-training (\eg, because of safety tuning disincentivizing self-reports), we repeat the detection experiment and its controls on every training stage of \model{olmo32}.
    The base model answers at chance, with or without perturbation. Furthermore, DPO and RLVR make the model answer `No' more often whenever the question has no clear answer, and not only when we ask about the perturbation, excluding our hypothesis.
    We discuss the details in \S\ref{subsec:detection_olmo_stages}.
    \section{Zero-shot classification}\label{sec:zero_shot}
    \begin{wrapfigure}[5]{r}{75mm}
        \vspace{-3ex}
        \begin{promptbox}
            \scriptsize
            \textbf{User:} Pay attention to the following sentence:\\
            \pind
                \text{\textcolor{green!50!black}{Target sentence}}
                \textcolor{gray!70!black}{
                    \textrm{
                        \scriptsize\itshape $\leftarrow$ perturb with either \textcolor{orange!80!black}{dropout} or \textcolor{blue!80!black}{noise}.}
                    }\\
            \pind What perturbation was applied? \\
            \pind \textcolor{orange!80!black}{A) Dropout} \textcolor{blue!80!black}{B) Noise}\\
            \textbf{Assistant:} The answer is:
        \end{promptbox}
    \end{wrapfigure}
    
    We now administer the much harder task of answering \emph{which} perturbation was applied, by showing models only one sentence, perturbed with either dropout or noise.
    To exclude biases towards disproportionally answering ` A' or ` B' we randomize the assignment between letters (\eg, ` A') and perturbations (\eg, `Dropout'), as well as the order in which the perturbation names appear in the prompt. 
    We also run the same experiments using synonyms of dropout and noise, \eg, 
   `\text{Masking}'
    and 
    `\text{Jitter}'. 
    
    \subsection{Results}\label{subsec:zero_shot_results}
    \vspace{-3ex}
\definecolor{pastelgreen}{RGB}{143,206,143}
\definecolor{pastelgreendark}{RGB}{70,140,70}
\begin{figure}[H]
\centering
\vspace{-5pt}
\def\plotheight{2.8cm}
\def\barswatch#1#2{\tikz[baseline=0ex]{\fill[#1] (0,0) -- (0,0.2) -- (0.2,0.2) -- cycle; \fill[#2] (0,0) -- (0.2,0) -- (0.2,0.2) -- cycle;}}
\def\curveswatch#1{\tikz[baseline=-0.6ex]\draw[#1, thick, mark=*, mark size=1.5] plot coordinates {(0,0) (0.4,0)};}
\begin{subfigure}[b]{0.47\textwidth}
\centering
\mbox{\barswatch{red!70!black}{pastelgreendark}~{\scriptsize Dropout \&\ Noise}}\hspace{1em}%
\mbox{\barswatch{red!40}{pastelgreen}~{\scriptsize Masking \&\ Jitter}}\\[2pt]
\begin{tikzpicture}
\begin{axis}[
    ybar,
    bar width=4.5pt,
    height=5.6cm,
    width=\linewidth,
    grid=major,
    ylabel={\small Avg. accuracy (\%)},
    ymin=40, ymax=70,
    ytick={40,50,60,70},
    extra y ticks={50},
    extra y tick labels={},
    extra y tick style={grid=major, major grid style={gray, dashed, line width=0.5pt}},
    symbolic x coords={\model{gemma31},\model{seed36},\model{qwen32},\model{qwen27},\model{mistral24},\model{qwen14},\model{llama8},\model{olmo32}},
    xtick={\model{gemma31},\model{seed36},\model{qwen32},\model{qwen27},\model{mistral24},\model{qwen14},\model{llama8},\model{olmo32}},
    tick label style={font=\footnotesize},
    xticklabel style={font=\scriptsize, rotate=45, anchor=north east},
    enlarge x limits=0.08,
]
\addplot[red!70!black, fill=red!70!black, fill opacity=0.8, bar shift=-2.5pt, error bars/.cd, y dir=both, y explicit, error bar style={black, line width=0.6pt}] coordinates {(\model{gemma31},58.26) +- (0,0.21) (\model{seed36},52.87) +- (0,0.20) (\model{mistral24},52.22) +- (0,0.30) (\model{qwen14},48.63) +- (0,0.22) (\model{llama8},48.20) +- (0,0.33) (\model{olmo32},43.20) +- (0,0.30)};
\addplot[pastelgreendark, fill=pastelgreendark, fill opacity=0.8, bar shift=-2.5pt, error bars/.cd, y dir=both, y explicit, error bar style={black, line width=0.6pt}] coordinates {(\model{qwen32},52.44) +- (0,0.14) (\model{qwen27},52.44) +- (0,0.33)};
\addplot[red!40, fill=red!40, fill opacity=0.8, bar shift=2.5pt, error bars/.cd, y dir=both, y explicit, error bar style={black, line width=0.6pt}] coordinates {(\model{qwen14},60.33) +- (0,0.31) (\model{mistral24},56.22) +- (0,0.24) (\model{seed36},52.75) +- (0,0.33) (\model{gemma31},51.45) +- (0,0.19) (\model{olmo32},51.25) +- (0,0.33) (\model{llama8},49.76) +- (0,0.34)};
\addplot[pastelgreen, fill=pastelgreen, fill opacity=0.8, bar shift=2.5pt, error bars/.cd, y dir=both, y explicit, error bar style={black, line width=0.6pt}] coordinates {(\model{qwen32},59.86) +- (0,0.32) (\model{qwen27},54.74) +- (0,0.33)};
\end{axis}
\end{tikzpicture}
\label{fig:zero_shot_average}
\end{subfigure}
\hfill
\begin{subfigure}[b]{0.49\textwidth}
\centering
\edef\plotwidth{\the\dimexpr(\linewidth-2.4cm)/2\relax}
\begin{tabular}[b]{@{}l@{}}
\multicolumn{1}{@{}c@{}}{{\scriptsize \model{qwen32}:}\hspace{0.6em}\mbox{\curveswatch{pastelgreendark}~{\scriptsize Dropout/Noise}}\hspace{0.6em}\mbox{\curveswatch{pastelgreen}~{\scriptsize Masking/Jitter}}}\\[2pt]
\begin{tikzpicture}[trim axis right]
\begin{axis}[
    height=\plotheight,
    width=\plotwidth,
    scale only axis,
    grid=major,
    ylabel={\small Accuracy (\%)},
    xmin=0, xmax=100,
    xtick={0,100},
    xticklabels={\makebox[0pt][l]{$p_{\min}$},\makebox[0pt][r]{$p_{\max}$}},
    ymin=0.0, ymax=100.0,
    ytick={0,50,100},
    tick label style={font=\footnotesize},
]
\addplot[pastelgreendark, fill=pastelgreendark, fill opacity=0.3, draw=none, forget plot] coordinates {(0.0,96.4343) (10.0,96.9891) (20.0,97.5394) (30.0,97.3566) (40.0,98.0840) (50.0,97.4481) (60.0,98.8844) (70.0,98.2639) (80.0,99.3146) (90.0,98.7968) (100.0,99.3146) (100.0,98.6854) (90.0,98.0032) (80.0,98.6854) (70.0,97.3361) (60.0,98.1156) (50.0,96.3519) (40.0,97.1160) (30.0,96.2434) (20.0,96.4606) (10.0,95.8109) (0.0,95.1657)} --cycle;
\addplot[pastelgreendark, mark=*, mark size=1, line width=1.5pt, forget plot] coordinates {(0.0,95.8000) (10.0,96.4000) (20.0,97.0000) (30.0,96.8000) (40.0,97.6000) (50.0,96.9000) (60.0,98.5000) (70.0,97.8000) (80.0,99.0000) (90.0,98.4000) (100.0,99.0000)};
\addplot[pastelgreen, fill=pastelgreen, fill opacity=0.3, draw=none, forget plot] coordinates {(0.0,48.7787) (10.0,48.7787) (20.0,55.0773) (30.0,51.0811) (40.0,52.6808) (50.0,53.3801) (60.0,53.2802) (70.0,55.3766) (80.0,58.8642) (90.0,61.8472) (100.0,67.7948) (100.0,64.8052) (90.0,58.7528) (80.0,55.7358) (70.0,52.2234) (60.0,50.1198) (50.0,50.2199) (40.0,49.5192) (30.0,47.9189) (20.0,51.9227) (10.0,45.6213) (0.0,45.6213)} --cycle;
\addplot[pastelgreen, mark=*, mark size=1, line width=1.5pt, forget plot] coordinates {(0.0,47.2000) (10.0,47.2000) (20.0,53.5000) (30.0,49.5000) (40.0,51.1000) (50.0,51.8000) (60.0,51.7000) (70.0,53.8000) (80.0,57.3000) (90.0,60.3000) (100.0,66.3000)};
\addplot[gray, dashed, line width=0.5pt, forget plot] coordinates {(0,50) (100,50)};
\end{axis}
\end{tikzpicture}%
\hspace{0.6cm}%
\begin{tikzpicture}[trim axis left]
\begin{axis}[
    height=\plotheight,
    width=\plotwidth,
    scale only axis,
    grid=major,
    yticklabels={},
    xmin=0, xmax=100,
    xtick={0,100},
    xticklabels={\makebox[0pt][l]{$\sigma_{\min}$},\makebox[0pt][r]{$\sigma_{\max}$}},
    ymin=0.0, ymax=100.0,
    ytick={0,50,100},
    tick label style={font=\footnotesize},
]
\addplot[pastelgreendark, fill=pastelgreendark, fill opacity=0.3, draw=none, forget plot] coordinates {(0.0,5.0486) (10.0,4.0812) (20.0,5.3693) (30.0,6.0085) (40.0,5.9021) (50.0,7.2796) (60.0,7.4906) (70.0,9.3819) (80.0,10.2184) (90.0,12.8202) (100.0,15.6134) (100.0,13.3866) (90.0,10.7798) (80.0,8.3816) (70.0,7.6181) (60.0,5.9094) (50.0,5.7204) (40.0,4.4979) (30.0,4.5915) (20.0,4.0307) (10.0,2.9188) (0.0,3.7514)} --cycle;
\addplot[pastelgreendark, mark=*, mark size=1, line width=1.5pt, forget plot] coordinates {(0.0,4.4000) (10.0,3.5000) (20.0,4.7000) (30.0,5.3000) (40.0,5.2000) (50.0,6.5000) (60.0,6.7000) (70.0,8.5000) (80.0,9.3000) (90.0,11.8000) (100.0,14.5000)};
\addplot[pastelgreen, fill=pastelgreen, fill opacity=0.3, draw=none, forget plot] coordinates {(0.0,55.6758) (10.0,56.3738) (20.0,55.3766) (30.0,60.4559) (40.0,58.1673) (50.0,64.4276) (60.0,68.9811) (70.0,73.9120) (80.0,79.4078) (90.0,82.5330) (100.0,87.7738) (100.0,85.6262) (90.0,80.0670) (80.0,76.7922) (70.0,71.0880) (60.0,66.0189) (50.0,61.3724) (40.0,55.0327) (30.0,57.3441) (20.0,52.2234) (10.0,53.2262) (0.0,52.5242)} --cycle;
\addplot[pastelgreen, mark=*, mark size=1, line width=1.5pt, forget plot] coordinates {(0.0,54.1000) (10.0,54.8000) (20.0,53.8000) (30.0,58.9000) (40.0,56.6000) (50.0,62.9000) (60.0,67.5000) (70.0,72.5000) (80.0,78.1000) (90.0,81.3000) (100.0,86.7000)};
\addplot[gray, dashed, line width=0.5pt, forget plot] coordinates {(0,50) (100,50)};
\end{axis}
\end{tikzpicture}\\[4pt]
\multicolumn{1}{@{}c@{}}{{\scriptsize \model{gemma31}:}\hspace{0.6em}\mbox{\curveswatch{red!70!black}~{\scriptsize Dropout/Noise}}\hspace{0.6em}\mbox{\curveswatch{red!40}~{\scriptsize Masking/Jitter}}}\\[2pt]
\begin{tikzpicture}[trim axis right]
\begin{axis}[
    height=\plotheight,
    width=\plotwidth,
    scale only axis,
    grid=major,
    xlabel={\small Dropout rate $p$},
    ylabel={\small Accuracy (\%)},
    xmin=0, xmax=100,
    xtick={0,100},
    xticklabels={\makebox[0pt][l]{$p_{\min}$},\makebox[0pt][r]{$p_{\max}$}},
    ymin=0.0, ymax=100.0,
    ytick={0,50,100},
    tick label style={font=\footnotesize},
]
\addplot[red!70!black, fill=red!70!black, fill opacity=0.3, draw=none, forget plot] coordinates {(0.0,1.7715) (10.0,1.4298) (20.0,2.1088) (30.0,3.7566) (40.0,9.1724) (50.0,18.8043) (60.0,34.0823) (70.0,48.5783) (80.0,54.2788) (90.0,51.4811) (100.0,41.7505) (100.0,38.6495) (90.0,48.3189) (80.0,51.1212) (70.0,45.4217) (60.0,31.1177) (50.0,16.3957) (40.0,7.4276) (30.0,2.6434) (20.0,1.2912) (10.0,0.7702) (0.0,1.0285)} --cycle;
\addplot[red!70!black, mark=*, mark size=1, line width=1.5pt, forget plot] coordinates {(0.0,1.4000) (10.0,1.1000) (20.0,1.7000) (30.0,3.2000) (40.0,8.3000) (50.0,17.6000) (60.0,32.6000) (70.0,47.0000) (80.0,52.7000) (90.0,49.9000) (100.0,40.2000)};
\addplot[red!40, fill=red!40, fill opacity=0.3, draw=none, forget plot] coordinates {(0.0,77.5467) (10.0,85.9353) (20.0,89.3164) (30.0,95.1271) (40.0,93.2380) (50.0,94.1851) (60.0,96.3415) (70.0,97.8126) (80.0,98.5317) (90.0,97.7217) (100.0,97.8126) (100.0,96.7874) (90.0,96.6783) (80.0,97.6683) (70.0,96.7874) (60.0,95.0585) (50.0,92.6149) (40.0,91.5620) (30.0,93.6729) (20.0,87.2836) (10.0,83.6647) (0.0,74.8533)} --cycle;
\addplot[red!40, mark=*, mark size=1, line width=1.5pt, forget plot] coordinates {(0.0,76.2000) (10.0,84.8000) (20.0,88.3000) (30.0,94.4000) (40.0,92.4000) (50.0,93.4000) (60.0,95.7000) (70.0,97.3000) (80.0,98.1000) (90.0,97.2000) (100.0,97.3000)};
\addplot[gray, dashed, line width=0.5pt, forget plot] coordinates {(0,50) (100,50)};
\end{axis}
\end{tikzpicture}%
\hspace{0.6cm}%
\begin{tikzpicture}[trim axis left]
\begin{axis}[
    height=\plotheight,
    width=\plotwidth,
    scale only axis,
    grid=major,
    xlabel={\small Noise SD $\sigma$},
    yticklabels={},
    xmin=0, xmax=100,
    xtick={0,100},
    xticklabels={\makebox[0pt][l]{$\sigma_{\min}$},\makebox[0pt][r]{$\sigma_{\max}$}},
    ymin=0.0, ymax=100.0,
    ytick={0,50,100},
    tick label style={font=\footnotesize},
]
\addplot[red!70!black, fill=red!70!black, fill opacity=0.3, draw=none, forget plot] coordinates {(0.0,98.8844) (10.0,98.5317) (20.0,98.2639) (30.0,97.9937) (40.0,96.6197) (50.0,96.1556) (60.0,93.9015) (70.0,92.1912) (80.0,90.1815) (90.0,87.0973) (100.0,84.1879) (100.0,81.8121) (90.0,84.9027) (80.0,88.2185) (70.0,90.4088) (60.0,92.2985) (50.0,94.8444) (40.0,95.3803) (30.0,97.0063) (20.0,97.3361) (10.0,97.6683) (0.0,98.1156)} --cycle;
\addplot[red!70!black, mark=*, mark size=1, line width=1.5pt, forget plot] coordinates {(0.0,98.5000) (10.0,98.1000) (20.0,97.8000) (30.0,97.5000) (40.0,96.0000) (50.0,95.5000) (60.0,93.1000) (70.0,91.3000) (80.0,89.2000) (90.0,86.0000) (100.0,83.0000)};
\addplot[red!40, fill=red!40, fill opacity=0.3, draw=none, forget plot] coordinates {(0.0,23.4121) (10.0,21.2649) (20.0,14.4772) (30.0,14.1670) (40.0,14.3738) (50.0,9.0676) (60.0,6.9626) (70.0,6.8568) (80.0,5.3693) (90.0,5.3693) (100.0,5.7957) (100.0,4.4043) (90.0,4.0307) (80.0,4.0307) (70.0,5.3432) (60.0,5.4374) (50.0,7.3324) (40.0,12.2262) (30.0,12.0330) (20.0,12.3228) (10.0,18.7351) (0.0,20.7879)} --cycle;
\addplot[red!40, mark=*, mark size=1, line width=1.5pt, forget plot] coordinates {(0.0,22.1000) (10.0,20.0000) (20.0,13.4000) (30.0,13.1000) (40.0,13.3000) (50.0,8.2000) (60.0,6.2000) (70.0,6.1000) (80.0,4.7000) (90.0,4.7000) (100.0,5.1000)};
\addplot[gray, dashed, line width=0.5pt, forget plot] coordinates {(0,50) (100,50)};
\end{axis}
\end{tikzpicture}
\end{tabular}%
\label{fig:zero_shot_detection}
\end{subfigure}
\caption{\small
We apply {dropout} or add {Gaussian noise} to the activations of a target sentence, ask \emph{``Did we apply dropout or noise?''}, and report accuracy with the Dropout/Noise labels (dark) and the synonyms Masking/Jitter (light). The average (over dropout and noise samples across perturbation strengths) of many models is above chance (\textbf{left}), with \textcolor{pastelgreendark}{two models} (\eg, \textcolor{pastelgreendark}{\model{qwen32}}) increasingly answering \emph{dropout} (or \emph{masking}) when in fact dropout is applied, and likewise for \emph{noise} (or \emph{jitter}) (\textbf{right}), while the \textcolor{red!70!black}{remaining models} (\eg, \textcolor{red!70!black}{\model{gemma31}}) do not.
}
\label{fig:zero_shot_main}
\end{figure}

    As expected, this task proves to be much harder than localizing or detecting a perturbed sentence. In fact, while most models overall perform above chance (Fig.~\ref{fig:zero_shot_main}, left), their accuracy as a function of each perturbation strength is harder to interpret, for it often decreases as the perturbation strength increases (\eg, see \model{gemma31} under noise (Fig.~\ref{fig:zero_shot_main}, right), and \S\ref{subsec:appendix_zero_shot_results} for the other models).
    Two notable exceptions are \model{qwen32} and \model{qwen27}. 
    Look at Fig.~\ref{fig:zero_shot_main}, right: \model{qwen32} has a very high prior to answer dropout over noise: even at $p_{\min}$, the model says `dropout' $95.80\%$ of the times when dropout is applied. Nonetheless, accuracy continues to increase as dropout rate increases, reaching $99.00\%$ at $p_{\max}$ (left plot, \textcolor{pastelgreendark}{dark green} curve). Similarly, despite the low prior to answer `noise' (with the model saying it only $4.40\%$ of the times when noise is injected at magnitude $\sigma_{\min}$), \model{qwen32} reaches $14.50\%$ accuracy in the noise samples at $\sigma_{\max}$ (right plot, \textcolor{pastelgreendark}{dark green} curve).
    This double increasing trend does not seem to be an artifact of the strong prior over dropout. In fact, the model has an almost $50\%$ prior between `masking' and `jitter', synonyms of dropout and noise, respectively. In this case, the double rise in accuracy is even more pronounced (\textcolor{pastelgreen}{light green} curves): 
    for instance, when noise is applied, the model identifies the correct perturbation $86.70\%$ of the times. 
    We remark that \model{qwen32} and \model{qwen27} are the only models with accuracy monotonically increasing in both perturbation strengths (\cf. \S\ref{subsec:appendix_zero_shot_results}), which motivates further controls. 

    \subsection{Controls}\label{subsec:zero_shot_controls}

    \begin{wrapfigure}{l}{75mm}
        \vspace{-3ex}
        \begin{promptbox}
            \scriptsize
            \textbf{User:} Pay attention to the following sentence:\\
            \pind
                \text{\textcolor{green!50!black}{Target sentence}}
                \textcolor{gray!70!black}{
                    \textrm{
                        \scriptsize\itshape $\leftarrow$ perturb with either \textcolor{orange!80!black}{dropout} or \textcolor{blue!80!black}{noise}.}
                    }\\
            \pind What perturbation was applied? \\
            \pind \textcolor[RGB]{180, 140, 20}{A) Foo} \textcolor[RGB]{150, 80, 50}{B) Bar}\\
            \textbf{Assistant:} The answer is:
        \end{promptbox}
        \vspace{1ex}
\centering
{\scriptsize
\tikz[baseline=-0.6ex]\draw[brown!70!black, dashed, line width=0.8pt] plot coordinates {(0,0) (0.4,0)}; Foo/Bar\hspace{0.8em}%
\tikz[baseline=-0.6ex]\draw[blue!60!black, dashed, line width=0.8pt] plot coordinates {(0,0) (0.4,0)}; Rotation/Permutation\\
\tikz[baseline=-0.6ex]\draw[green!70!black, dashed, line width=0.8pt] plot coordinates {(0,0) (0.4,0)}; Vanilla/Chocolate\hspace{0.8em}%
\tikz[baseline=-0.6ex]\draw[cyan!40!blue!70, dashed, line width=0.8pt] plot coordinates {(0,0) (0.4,0)}; X/Y}\\[2pt]
\tikzset{external/export next=false}%
\begin{tikzpicture}[trim axis right]
\begin{axis}[
    scale only axis,
    width=2.4cm,
    height=2.2cm,
    grid=major,
    xlabel={\scriptsize Dropout rate $p$},
    ylabel={\scriptsize 1st label (\%)},
    xmin=0, xmax=100,
    xtick={0,100},
    xticklabels={\makebox[0pt][l]{$p_{\min}$},\makebox[0pt][r]{$p_{\max}$}},
    ymin=0.0, ymax=100.0,
    ytick={0,50,100},
    tick label style={font=\tiny},
    xlabel style={yshift=1ex},
    ylabel style={yshift=-1ex},
]
\addplot[brown!70!black, fill=brown!70!black, fill opacity=0.15, draw=none, forget plot] coordinates {(0.0,90.6612) (10.0,86.5166) (20.0,87.3873) (30.0,84.7709) (40.0,83.2149) (50.0,81.2649) (60.0,79.1142) (70.0,74.8956) (80.0,76.1729) (90.0,75.4853) (100.0,74.5023) (100.0,71.6977) (90.0,72.7147) (80.0,73.4271) (70.0,72.1044) (60.0,76.4858) (50.0,78.7351) (40.0,80.7851) (30.0,82.4291) (20.0,85.2127) (10.0,84.2834) (0.0,88.7388)} --cycle;
\addplot[brown!70!black, dashed, line width=0.8pt, forget plot] coordinates {(0.0,89.7000) (10.0,85.4000) (20.0,86.3000) (30.0,83.6000) (40.0,82.0000) (50.0,80.0000) (60.0,77.8000) (70.0,73.5000) (80.0,74.8000) (90.0,74.1000) (100.0,73.1000)};
\addplot[blue!60!black, fill=blue!60!black, fill opacity=0.15, draw=none, forget plot] coordinates {(0.0,1.6582) (10.0,1.0817) (20.0,0.9636) (30.0,0.4729) (40.0,1.0817) (50.0,0.5996) (60.0,0.4729) (70.0,0.5996) (80.0,0.3413) (90.0,0.5996) (100.0,0.5996) (100.0,0.2004) (90.0,0.2004) (80.0,0.0587) (70.0,0.2004) (60.0,0.1271) (50.0,0.2004) (40.0,0.5183) (30.0,0.1271) (20.0,0.4364) (10.0,0.5183) (0.0,0.9418)} --cycle;
\addplot[blue!60!black, dashed, line width=0.8pt, forget plot] coordinates {(0.0,1.3000) (10.0,0.8000) (20.0,0.7000) (30.0,0.3000) (40.0,0.8000) (50.0,0.4000) (60.0,0.3000) (70.0,0.4000) (80.0,0.2000) (90.0,0.4000) (100.0,0.4000)};
\addplot[green!70!black, fill=green!70!black, fill opacity=0.15, draw=none, forget plot] coordinates {(0.0,61.2511) (10.0,56.0747) (20.0,50.9810) (30.0,44.1637) (40.0,39.6357) (50.0,29.5214) (60.0,25.3506) (70.0,20.1381) (80.0,22.2880) (90.0,27.3871) (100.0,35.3969) (100.0,32.4031) (90.0,24.6129) (80.0,19.7120) (70.0,17.6619) (60.0,22.6494) (50.0,26.6786) (40.0,36.5643) (30.0,41.0363) (20.0,47.8190) (10.0,52.9253) (0.0,58.1489)} --cycle;
\addplot[green!70!black, dashed, line width=0.8pt, forget plot] coordinates {(0.0,59.7000) (10.0,54.5000) (20.0,49.4000) (30.0,42.6000) (40.0,38.1000) (50.0,28.1000) (60.0,24.0000) (70.0,18.9000) (80.0,21.0000) (90.0,26.0000) (100.0,33.9000)};
\addplot[cyan!40!blue!70, fill=cyan!40!blue!70, fill opacity=0.15, draw=none, forget plot] coordinates {(0.0,76.1729) (10.0,73.5183) (20.0,73.7152) (30.0,75.3871) (40.0,71.5478) (50.0,74.3056) (60.0,75.5836) (70.0,72.7305) (80.0,70.8573) (90.0,63.7333) (100.0,59.0632) (100.0,55.9368) (90.0,60.6667) (80.0,67.9427) (70.0,69.8695) (60.0,72.8164) (50.0,71.4944) (40.0,68.6522) (30.0,72.6129) (20.0,70.8848) (10.0,70.6817) (0.0,73.4271)} --cycle;
\addplot[cyan!40!blue!70, dashed, line width=0.8pt, forget plot] coordinates {(0.0,74.8000) (10.0,72.1000) (20.0,72.3000) (30.0,74.0000) (40.0,70.1000) (50.0,72.9000) (60.0,74.2000) (70.0,71.3000) (80.0,69.4000) (90.0,62.2000) (100.0,57.5000)};
\addplot[gray, dashed, line width=0.5pt, forget plot] coordinates {(0,50) (100,50)};
\end{axis}
\end{tikzpicture}%
\hspace{0.6cm}%
\tikzset{external/export next=false}%
\begin{tikzpicture}[trim axis left]
\begin{axis}[
    scale only axis,
    width=2.4cm,
    height=2.2cm,
    grid=major,
    xlabel={\scriptsize Noise SD $\sigma$},
    ylabel={\scriptsize 2nd label (\%)},
    yticklabel pos=right,
    ylabel near ticks,
    xmin=0, xmax=100,
    xtick={0,100},
    xticklabels={\makebox[0pt][l]{$\sigma_{\min}$},\makebox[0pt][r]{$\sigma_{\max}$}},
    ymin=0.0, ymax=100.0,
    ytick={0,50,100},
    tick label style={font=\tiny},
    xlabel style={yshift=1ex},
]
\addplot[brown!70!black, fill=brown!70!black, fill opacity=0.15, draw=none, forget plot] coordinates {(0.0,13.2350) (10.0,12.1973) (20.0,11.4694) (30.0,11.2612) (40.0,13.5458) (50.0,10.2184) (60.0,10.9487) (70.0,10.0095) (80.0,11.7815) (90.0,11.8855) (100.0,15.2005) (100.0,12.9995) (90.0,9.9145) (80.0,9.8185) (70.0,8.1905) (60.0,9.0513) (50.0,8.3816) (40.0,11.4542) (30.0,9.3388) (20.0,9.5306) (10.0,10.2027) (0.0,11.1650)} --cycle;
\addplot[brown!70!black, dashed, line width=0.8pt, forget plot] coordinates {(0.0,12.2000) (10.0,11.2000) (20.0,10.5000) (30.0,10.3000) (40.0,12.5000) (50.0,9.3000) (60.0,10.0000) (70.0,9.1000) (80.0,10.8000) (90.0,10.9000) (100.0,14.1000)};
\addplot[blue!60!black, fill=blue!60!black, fill opacity=0.15, draw=none, forget plot] coordinates {(0.0,99.7996) (10.0,99.3986) (20.0,99.1443) (30.0,99.6442) (40.0,99.8729) (50.0,99.8729) (60.0,99.8729) (70.0,99.9999) (80.0,99.9413) (90.0,100.0000) (100.0,99.9413) (100.0,99.6587) (90.0,100.0000) (80.0,99.6587) (70.0,99.8001) (60.0,99.5271) (50.0,99.5271) (40.0,99.5271) (30.0,99.1558) (20.0,98.4557) (10.0,98.8014) (0.0,99.4004)} --cycle;
\addplot[blue!60!black, dashed, line width=0.8pt, forget plot] coordinates {(0.0,99.6000) (10.0,99.1000) (20.0,98.8000) (30.0,99.4000) (40.0,99.7000) (50.0,99.7000) (60.0,99.7000) (70.0,99.9000) (80.0,99.8000) (90.0,100.0000) (100.0,99.8000)};
\addplot[green!70!black, fill=green!70!black, fill opacity=0.15, draw=none, forget plot] coordinates {(0.0,36.6093) (10.0,39.9380) (20.0,47.8768) (30.0,54.6781) (40.0,66.7063) (50.0,82.8253) (60.0,91.0445) (70.0,95.5957) (80.0,97.7217) (90.0,96.5270) (100.0,96.1556) (100.0,94.8444) (90.0,95.2730) (80.0,96.6783) (70.0,94.2043) (60.0,89.1555) (50.0,80.3747) (40.0,63.6937) (30.0,51.5219) (20.0,44.7232) (10.0,36.8620) (0.0,33.5907)} --cycle;
\addplot[green!70!black, dashed, line width=0.8pt, forget plot] coordinates {(0.0,35.1000) (10.0,38.4000) (20.0,46.3000) (30.0,53.1000) (40.0,65.2000) (50.0,81.6000) (60.0,90.1000) (70.0,94.9000) (80.0,97.2000) (90.0,95.9000) (100.0,95.5000)};
\addplot[cyan!40!blue!70, fill=cyan!40!blue!70, fill opacity=0.15, draw=none, forget plot] coordinates {(0.0,27.8956) (10.0,28.5056) (20.0,26.9801) (30.0,22.9013) (40.0,21.3673) (50.0,17.3651) (60.0,20.0355) (70.0,18.4961) (80.0,21.2649) (90.0,22.9013) (100.0,22.2880) (100.0,19.7120) (90.0,20.2987) (80.0,18.7351) (70.0,16.1039) (60.0,17.5645) (50.0,15.0349) (40.0,18.8327) (30.0,20.2987) (20.0,24.2199) (10.0,25.6944) (0.0,25.1044)} --cycle;
\addplot[cyan!40!blue!70, dashed, line width=0.8pt, forget plot] coordinates {(0.0,26.5000) (10.0,27.1000) (20.0,25.6000) (30.0,21.6000) (40.0,20.1000) (50.0,16.2000) (60.0,18.8000) (70.0,17.3000) (80.0,20.0000) (90.0,21.6000) (100.0,21.0000)};
\addplot[gray, dashed, line width=0.5pt, forget plot] coordinates {(0,50) (100,50)};
\end{axis}
\end{tikzpicture}
        \caption{\small Zero-shot controls of \model{qwen32}: how often the model answers the first label when dropout is applied (left), and the second label when noise is applied (right).}
        \label{fig:zero_shot_controls_wrap}
    \end{wrapfigure}
    For \model{qwen32} and \model{qwen27} we ran $50$ control prompts; for the other models, which do not perform equally well at zero-shot, we ran $6$ controls, \S\ref{subsec:appendix_zero_shot_controls}.
    For instance, we ask ``Did we apply X or Y?'', yet the underlying perturbations remain exclusively dropout or noise.
    We keep track of the number of times the models say `X' (resp.\ `Y') when the underlying perturbation is dropout (resp.\ noise), or vice versa.
    This controls for the possibility that each perturbation steers the activations in a way that systematically favors one label over another, irrespective of its semantic content. For instance, dropout might steer to choose in the same way that it steers to pick `X' over `Y', or vice versa. We also track of the average entropy of the distribution of tokens (\cf. \S\ref{subsec:appendix_zero_shot_entropy}).
    For \model{qwen32} and \model{qwen27}, we found that none of the $50$ controls induce a behavior similar to $\langle\text{Dropout}, \text{Noise}\rangle$ or their synonyms (\cf. Fig.~\ref{fig:zero_shot_controls_wrap}).
    This corroborates that these two models can indeed distinguish dropout from noise. Can the others \emph{learn} to distinguish them?
    \section{Few-shot classification}\label{sec:icl}
    
    \begin{wrapfigure}[8]{r}{.5\linewidth}
    \vspace{-5ex}
    \begin{promptbox}
    \scriptsize
    \textbf{User:} I will teach you to detect perturbations.\\
    \textbf{Assistant:} Understood.\\[4pt]
    $\left\{\begin{array}{@{}l@{}}
        \textbf{User:}\text{ \textcolor{green!50!black}{Sentence\_1}}
            \hspace{1cm}\textrm{\textcolor{gray!70!black}{[\reflectbox{\ensuremath{\rightsquigarrow}} apply \textcolor{orange!80!black}{dropout}]
            }}\\[1pt]
        \textbf{Assistant:}\text{ That was \textcolor{orange!80!black}{dropout}.}\\[3pt]
        \textbf{User:}\text{ \textcolor{green!50!black}{Sentence\_2}}
            \hspace{1cm}\textrm{\textcolor{gray!70!black}{[\reflectbox{\ensuremath{\rightsquigarrow}} add \textcolor{blue!80!black}{Gaussian noise}]}}\\[1pt]
        \textbf{Assistant:}\text{ That was \textcolor{blue!80!black}{noise}.}\\[3pt]
        \quad\vdots
    \end{array}\right\}$
    \textcolor{gray!70!black}
        {$\times k$}
        \\[4pt]
    \textbf{User:} \textcolor{green!50!black}{Test\_sentence}.
    What perturbation was applied?\\
    \pind \textcolor{orange!80!black}{A) Dropout}\\
    \pind \textcolor{blue!80!black}{B) Noise}\\
    \textbf{Assistant:} The answer is:
    \end{promptbox}
    \end{wrapfigure}

    We show models $k \in \{0, 1,3,5,7,9\}$ pairs of labels of the form
    $\langle\textcolor{orange!80!black}{\texttt{dropout}};\textcolor{blue!80!black}{\texttt{noise}}\rangle$, or vice versa, and then ask them to predict the perturbation on a test sentence.
    (Observe that the case with zero examples coincides with the zero-shot experiment presented in \S\ref{sec:zero_shot}.)
    We randomize the assignment between letters and perturbations, the order in which the perturbations are presented in the final question, and the order of dropout and noise in each teaching pair.
    \subsection{Results}\label{subsec:icl_results}
        \textbf{Accuracy with one example.} The left-hand side of Fig.~\ref{fig:few_shots_heatmap_examples} shows accuracy (heat) as a function of both dropout rate $p$ and noise SD $\sigma$ when \model{qwen32} is supervised with $k=1$ example (see \S\ref{subsec:appendix_few_shot_heatmaps} for the analogous plots for the other models and number of examples). 
        Accuracy improves if dropout rate increases while noise SD remains fixed (bottom-right quadrant), and vice versa (top-left quadrant).
        It also rises when both perturbations increase in strength along each other: compare the almost-chance accuracy when both $p$ and $\sigma$ are low (bottom-left quadrant) with the $\geq 60\%$ accuracy when both $p$ and $\sigma$ are high (top-right quadrant). This excludes that \model{qwen32} learns only by virtue of comparing a stronger perturbation against a weaker one. 
        This trend is shared with, \eg, \model{gemma31} or \model{seed36}, but not with \model{llama8} (which always remains around chance) nor with \model{olmo32} (which shows rising accuracy only as dropout rate increases); \S\ref{subsec:appendix_few_shot_heatmaps}. 

        \textbf{The overall effect of learning on all models.} The right-hand side of Fig.~\ref{fig:few_shots_heatmap_examples} shows the average accuracy of every model as a function of the number of example pairs. For instance, the data-point corresponding to $k=1$ and $M=\text{\model{qwen32}}$ (\textcolor{blue!80!black}{blue} curve) represents the average of all the accuracies depicted in the heatmap on the left-hand side of Fig.~\ref{fig:few_shots_heatmap_examples}.
        Most models improve their accuracy as the number of examples increases, with three of them surpassing $75\%$. 
        
        \begin{figure}[ht]
        \centering
        \hfill
        \scalebox{0.80}{%
            \begin{minipage}[c]{0.58\textwidth}
                \centering
\begin{tikzpicture}
\begin{axis}[
    width=0.72\linewidth,
    height=0.72\linewidth,
    xlabel={\small Dropout rate $p$},
    ylabel={\small Noise SD $\sigma$},
    colorbar,
    colormap={pastel_rwg}{rgb255(0cm)=(206,143,143); rgb255(50cm)=(255,255,255); rgb255(100cm)=(143,206,143)},
    point meta min=0.0,
    point meta max=100.0,
    xtick={0,1,2,3,4,5,6,7,8,9,10},
    xticklabels={$p_{\min}$,,,,,,,,,,$p_{\max}$},
    ytick={0,1,2,3,4,5,6,7,8,9,10},
    yticklabels={$\sigma_{\min}$,,,,,,,,,,$\sigma_{\max}$},
    x tick label style={font=\scriptsize},
    y tick label style={font=\scriptsize},
    view={0}{90},
    axis line style={draw=none},
    major tick length=0pt,
    enlargelimits=false,
    colorbar style={font=\tiny},
    title style={font=\small},
    title={\model{qwen32}},
]
\addplot[matrix plot*, mesh/cols=11, mesh/rows=11, point meta=explicit] table[meta=C] {
x y C
0 10 71.60
1 10 72.20
2 10 67.90
3 10 71.40
4 10 67.90
5 10 67.00
6 10 66.00
7 10 62.50
8 10 63.80
9 10 65.50
10 10 63.50
0 9 68.80
1 9 69.40
2 9 69.40
3 9 66.50
4 9 64.90
5 9 66.90
6 9 64.60
7 9 61.40
8 9 61.30
9 9 62.60
10 9 62.30
0 8 70.20
1 8 68.30
2 8 68.40
3 8 65.30
4 8 64.10
5 8 62.90
6 8 62.00
7 8 62.50
8 8 61.50
9 8 62.00
10 8 60.50
0 7 71.50
1 7 65.50
2 7 66.10
3 7 61.10
4 7 59.50
5 7 59.70
6 7 60.70
7 7 58.60
8 7 60.10
9 7 63.90
10 7 65.00
0 6 63.50
1 6 65.50
2 6 64.50
3 6 61.00
4 6 54.80
5 6 58.70
6 6 58.00
7 6 61.30
8 6 64.70
9 6 62.40
10 6 63.00
0 5 62.10
1 5 62.50
2 5 56.10
3 5 56.80
4 5 55.00
5 5 55.20
6 5 55.30
7 5 58.40
8 5 60.20
9 5 63.30
10 5 63.60
0 4 56.70
1 4 54.70
2 4 53.90
3 4 54.10
4 4 50.50
5 4 51.70
6 4 56.60
7 4 55.80
8 4 59.10
9 4 63.40
10 4 63.80
0 3 50.00
1 3 54.20
2 3 52.60
3 3 50.80
4 3 51.80
5 3 51.20
6 3 54.40
7 3 58.40
8 3 61.40
9 3 59.00
10 3 63.90
0 2 47.60
1 2 52.50
2 2 50.20
3 2 48.30
4 2 51.40
5 2 52.40
6 2 56.30
7 2 59.00
8 2 60.00
9 2 65.40
10 2 65.70
0 1 51.30
1 1 52.10
2 1 48.30
3 1 50.70
4 1 51.80
5 1 52.50
6 1 58.20
7 1 60.10
8 1 66.00
9 1 63.20
10 1 64.80
0 0 51.20
1 0 55.00
2 0 50.50
3 0 50.70
4 0 52.10
5 0 56.40
6 0 58.20
7 0 60.00
8 0 63.80
9 0 66.00
10 0 64.40
};
\node[font=\tiny] at (axis cs:0,0) {51.2};
\node[font=\tiny] at (axis cs:1,0) {55.0};
\node[font=\tiny] at (axis cs:2,0) {50.5};
\node[font=\tiny] at (axis cs:3,0) {50.7};
\node[font=\tiny] at (axis cs:4,0) {52.1};
\node[font=\tiny] at (axis cs:5,0) {56.4};
\node[font=\tiny] at (axis cs:6,0) {58.2};
\node[font=\tiny] at (axis cs:7,0) {60.0};
\node[font=\tiny] at (axis cs:8,0) {63.8};
\node[font=\tiny] at (axis cs:9,0) {66.0};
\node[font=\tiny] at (axis cs:10,0) {64.4};
\node[font=\tiny] at (axis cs:0,1) {51.3};
\node[font=\tiny] at (axis cs:1,1) {52.1};
\node[font=\tiny] at (axis cs:2,1) {48.3};
\node[font=\tiny] at (axis cs:3,1) {50.7};
\node[font=\tiny] at (axis cs:4,1) {51.8};
\node[font=\tiny] at (axis cs:5,1) {52.5};
\node[font=\tiny] at (axis cs:6,1) {58.2};
\node[font=\tiny] at (axis cs:7,1) {60.1};
\node[font=\tiny] at (axis cs:8,1) {66.0};
\node[font=\tiny] at (axis cs:9,1) {63.2};
\node[font=\tiny] at (axis cs:10,1) {64.8};
\node[font=\tiny] at (axis cs:0,2) {47.6};
\node[font=\tiny] at (axis cs:1,2) {52.5};
\node[font=\tiny] at (axis cs:2,2) {50.2};
\node[font=\tiny] at (axis cs:3,2) {48.3};
\node[font=\tiny] at (axis cs:4,2) {51.4};
\node[font=\tiny] at (axis cs:5,2) {52.4};
\node[font=\tiny] at (axis cs:6,2) {56.3};
\node[font=\tiny] at (axis cs:7,2) {59.0};
\node[font=\tiny] at (axis cs:8,2) {60.0};
\node[font=\tiny] at (axis cs:9,2) {65.4};
\node[font=\tiny] at (axis cs:10,2) {65.7};
\node[font=\tiny] at (axis cs:0,3) {50.0};
\node[font=\tiny] at (axis cs:1,3) {54.2};
\node[font=\tiny] at (axis cs:2,3) {52.6};
\node[font=\tiny] at (axis cs:3,3) {50.8};
\node[font=\tiny] at (axis cs:4,3) {51.8};
\node[font=\tiny] at (axis cs:5,3) {51.2};
\node[font=\tiny] at (axis cs:6,3) {54.4};
\node[font=\tiny] at (axis cs:7,3) {58.4};
\node[font=\tiny] at (axis cs:8,3) {61.4};
\node[font=\tiny] at (axis cs:9,3) {59.0};
\node[font=\tiny] at (axis cs:10,3) {63.9};
\node[font=\tiny] at (axis cs:0,4) {56.7};
\node[font=\tiny] at (axis cs:1,4) {54.7};
\node[font=\tiny] at (axis cs:2,4) {53.9};
\node[font=\tiny] at (axis cs:3,4) {54.1};
\node[font=\tiny] at (axis cs:4,4) {50.5};
\node[font=\tiny] at (axis cs:5,4) {51.7};
\node[font=\tiny] at (axis cs:6,4) {56.6};
\node[font=\tiny] at (axis cs:7,4) {55.8};
\node[font=\tiny] at (axis cs:8,4) {59.1};
\node[font=\tiny] at (axis cs:9,4) {63.4};
\node[font=\tiny] at (axis cs:10,4) {63.8};
\node[font=\tiny] at (axis cs:0,5) {62.1};
\node[font=\tiny] at (axis cs:1,5) {62.5};
\node[font=\tiny] at (axis cs:2,5) {56.1};
\node[font=\tiny] at (axis cs:3,5) {56.8};
\node[font=\tiny] at (axis cs:4,5) {55.0};
\node[font=\tiny] at (axis cs:5,5) {55.2};
\node[font=\tiny] at (axis cs:6,5) {55.3};
\node[font=\tiny] at (axis cs:7,5) {58.4};
\node[font=\tiny] at (axis cs:8,5) {60.2};
\node[font=\tiny] at (axis cs:9,5) {63.3};
\node[font=\tiny] at (axis cs:10,5) {63.6};
\node[font=\tiny] at (axis cs:0,6) {63.5};
\node[font=\tiny] at (axis cs:1,6) {65.5};
\node[font=\tiny] at (axis cs:2,6) {64.5};
\node[font=\tiny] at (axis cs:3,6) {61.0};
\node[font=\tiny] at (axis cs:4,6) {54.8};
\node[font=\tiny] at (axis cs:5,6) {58.7};
\node[font=\tiny] at (axis cs:6,6) {58.0};
\node[font=\tiny] at (axis cs:7,6) {61.3};
\node[font=\tiny] at (axis cs:8,6) {64.7};
\node[font=\tiny] at (axis cs:9,6) {62.4};
\node[font=\tiny] at (axis cs:10,6) {63.0};
\node[font=\tiny] at (axis cs:0,7) {71.5};
\node[font=\tiny] at (axis cs:1,7) {65.5};
\node[font=\tiny] at (axis cs:2,7) {66.1};
\node[font=\tiny] at (axis cs:3,7) {61.1};
\node[font=\tiny] at (axis cs:4,7) {59.5};
\node[font=\tiny] at (axis cs:5,7) {59.7};
\node[font=\tiny] at (axis cs:6,7) {60.7};
\node[font=\tiny] at (axis cs:7,7) {58.6};
\node[font=\tiny] at (axis cs:8,7) {60.1};
\node[font=\tiny] at (axis cs:9,7) {63.9};
\node[font=\tiny] at (axis cs:10,7) {65.0};
\node[font=\tiny] at (axis cs:0,8) {70.2};
\node[font=\tiny] at (axis cs:1,8) {68.3};
\node[font=\tiny] at (axis cs:2,8) {68.4};
\node[font=\tiny] at (axis cs:3,8) {65.3};
\node[font=\tiny] at (axis cs:4,8) {64.1};
\node[font=\tiny] at (axis cs:5,8) {62.9};
\node[font=\tiny] at (axis cs:6,8) {62.0};
\node[font=\tiny] at (axis cs:7,8) {62.5};
\node[font=\tiny] at (axis cs:8,8) {61.5};
\node[font=\tiny] at (axis cs:9,8) {62.0};
\node[font=\tiny] at (axis cs:10,8) {60.5};
\node[font=\tiny] at (axis cs:0,9) {68.8};
\node[font=\tiny] at (axis cs:1,9) {69.4};
\node[font=\tiny] at (axis cs:2,9) {69.4};
\node[font=\tiny] at (axis cs:3,9) {66.5};
\node[font=\tiny] at (axis cs:4,9) {64.9};
\node[font=\tiny] at (axis cs:5,9) {66.9};
\node[font=\tiny] at (axis cs:6,9) {64.6};
\node[font=\tiny] at (axis cs:7,9) {61.4};
\node[font=\tiny] at (axis cs:8,9) {61.3};
\node[font=\tiny] at (axis cs:9,9) {62.6};
\node[font=\tiny] at (axis cs:10,9) {62.3};
\node[font=\tiny] at (axis cs:0,10) {71.6};
\node[font=\tiny] at (axis cs:1,10) {72.2};
\node[font=\tiny] at (axis cs:2,10) {67.9};
\node[font=\tiny] at (axis cs:3,10) {71.4};
\node[font=\tiny] at (axis cs:4,10) {67.9};
\node[font=\tiny] at (axis cs:5,10) {67.0};
\node[font=\tiny] at (axis cs:6,10) {66.0};
\node[font=\tiny] at (axis cs:7,10) {62.5};
\node[font=\tiny] at (axis cs:8,10) {63.8};
\node[font=\tiny] at (axis cs:9,10) {65.5};
\node[font=\tiny] at (axis cs:10,10) {63.5};
\end{axis}
\end{tikzpicture}
            \end{minipage}
        }
        \hfill
        \scalebox{0.950}{%
            \hspace{-2em}
            \begin{minipage}[c]{0.58\textwidth}
                \centering
%
\begin{tikzpicture}[baseline=(current bounding box.center)]
\begin{axis}[
    height=5.2cm,
    width=4cm,
    grid=major,
    xlabel={\small teaching pairs},
    ylabel={\small Accuracy (\%)},
    xtick={0,1,3,5,7,9},
    legend entries={},
]
\addplot[red!40, fill=red!40, fill opacity=0.3, draw=none, forget plot] coordinates {(0,49.53) (1,50.15) (3,50.33) (5,51.00) (7,50.92) (9,50.88) (9,50.59) (7,50.63) (5,50.71) (3,50.05) (1,49.86) (0,49.24)} --cycle;
\addplot[red!40, mark=*, mark size=1.5, line width=1.5pt, forget plot] coordinates {(0,49.39) (1,50.00) (3,50.19) (5,50.86) (7,50.77) (9,50.73)};
\addplot[green!40!black, fill=green!40!black, fill opacity=0.3, draw=none, forget plot] coordinates {(0,50.14) (1,50.94) (3,54.60) (5,57.03) (7,59.01) (9,59.81) (9,59.53) (7,58.73) (5,56.74) (3,54.31) (1,50.65) (0,49.85)} --cycle;
\addplot[green!40!black, mark=*, mark size=1.5, line width=1.5pt, forget plot] coordinates {(0,49.99) (1,50.79) (3,54.46) (5,56.88) (7,58.87) (9,59.67)};
\addplot[blue!70!black, fill=blue!70!black, fill opacity=0.3, draw=none, forget plot] coordinates {(0,56.92) (1,60.20) (3,61.12) (5,61.48) (7,62.94) (9,63.91) (9,63.64) (7,62.67) (5,61.20) (3,60.84) (1,59.92) (0,56.63)} --cycle;
\addplot[blue!70!black, mark=*, mark size=1.5, line width=1.5pt, forget plot] coordinates {(0,56.78) (1,60.06) (3,60.98) (5,61.34) (7,62.81) (9,63.77)};
\addplot[red!70!black, fill=red!70!black, fill opacity=0.3, draw=none, forget plot] coordinates {(0,47.42) (1,52.62) (3,52.57) (5,53.40) (7,53.65) (9,54.14) (9,53.85) (7,53.36) (5,53.12) (3,52.28) (1,52.33) (0,47.13)} --cycle;
\addplot[red!70!black, mark=*, mark size=1.5, line width=1.5pt, forget plot] coordinates {(0,47.27) (1,52.47) (3,52.42) (5,53.26) (7,53.50) (9,53.99)};
\addplot[teal!70!black, fill=teal!70!black, fill opacity=0.3, draw=none, forget plot] coordinates {(0,53.47) (1,57.37) (3,69.07) (5,73.94) (7,76.60) (9,78.05) (9,77.81) (7,76.36) (5,73.69) (3,68.80) (1,57.08) (0,53.18)} --cycle;
\addplot[teal!70!black, mark=*, mark size=1.5, line width=1.5pt, forget plot] coordinates {(0,53.32) (1,57.23) (3,68.93) (5,73.82) (7,76.48) (9,77.93)};
\addplot[brown!80!black, fill=brown!80!black, fill opacity=0.3, draw=none, forget plot] coordinates {(0,53.78) (1,64.00) (3,72.23) (5,75.69) (7,75.95) (9,77.19) (9,76.95) (7,75.70) (5,75.44) (3,71.97) (1,63.72) (0,53.50)} --cycle;
\addplot[brown!80!black, mark=*, mark size=1.5, line width=1.5pt, forget plot] coordinates {(0,53.64) (1,63.86) (3,72.10) (5,75.57) (7,75.82) (9,77.07)};
\addplot[yellow!60!black, fill=yellow!60!black, fill opacity=0.3, draw=none, forget plot] coordinates {(0,53.77) (1,55.02) (3,58.64) (5,60.57) (7,62.03) (9,63.75) (9,63.47) (7,61.76) (5,60.29) (3,58.35) (1,54.73) (0,53.48)} --cycle;
\addplot[yellow!60!black, mark=*, mark size=1.5, line width=1.5pt, forget plot] coordinates {(0,53.63) (1,54.87) (3,58.50) (5,60.43) (7,61.90) (9,63.61)};
\addplot[cyan!60!black, fill=cyan!60!black, fill opacity=0.3, draw=none, forget plot] coordinates {(0,57.83) (1,68.57) (3,80.79) (5,86.11) (7,88.82) (9,90.31) (9,90.14) (7,88.64) (5,85.91) (3,80.57) (1,68.31) (0,57.55)} --cycle;
\addplot[cyan!60!black, mark=*, mark size=1.5, line width=1.5pt, forget plot] coordinates {(0,57.69) (1,68.44) (3,80.68) (5,86.01) (7,88.73) (9,90.22)};
\addplot[gray, dashed, line width=0.5pt, forget plot] coordinates {(0,50) (9,50)};
\end{axis}
\end{tikzpicture}
{ 
\small
\def\drawsymbol[#1]{\tikz\draw[#1,thick, mark=*, mark size=1.5] plot coordinates {(0,0) (0.4,0)};}
\begin{tabular}{@{}ll@{}}
\drawsymbol[cyan!60!black] &\model{gemma31} \\
\drawsymbol[red!40] &\model{llama8} \\
\drawsymbol[yellow!60!black] &\model{mistral24} \\
\drawsymbol[red!70!black] &\model{olmo32} \\
\drawsymbol[green!40!black] &\model{qwen14} \\
\drawsymbol[blue!70!black] &\model{qwen32} \\
\drawsymbol[teal!70!black] &\model{qwen27} \\
\drawsymbol[brown!80!black] &\model{seed36}
\end{tabular}
}
            \end{minipage}%
            }
        \caption{
            \small 
            \textbf{Left:} the accuracy of \model{qwen32} with a single pair of in-context examples, as a function of dropout rate and noise SD (standard errors are not depicted but below $1.58\%$). 
            \textbf{Right:} the average accuracy of every model as a function of number of in-context examples (standard errors are below $0.15\%$ and hence not visible).
            }
        \label{fig:few_shots_heatmap_examples}
        \end{figure}
        
        \subsection{Controls}\label{subsec:icl_controls}

        We have seen that the difference between dropout and noise can be learned in-context. 
        But should we understand this as (a) sharpening the model's latent understanding of the concepts of dropout and Gaussian noise, or as (b) a new signal whose meaning has nothing to do with what the model already knows about these two perturbations?
        To the extent that a model $M$ learns to separate dropout from noise even when dropout is mislabeled as \textcolor{blue!80!black}{noise} and noise is mislabeled as \textcolor{orange!80!black}{dropout},
        we should adopt interpretation (b).
        This task of identifying perturbations with the incorrect labels can be thought of as a kind of Stroop test (\citeyear{stroop1935studies}).
        \unskip\footnote{In Stroop's classical psychological experiment, participants are given a list of color words printed in other colors (\eg, \textcolor{green!50!black}{blue}, \textcolor{red!80!black}{white}, \textcolor{blue!50!black}{orange}), and asked to list the actual colors (\eg, ``green, red, blue'') rather than reading the printed words.
        Literate participants are much slower and make more mistakes when the labels differ.
        }
        Fig.~\ref{fig:icl_flipped_curves} depicts accuracy as a function of the number of examples, with both correct and inverted labels, for every model (see \S\ref{subsec:appendix_delta_heatmaps} for more details). Notably, the same models whose zero-shot accuracy is above chance (recall \S\ref{subsec:zero_shot_results}, Fig.~\ref{fig:zero_shot_main}, left) do learn even when supervised with wrong labels, but do so worse, or slower than with correct ones. By contrast, \model{llama8} and \model{olmo32}'s accuracy remains flat around chance. For the Qwen models, this suggests a prior towards the correct ones---thus corroborating the findings of \S\ref{subsec:zero_shot_results}. The cases of, \eg, \model{gemma31} or \model{mistral24}, are harder to interpret, for this apparently positive signal could manifest even with control labels, analogously to \S\ref{subsec:zero_shot_results} Fig.~\ref{fig:zero_shot_main} (right). Accordingly, we run one last suite of controls. 
        \begin{figure}[tb]
        \centering
            {\scriptsize \tikz\draw[gray, thick, mark=*, mark size=1] plot coordinates {(0,0) (0.4,0)}; correct labels\hspace{1em}%
            \tikz\draw[gray, dashed, mark=*, mark size=1] plot coordinates {(0,0) (0.4,0)}; inverted labels}\\[2pt]
            \tikzset{external/export next=false}%
\begin{tikzpicture}[trim axis right]
\begin{axis}[
    scale only axis,
    width=1.75cm,
    height=2.65cm,
    grid=major,
    title={\model{gemma31}},
    title style={font=\scriptsize, yshift=-1.2ex},
    ylabel={\scriptsize Accuracy (\%)}, ylabel style={yshift=-0.8ex},
    xtick={0,1,3,5,7,9}, xmin=0, xmax=9, ymin=45, ymax=95,
    tick label style={font=\tiny},
]
\addplot[cyan!60!black, fill=cyan!60!black, fill opacity=0.3, draw=none, forget plot] coordinates {(0,57.83) (1,68.57) (3,80.79) (5,86.11) (7,88.82) (9,90.31) (9,90.14) (7,88.64) (5,85.91) (3,80.57) (1,68.31) (0,57.55)} --cycle;
\addplot[cyan!60!black, mark=*, mark size=1, line width=1.2pt, forget plot] coordinates {(0,57.69) (1,68.44) (3,80.68) (5,86.01) (7,88.73) (9,90.22)};
\addplot[cyan!60!black, fill=cyan!60!black, fill opacity=0.15, draw=none, forget plot] coordinates {(1,57.79) (3,68.31) (5,77.56) (7,84.02) (9,87.75) (9,87.56) (7,83.81) (5,77.32) (3,68.05) (1,57.51)} --cycle;
\addplot[cyan!60!black, dashed, mark=o, mark size=1, line width=0.9pt, forget plot] coordinates {(1,57.65) (3,68.18) (5,77.44) (7,83.91) (9,87.66)};
\addplot[gray, dashed, line width=0.5pt, forget plot] coordinates {(0,50) (9,50)};
\end{axis}
\end{tikzpicture}%
%
            \hfill
            \tikzset{external/export next=false}%
\begin{tikzpicture}[trim axis left, trim axis right]
\begin{axis}[
    scale only axis,
    width=1.75cm,
    height=2.65cm,
    grid=major,
    title={\model{llama8}},
    title style={font=\scriptsize, yshift=-1.2ex},
    ylabel={}, yticklabels={},
    xtick={0,1,3,5,7,9}, xmin=0, xmax=9, ymin=45, ymax=95,
    tick label style={font=\tiny},
]
\addplot[red!40, fill=red!40, fill opacity=0.3, draw=none, forget plot] coordinates {(0,49.53) (1,50.15) (3,50.33) (5,51.00) (7,50.92) (9,50.88) (9,50.59) (7,50.63) (5,50.71) (3,50.05) (1,49.86) (0,49.24)} --cycle;
\addplot[red!40, mark=*, mark size=1, line width=1.2pt, forget plot] coordinates {(0,49.39) (1,50.00) (3,50.19) (5,50.86) (7,50.77) (9,50.73)};
\addplot[red!40, fill=red!40, fill opacity=0.15, draw=none, forget plot] coordinates {(1,50.44) (3,51.13) (5,51.25) (7,51.24) (9,51.63) (9,51.34) (7,50.96) (5,50.97) (3,50.85) (1,50.15)} --cycle;
\addplot[red!40, dashed, mark=o, mark size=1, line width=0.9pt, forget plot] coordinates {(1,50.29) (3,50.99) (5,51.11) (7,51.10) (9,51.48)};
\addplot[gray, dashed, line width=0.5pt, forget plot] coordinates {(0,50) (9,50)};
\end{axis}
\end{tikzpicture}%
%
            \hfill
            \tikzset{external/export next=false}%
\begin{tikzpicture}[trim axis left, trim axis right]
\begin{axis}[
    scale only axis,
    width=1.75cm,
    height=2.65cm,
    grid=major,
    title={\model{mistral24}},
    title style={font=\scriptsize, yshift=-1.2ex},
    ylabel={}, yticklabels={},
    xtick={0,1,3,5,7,9}, xmin=0, xmax=9, ymin=45, ymax=95,
    tick label style={font=\tiny},
]
\addplot[yellow!60!black, fill=yellow!60!black, fill opacity=0.3, draw=none, forget plot] coordinates {(0,53.77) (1,55.02) (3,58.64) (5,60.57) (7,62.03) (9,63.75) (9,63.47) (7,61.76) (5,60.29) (3,58.35) (1,54.73) (0,53.48)} --cycle;
\addplot[yellow!60!black, mark=*, mark size=1, line width=1.2pt, forget plot] coordinates {(0,53.63) (1,54.87) (3,58.50) (5,60.43) (7,61.90) (9,63.61)};
\addplot[yellow!60!black, fill=yellow!60!black, fill opacity=0.15, draw=none, forget plot] coordinates {(1,48.98) (3,53.59) (5,55.37) (7,56.53) (9,57.84) (9,57.55) (7,56.24) (5,55.08) (3,53.31) (1,48.69)} --cycle;
\addplot[yellow!60!black, dashed, mark=o, mark size=1, line width=0.9pt, forget plot] coordinates {(1,48.83) (3,53.45) (5,55.23) (7,56.38) (9,57.70)};
\addplot[gray, dashed, line width=0.5pt, forget plot] coordinates {(0,50) (9,50)};
\end{axis}
\end{tikzpicture}%
%
            \hfill
            \tikzset{external/export next=false}%
\begin{tikzpicture}[trim axis left, trim axis right]
\begin{axis}[
    scale only axis,
    width=1.75cm,
    height=2.65cm,
    grid=major,
    title={\model{olmo32}},
    title style={font=\scriptsize, yshift=-1.2ex},
    ylabel={}, yticklabels={},
    xtick={0,1,3,5,7,9}, xmin=0, xmax=9, ymin=45, ymax=95,
    tick label style={font=\tiny},
]
\addplot[red!70!black, fill=red!70!black, fill opacity=0.3, draw=none, forget plot] coordinates {(0,47.42) (1,52.62) (3,52.57) (5,53.40) (7,53.65) (9,54.14) (9,53.85) (7,53.36) (5,53.12) (3,52.28) (1,52.33) (0,47.13)} --cycle;
\addplot[red!70!black, mark=*, mark size=1, line width=1.2pt, forget plot] coordinates {(0,47.27) (1,52.47) (3,52.42) (5,53.26) (7,53.50) (9,53.99)};
\addplot[red!70!black, fill=red!70!black, fill opacity=0.15, draw=none, forget plot] coordinates {(1,49.57) (3,52.13) (5,52.77) (7,53.42) (9,53.29) (9,53.01) (7,53.14) (5,52.48) (3,51.85) (1,49.28)} --cycle;
\addplot[red!70!black, dashed, mark=o, mark size=1, line width=0.9pt, forget plot] coordinates {(1,49.43) (3,51.99) (5,52.63) (7,53.28) (9,53.15)};
\addplot[gray, dashed, line width=0.5pt, forget plot] coordinates {(0,50) (9,50)};
\end{axis}
\end{tikzpicture}%
%
            \hfill
            \tikzset{external/export next=false}%
\begin{tikzpicture}[trim axis left, trim axis right]
\begin{axis}[
    scale only axis,
    width=1.75cm,
    height=2.65cm,
    grid=major,
    title={\model{qwen14}},
    title style={font=\scriptsize, yshift=-1.2ex},
    ylabel={}, yticklabels={},
    xtick={0,1,3,5,7,9}, xmin=0, xmax=9, ymin=45, ymax=95,
    tick label style={font=\tiny},
]
\addplot[green!40!black, fill=green!40!black, fill opacity=0.3, draw=none, forget plot] coordinates {(0,50.14) (1,50.94) (3,54.60) (5,57.03) (7,59.01) (9,59.81) (9,59.53) (7,58.73) (5,56.74) (3,54.31) (1,50.65) (0,49.85)} --cycle;
\addplot[green!40!black, mark=*, mark size=1, line width=1.2pt, forget plot] coordinates {(0,49.99) (1,50.79) (3,54.46) (5,56.88) (7,58.87) (9,59.67)};
\addplot[green!40!black, fill=green!40!black, fill opacity=0.15, draw=none, forget plot] coordinates {(1,51.40) (3,53.92) (5,56.47) (7,58.25) (9,58.68) (9,58.40) (7,57.96) (5,56.18) (3,53.63) (1,51.11)} --cycle;
\addplot[green!40!black, dashed, mark=o, mark size=1, line width=0.9pt, forget plot] coordinates {(1,51.25) (3,53.78) (5,56.33) (7,58.11) (9,58.54)};
\addplot[gray, dashed, line width=0.5pt, forget plot] coordinates {(0,50) (9,50)};
\end{axis}
\end{tikzpicture}%
%
            \hfill
            \tikzset{external/export next=false}%
\begin{tikzpicture}[trim axis left, trim axis right]
\begin{axis}[
    scale only axis,
    width=1.75cm,
    height=2.65cm,
    grid=major,
    title={\model{qwen32}},
    title style={font=\scriptsize, yshift=-1.2ex},
    ylabel={}, yticklabels={},
    xtick={0,1,3,5,7,9}, xmin=0, xmax=9, ymin=45, ymax=95,
    tick label style={font=\tiny},
]
\addplot[blue!70!black, fill=blue!70!black, fill opacity=0.3, draw=none, forget plot] coordinates {(0,56.92) (1,60.20) (3,61.12) (5,61.48) (7,62.94) (9,63.91) (9,63.64) (7,62.67) (5,61.20) (3,60.84) (1,59.92) (0,56.63)} --cycle;
\addplot[blue!70!black, mark=*, mark size=1, line width=1.2pt, forget plot] coordinates {(0,56.78) (1,60.06) (3,60.98) (5,61.34) (7,62.81) (9,63.77)};
\addplot[blue!70!black, fill=blue!70!black, fill opacity=0.15, draw=none, forget plot] coordinates {(1,50.54) (3,55.35) (5,57.88) (7,58.84) (9,59.95) (9,59.67) (7,58.56) (5,57.60) (3,55.06) (1,50.26)} --cycle;
\addplot[blue!70!black, dashed, mark=o, mark size=1, line width=0.9pt, forget plot] coordinates {(1,50.40) (3,55.20) (5,57.74) (7,58.70) (9,59.81)};
\addplot[gray, dashed, line width=0.5pt, forget plot] coordinates {(0,50) (9,50)};
\end{axis}
\end{tikzpicture}%
%
            \hfill
            \tikzset{external/export next=false}%
\begin{tikzpicture}[trim axis left, trim axis right]
\begin{axis}[
    scale only axis,
    width=1.75cm,
    height=2.65cm,
    grid=major,
    title={\model{qwen27}},
    title style={font=\scriptsize, yshift=-1.2ex},
    ylabel={}, yticklabels={},
    xtick={0,1,3,5,7,9}, xmin=0, xmax=9, ymin=45, ymax=95,
    tick label style={font=\tiny},
]
\addplot[teal!70!black, fill=teal!70!black, fill opacity=0.3, draw=none, forget plot] coordinates {(0,53.47) (1,57.37) (3,69.07) (5,73.94) (7,76.60) (9,78.05) (9,77.81) (7,76.36) (5,73.69) (3,68.80) (1,57.08) (0,53.18)} --cycle;
\addplot[teal!70!black, mark=*, mark size=1, line width=1.2pt, forget plot] coordinates {(0,53.32) (1,57.23) (3,68.93) (5,73.82) (7,76.48) (9,77.93)};
\addplot[teal!70!black, fill=teal!70!black, fill opacity=0.15, draw=none, forget plot] coordinates {(1,51.83) (3,66.86) (5,74.05) (7,77.21) (9,78.46) (9,78.22) (7,76.97) (5,73.79) (3,66.59) (1,51.54)} --cycle;
\addplot[teal!70!black, dashed, mark=o, mark size=1, line width=0.9pt, forget plot] coordinates {(1,51.68) (3,66.73) (5,73.92) (7,77.09) (9,78.34)};
\addplot[gray, dashed, line width=0.5pt, forget plot] coordinates {(0,50) (9,50)};
\end{axis}
\end{tikzpicture}%
%
            \hfill
            \tikzset{external/export next=false}%
\begin{tikzpicture}[trim axis left, trim axis right]
\begin{axis}[
    scale only axis,
    width=1.75cm,
    height=2.65cm,
    grid=major,
    title={\model{seed36}},
    title style={font=\scriptsize, yshift=-1.2ex},
    ylabel={}, yticklabels={},
    xtick={0,1,3,5,7,9}, xmin=0, xmax=9, ymin=45, ymax=95,
    tick label style={font=\tiny},
]
\addplot[brown!80!black, fill=brown!80!black, fill opacity=0.3, draw=none, forget plot] coordinates {(0,53.78) (1,64.00) (3,72.23) (5,75.69) (7,75.95) (9,77.19) (9,76.95) (7,75.70) (5,75.44) (3,71.97) (1,63.72) (0,53.50)} --cycle;
\addplot[brown!80!black, mark=*, mark size=1, line width=1.2pt, forget plot] coordinates {(0,53.64) (1,63.86) (3,72.10) (5,75.57) (7,75.82) (9,77.07)};
\addplot[brown!80!black, fill=brown!80!black, fill opacity=0.15, draw=none, forget plot] coordinates {(1,50.02) (3,59.46) (5,64.57) (7,66.94) (9,69.00) (9,68.73) (7,66.67) (5,64.30) (3,59.18) (1,49.74)} --cycle;
\addplot[brown!80!black, dashed, mark=o, mark size=1, line width=0.9pt, forget plot] coordinates {(1,49.88) (3,59.32) (5,64.44) (7,66.81) (9,68.86)};
\addplot[gray, dashed, line width=0.5pt, forget plot] coordinates {(0,50) (9,50)};
\end{axis}
\end{tikzpicture}%
\\[-1pt]
            {\scriptsize \# teaching pairs}
            \vspace{-5pt}
        \caption{\small
            Average accuracy of every model as a function of the number of examples, with correct and inverted labels.
        }
        \label{fig:icl_flipped_curves}
        \end{figure}

        In particular, we teach models to distinguish dropout from noise by naming the perturbations with control labels (for instance, $\langle\texttt{X};\texttt{Y}\rangle$), repeat the experiment with such aliases flipped (for instance, $\langle\texttt{Y};\texttt{X}\rangle$), and, for every value of $p$ and $\sigma$, report the difference in accuracy $\delta_{\langle p;\sigma\rangle}$.
        Fig.~\ref{fig:icl_deltas_qwen32} reports this difference for \model{qwen32} and some control labels.
        With the correct labels (top left) and their synonyms (top right), $\delta_{\langle p;\sigma\rangle}$ is positive for most pairs pairs. By contrast, with $\langle\texttt{X};\texttt{Y}\rangle$, for instance, not only $\delta_{\langle p;\sigma\rangle}$ magnitude is much smaller, but its sign depends on which perturbation is stronger: it is positive when noise is strong and dropout is weak, and negative in the opposite case. This corroborates that \model{qwen32} indeed learns better only with the correct labels.
        Surprisingly, \model{gemma31} shows a similar (however much weaker) pattern (see \S\ref{subsec:appendix_delta_heatmaps} for the all controls and models), raising the possibility that in-context learning exposes an underlying prior towards the correct labels too weak to manifest zero-shot.

        \begin{figure}[H]
        \centering
            \begin{minipage}[b]{0.44\textwidth}
                \centering
                \tikzset{external/export next=false}
                \input{plots/07_few_shots/delta_qwen32_dropout_noise}
            \end{minipage}%
            \hspace{0.02\textwidth}%
            \begin{minipage}[b]{0.50\textwidth}
                \centering
                \tikzset{external/export next=false}
                \centering
\begin{tikzpicture}
\begin{axis}[
    scale only axis,
    width=3.3cm,
    height=3.3cm,
    xlabel={},
    ylabel={},
    yticklabels={},
    colorbar,
    colormap={pastel_po}{rgb255(0cm)=(225,160,100); rgb255(50cm)=(255,255,255); rgb255(100cm)=(160,135,200)},
    point meta min=-25.0,
    point meta max=25.0,
    xtick={0,1,2,3,4,5,6,7,8,9,10},
    xticklabels={$p_{\min}$,,,,,,,,,,$p_{\max}$},
    ytick={0,1,2,3,4,5,6,7,8,9,10},
    x tick label style={font=\scriptsize},
    y tick label style={font=\scriptsize},
    view={0}{90},
    axis line style={draw=none},
    major tick length=0pt,
    enlargelimits=false,
    colorbar style={font=\tiny},
    title style={font=\small},
    title={Masking / Jitter},
    title style={yshift=-1.0ex},
]
\addplot[matrix plot*, mesh/cols=11, mesh/rows=11, point meta=explicit] table[meta=C] {
x y C
0 10 -0.70
1 10 -2.40
2 10 -0.70
3 10 -2.70
4 10 1.40
5 10 -1.20
6 10 -3.30
7 10 3.50
8 10 11.60
9 10 9.70
10 10 9.00
0 9 0.20
1 9 2.60
2 9 4.70
3 9 5.00
4 9 0.40
5 9 -1.80
6 9 1.10
7 9 8.80
8 9 11.00
9 9 10.10
10 9 22.80
0 8 3.80
1 8 4.50
2 8 6.30
3 8 0.50
4 8 2.90
5 8 5.40
6 8 5.80
7 8 8.50
8 8 13.40
9 8 19.90
10 8 21.60
0 7 6.80
1 7 8.00
2 7 6.20
3 7 0.80
4 7 5.40
5 7 4.90
6 7 6.00
7 7 11.80
8 7 13.70
9 7 19.50
10 7 20.40
0 6 1.00
1 6 4.80
2 6 7.40
3 6 6.10
4 6 8.10
5 6 9.50
6 6 8.90
7 6 13.10
8 6 15.60
9 6 20.10
10 6 17.10
0 5 8.20
1 5 6.30
2 5 3.40
3 5 2.40
4 5 4.60
5 5 8.70
6 5 11.00
7 5 7.20
8 5 16.00
9 5 18.90
10 5 20.20
0 4 4.20
1 4 5.70
2 4 0.90
3 4 1.50
4 4 4.30
5 4 7.20
6 4 5.60
7 4 11.80
8 4 14.90
9 4 16.50
10 4 21.70
0 3 5.10
1 3 7.00
2 3 6.40
3 3 2.80
4 3 0.50
5 3 3.80
6 3 5.20
7 3 11.60
8 3 11.40
9 3 8.80
10 3 12.70
0 2 0.20
1 2 4.30
2 2 2.50
3 2 -2.20
4 2 3.80
5 2 2.80
6 2 2.90
7 2 5.70
8 2 9.90
9 2 14.10
10 2 7.70
0 1 0.50
1 1 -0.90
2 1 3.50
3 1 1.40
4 1 2.00
5 1 2.40
6 1 3.80
7 1 7.90
8 1 1.90
9 1 11.40
10 1 11.70
0 0 1.90
1 0 0.10
2 0 -2.10
3 0 1.10
4 0 3.30
5 0 -0.30
6 0 0.80
7 0 7.30
8 0 5.10
9 0 2.10
10 0 12.40
};
\node[font=\fontsize{4}{4.8}\selectfont] at (axis cs:0,0) {1.9};
\node[font=\fontsize{4}{4.8}\selectfont] at (axis cs:1,0) {0.1};
\node[font=\fontsize{4}{4.8}\selectfont] at (axis cs:2,0) {-2.1};
\node[font=\fontsize{4}{4.8}\selectfont] at (axis cs:3,0) {1.1};
\node[font=\fontsize{4}{4.8}\selectfont] at (axis cs:4,0) {3.3};
\node[font=\fontsize{4}{4.8}\selectfont] at (axis cs:5,0) {-0.3};
\node[font=\fontsize{4}{4.8}\selectfont] at (axis cs:6,0) {0.8};
\node[font=\fontsize{4}{4.8}\selectfont] at (axis cs:7,0) {7.3};
\node[font=\fontsize{4}{4.8}\selectfont] at (axis cs:8,0) {5.1};
\node[font=\fontsize{4}{4.8}\selectfont] at (axis cs:9,0) {2.1};
\node[font=\fontsize{4}{4.8}\selectfont] at (axis cs:10,0) {12.4};
\node[font=\fontsize{4}{4.8}\selectfont] at (axis cs:0,1) {0.5};
\node[font=\fontsize{4}{4.8}\selectfont] at (axis cs:1,1) {-0.9};
\node[font=\fontsize{4}{4.8}\selectfont] at (axis cs:2,1) {3.5};
\node[font=\fontsize{4}{4.8}\selectfont] at (axis cs:3,1) {1.4};
\node[font=\fontsize{4}{4.8}\selectfont] at (axis cs:4,1) {2.0};
\node[font=\fontsize{4}{4.8}\selectfont] at (axis cs:5,1) {2.4};
\node[font=\fontsize{4}{4.8}\selectfont] at (axis cs:6,1) {3.8};
\node[font=\fontsize{4}{4.8}\selectfont] at (axis cs:7,1) {7.9};
\node[font=\fontsize{4}{4.8}\selectfont] at (axis cs:8,1) {1.9};
\node[font=\fontsize{4}{4.8}\selectfont] at (axis cs:9,1) {11.4};
\node[font=\fontsize{4}{4.8}\selectfont] at (axis cs:10,1) {11.7};
\node[font=\fontsize{4}{4.8}\selectfont] at (axis cs:0,2) {0.2};
\node[font=\fontsize{4}{4.8}\selectfont] at (axis cs:1,2) {4.3};
\node[font=\fontsize{4}{4.8}\selectfont] at (axis cs:2,2) {2.5};
\node[font=\fontsize{4}{4.8}\selectfont] at (axis cs:3,2) {-2.2};
\node[font=\fontsize{4}{4.8}\selectfont] at (axis cs:4,2) {3.8};
\node[font=\fontsize{4}{4.8}\selectfont] at (axis cs:5,2) {2.8};
\node[font=\fontsize{4}{4.8}\selectfont] at (axis cs:6,2) {2.9};
\node[font=\fontsize{4}{4.8}\selectfont] at (axis cs:7,2) {5.7};
\node[font=\fontsize{4}{4.8}\selectfont] at (axis cs:8,2) {9.9};
\node[font=\fontsize{4}{4.8}\selectfont] at (axis cs:9,2) {14.1};
\node[font=\fontsize{4}{4.8}\selectfont] at (axis cs:10,2) {7.7};
\node[font=\fontsize{4}{4.8}\selectfont] at (axis cs:0,3) {5.1};
\node[font=\fontsize{4}{4.8}\selectfont] at (axis cs:1,3) {7.0};
\node[font=\fontsize{4}{4.8}\selectfont] at (axis cs:2,3) {6.4};
\node[font=\fontsize{4}{4.8}\selectfont] at (axis cs:3,3) {2.8};
\node[font=\fontsize{4}{4.8}\selectfont] at (axis cs:4,3) {0.5};
\node[font=\fontsize{4}{4.8}\selectfont] at (axis cs:5,3) {3.8};
\node[font=\fontsize{4}{4.8}\selectfont] at (axis cs:6,3) {5.2};
\node[font=\fontsize{4}{4.8}\selectfont] at (axis cs:7,3) {11.6};
\node[font=\fontsize{4}{4.8}\selectfont] at (axis cs:8,3) {11.4};
\node[font=\fontsize{4}{4.8}\selectfont] at (axis cs:9,3) {8.8};
\node[font=\fontsize{4}{4.8}\selectfont] at (axis cs:10,3) {12.7};
\node[font=\fontsize{4}{4.8}\selectfont] at (axis cs:0,4) {4.2};
\node[font=\fontsize{4}{4.8}\selectfont] at (axis cs:1,4) {5.7};
\node[font=\fontsize{4}{4.8}\selectfont] at (axis cs:2,4) {0.9};
\node[font=\fontsize{4}{4.8}\selectfont] at (axis cs:3,4) {1.5};
\node[font=\fontsize{4}{4.8}\selectfont] at (axis cs:4,4) {4.3};
\node[font=\fontsize{4}{4.8}\selectfont] at (axis cs:5,4) {7.2};
\node[font=\fontsize{4}{4.8}\selectfont] at (axis cs:6,4) {5.6};
\node[font=\fontsize{4}{4.8}\selectfont] at (axis cs:7,4) {11.8};
\node[font=\fontsize{4}{4.8}\selectfont] at (axis cs:8,4) {14.9};
\node[font=\fontsize{4}{4.8}\selectfont] at (axis cs:9,4) {16.5};
\node[font=\fontsize{4}{4.8}\selectfont] at (axis cs:10,4) {21.7};
\node[font=\fontsize{4}{4.8}\selectfont] at (axis cs:0,5) {8.2};
\node[font=\fontsize{4}{4.8}\selectfont] at (axis cs:1,5) {6.3};
\node[font=\fontsize{4}{4.8}\selectfont] at (axis cs:2,5) {3.4};
\node[font=\fontsize{4}{4.8}\selectfont] at (axis cs:3,5) {2.4};
\node[font=\fontsize{4}{4.8}\selectfont] at (axis cs:4,5) {4.6};
\node[font=\fontsize{4}{4.8}\selectfont] at (axis cs:5,5) {8.7};
\node[font=\fontsize{4}{4.8}\selectfont] at (axis cs:6,5) {11.0};
\node[font=\fontsize{4}{4.8}\selectfont] at (axis cs:7,5) {7.2};
\node[font=\fontsize{4}{4.8}\selectfont] at (axis cs:8,5) {16.0};
\node[font=\fontsize{4}{4.8}\selectfont] at (axis cs:9,5) {18.9};
\node[font=\fontsize{4}{4.8}\selectfont] at (axis cs:10,5) {20.2};
\node[font=\fontsize{4}{4.8}\selectfont] at (axis cs:0,6) {1.0};
\node[font=\fontsize{4}{4.8}\selectfont] at (axis cs:1,6) {4.8};
\node[font=\fontsize{4}{4.8}\selectfont] at (axis cs:2,6) {7.4};
\node[font=\fontsize{4}{4.8}\selectfont] at (axis cs:3,6) {6.1};
\node[font=\fontsize{4}{4.8}\selectfont] at (axis cs:4,6) {8.1};
\node[font=\fontsize{4}{4.8}\selectfont] at (axis cs:5,6) {9.5};
\node[font=\fontsize{4}{4.8}\selectfont] at (axis cs:6,6) {8.9};
\node[font=\fontsize{4}{4.8}\selectfont] at (axis cs:7,6) {13.1};
\node[font=\fontsize{4}{4.8}\selectfont] at (axis cs:8,6) {15.6};
\node[font=\fontsize{4}{4.8}\selectfont] at (axis cs:9,6) {20.1};
\node[font=\fontsize{4}{4.8}\selectfont] at (axis cs:10,6) {17.1};
\node[font=\fontsize{4}{4.8}\selectfont] at (axis cs:0,7) {6.8};
\node[font=\fontsize{4}{4.8}\selectfont] at (axis cs:1,7) {8.0};
\node[font=\fontsize{4}{4.8}\selectfont] at (axis cs:2,7) {6.2};
\node[font=\fontsize{4}{4.8}\selectfont] at (axis cs:3,7) {0.8};
\node[font=\fontsize{4}{4.8}\selectfont] at (axis cs:4,7) {5.4};
\node[font=\fontsize{4}{4.8}\selectfont] at (axis cs:5,7) {4.9};
\node[font=\fontsize{4}{4.8}\selectfont] at (axis cs:6,7) {6.0};
\node[font=\fontsize{4}{4.8}\selectfont] at (axis cs:7,7) {11.8};
\node[font=\fontsize{4}{4.8}\selectfont] at (axis cs:8,7) {13.7};
\node[font=\fontsize{4}{4.8}\selectfont] at (axis cs:9,7) {19.5};
\node[font=\fontsize{4}{4.8}\selectfont] at (axis cs:10,7) {20.4};
\node[font=\fontsize{4}{4.8}\selectfont] at (axis cs:0,8) {3.8};
\node[font=\fontsize{4}{4.8}\selectfont] at (axis cs:1,8) {4.5};
\node[font=\fontsize{4}{4.8}\selectfont] at (axis cs:2,8) {6.3};
\node[font=\fontsize{4}{4.8}\selectfont] at (axis cs:3,8) {0.5};
\node[font=\fontsize{4}{4.8}\selectfont] at (axis cs:4,8) {2.9};
\node[font=\fontsize{4}{4.8}\selectfont] at (axis cs:5,8) {5.4};
\node[font=\fontsize{4}{4.8}\selectfont] at (axis cs:6,8) {5.8};
\node[font=\fontsize{4}{4.8}\selectfont] at (axis cs:7,8) {8.5};
\node[font=\fontsize{4}{4.8}\selectfont] at (axis cs:8,8) {13.4};
\node[font=\fontsize{4}{4.8}\selectfont] at (axis cs:9,8) {19.9};
\node[font=\fontsize{4}{4.8}\selectfont] at (axis cs:10,8) {21.6};
\node[font=\fontsize{4}{4.8}\selectfont] at (axis cs:0,9) {0.2};
\node[font=\fontsize{4}{4.8}\selectfont] at (axis cs:1,9) {2.6};
\node[font=\fontsize{4}{4.8}\selectfont] at (axis cs:2,9) {4.7};
\node[font=\fontsize{4}{4.8}\selectfont] at (axis cs:3,9) {5.0};
\node[font=\fontsize{4}{4.8}\selectfont] at (axis cs:4,9) {0.4};
\node[font=\fontsize{4}{4.8}\selectfont] at (axis cs:5,9) {-1.8};
\node[font=\fontsize{4}{4.8}\selectfont] at (axis cs:6,9) {1.1};
\node[font=\fontsize{4}{4.8}\selectfont] at (axis cs:7,9) {8.8};
\node[font=\fontsize{4}{4.8}\selectfont] at (axis cs:8,9) {11.0};
\node[font=\fontsize{4}{4.8}\selectfont] at (axis cs:9,9) {10.1};
\node[font=\fontsize{4}{4.8}\selectfont] at (axis cs:10,9) {22.8};
\node[font=\fontsize{4}{4.8}\selectfont] at (axis cs:0,10) {-0.7};
\node[font=\fontsize{4}{4.8}\selectfont] at (axis cs:1,10) {-2.4};
\node[font=\fontsize{4}{4.8}\selectfont] at (axis cs:2,10) {-0.7};
\node[font=\fontsize{4}{4.8}\selectfont] at (axis cs:3,10) {-2.7};
\node[font=\fontsize{4}{4.8}\selectfont] at (axis cs:4,10) {1.4};
\node[font=\fontsize{4}{4.8}\selectfont] at (axis cs:5,10) {-1.2};
\node[font=\fontsize{4}{4.8}\selectfont] at (axis cs:6,10) {-3.3};
\node[font=\fontsize{4}{4.8}\selectfont] at (axis cs:7,10) {3.5};
\node[font=\fontsize{4}{4.8}\selectfont] at (axis cs:8,10) {11.6};
\node[font=\fontsize{4}{4.8}\selectfont] at (axis cs:9,10) {9.7};
\node[font=\fontsize{4}{4.8}\selectfont] at (axis cs:10,10) {9.0};
\end{axis}
\end{tikzpicture}
            \end{minipage}\\[4pt]
            \begin{minipage}[b]{0.44\textwidth}
                \centering
                \tikzset{external/export next=false}
                \input{plots/07_few_shots/delta_qwen32_x_y}
            \end{minipage}%
            \hspace{0.02\textwidth}%
            \begin{minipage}[b]{0.50\textwidth}
                \centering
                \tikzset{external/export next=false}
                \centering
\begin{tikzpicture}
\begin{axis}[
    scale only axis,
    width=3.3cm,
    height=3.3cm,
    xlabel={\small Dropout rate $p$},
    ylabel={},
    yticklabels={},
    colorbar,
    colormap={pastel_po}{rgb255(0cm)=(225,160,100); rgb255(50cm)=(255,255,255); rgb255(100cm)=(160,135,200)},
    point meta min=-25.0,
    point meta max=25.0,
    xtick={0,1,2,3,4,5,6,7,8,9,10},
    xticklabels={$p_{\min}$,,,,,,,,,,$p_{\max}$},
    ytick={0,1,2,3,4,5,6,7,8,9,10},
    x tick label style={font=\scriptsize},
    y tick label style={font=\scriptsize},
    view={0}{90},
    axis line style={draw=none},
    major tick length=0pt,
    enlargelimits=false,
    colorbar style={font=\tiny},
    title style={font=\small},
    title={Rotation / Permutation},
    title style={yshift=-1.0ex},
]
\addplot[matrix plot*, mesh/cols=11, mesh/rows=11, point meta=explicit] table[meta=C] {
x y C
0 10 0.90
1 10 3.00
2 10 3.70
3 10 0.80
4 10 5.40
5 10 5.10
6 10 3.30
7 10 4.20
8 10 4.10
9 10 6.20
10 10 8.00
0 9 3.20
1 9 1.90
2 9 2.40
3 9 4.20
4 9 3.70
5 9 4.10
6 9 1.20
7 9 5.70
8 9 4.50
9 9 6.10
10 9 6.20
0 8 6.60
1 8 2.90
2 8 1.50
3 8 4.40
4 8 3.80
5 8 2.90
6 8 4.80
7 8 6.80
8 8 3.60
9 8 4.40
10 8 5.00
0 7 3.30
1 7 4.00
2 7 3.10
3 7 3.10
4 7 4.60
5 7 4.20
6 7 3.20
7 7 0.90
8 7 3.30
9 7 6.50
10 7 7.30
0 6 1.60
1 6 4.00
2 6 1.60
3 6 3.20
4 6 1.80
5 6 5.10
6 6 2.00
7 6 1.70
8 6 1.60
9 6 3.20
10 6 7.30
0 5 0.20
1 5 -0.30
2 5 3.80
3 5 4.70
4 5 0.80
5 5 2.20
6 5 0.70
7 5 1.70
8 5 4.80
9 5 5.20
10 5 4.40
0 4 1.90
1 4 1.90
2 4 2.20
3 4 2.00
4 4 0.90
5 4 3.80
6 4 0.20
7 4 5.20
8 4 0.40
9 4 3.10
10 4 3.70
0 3 1.30
1 3 1.40
2 3 0.80
3 3 1.00
4 3 1.20
5 3 0.90
6 3 3.50
7 3 0.00
8 3 0.90
9 3 -0.30
10 3 2.60
0 2 0.80
1 2 -1.50
2 2 1.60
3 2 -1.30
4 2 0.60
5 2 2.30
6 2 1.50
7 2 -0.70
8 2 1.00
9 2 1.90
10 2 1.40
0 1 -1.00
1 1 0.00
2 1 -0.70
3 1 0.30
4 1 -0.70
5 1 -1.00
6 1 2.20
7 1 2.50
8 1 2.10
9 1 0.80
10 1 -2.50
0 0 -1.70
1 0 -3.70
2 0 -3.90
3 0 -1.10
4 0 0.20
5 0 -0.60
6 0 -2.30
7 0 -0.30
8 0 -0.40
9 0 -0.50
10 0 0.40
};
\node[font=\fontsize{4}{4.8}\selectfont] at (axis cs:0,0) {-1.7};
\node[font=\fontsize{4}{4.8}\selectfont] at (axis cs:1,0) {-3.7};
\node[font=\fontsize{4}{4.8}\selectfont] at (axis cs:2,0) {-3.9};
\node[font=\fontsize{4}{4.8}\selectfont] at (axis cs:3,0) {-1.1};
\node[font=\fontsize{4}{4.8}\selectfont] at (axis cs:4,0) {0.2};
\node[font=\fontsize{4}{4.8}\selectfont] at (axis cs:5,0) {-0.6};
\node[font=\fontsize{4}{4.8}\selectfont] at (axis cs:6,0) {-2.3};
\node[font=\fontsize{4}{4.8}\selectfont] at (axis cs:7,0) {-0.3};
\node[font=\fontsize{4}{4.8}\selectfont] at (axis cs:8,0) {-0.4};
\node[font=\fontsize{4}{4.8}\selectfont] at (axis cs:9,0) {-0.5};
\node[font=\fontsize{4}{4.8}\selectfont] at (axis cs:10,0) {0.4};
\node[font=\fontsize{4}{4.8}\selectfont] at (axis cs:0,1) {-1.0};
\node[font=\fontsize{4}{4.8}\selectfont] at (axis cs:1,1) {0.0};
\node[font=\fontsize{4}{4.8}\selectfont] at (axis cs:2,1) {-0.7};
\node[font=\fontsize{4}{4.8}\selectfont] at (axis cs:3,1) {0.3};
\node[font=\fontsize{4}{4.8}\selectfont] at (axis cs:4,1) {-0.7};
\node[font=\fontsize{4}{4.8}\selectfont] at (axis cs:5,1) {-1.0};
\node[font=\fontsize{4}{4.8}\selectfont] at (axis cs:6,1) {2.2};
\node[font=\fontsize{4}{4.8}\selectfont] at (axis cs:7,1) {2.5};
\node[font=\fontsize{4}{4.8}\selectfont] at (axis cs:8,1) {2.1};
\node[font=\fontsize{4}{4.8}\selectfont] at (axis cs:9,1) {0.8};
\node[font=\fontsize{4}{4.8}\selectfont] at (axis cs:10,1) {-2.5};
\node[font=\fontsize{4}{4.8}\selectfont] at (axis cs:0,2) {0.8};
\node[font=\fontsize{4}{4.8}\selectfont] at (axis cs:1,2) {-1.5};
\node[font=\fontsize{4}{4.8}\selectfont] at (axis cs:2,2) {1.6};
\node[font=\fontsize{4}{4.8}\selectfont] at (axis cs:3,2) {-1.3};
\node[font=\fontsize{4}{4.8}\selectfont] at (axis cs:4,2) {0.6};
\node[font=\fontsize{4}{4.8}\selectfont] at (axis cs:5,2) {2.3};
\node[font=\fontsize{4}{4.8}\selectfont] at (axis cs:6,2) {1.5};
\node[font=\fontsize{4}{4.8}\selectfont] at (axis cs:7,2) {-0.7};
\node[font=\fontsize{4}{4.8}\selectfont] at (axis cs:8,2) {1.0};
\node[font=\fontsize{4}{4.8}\selectfont] at (axis cs:9,2) {1.9};
\node[font=\fontsize{4}{4.8}\selectfont] at (axis cs:10,2) {1.4};
\node[font=\fontsize{4}{4.8}\selectfont] at (axis cs:0,3) {1.3};
\node[font=\fontsize{4}{4.8}\selectfont] at (axis cs:1,3) {1.4};
\node[font=\fontsize{4}{4.8}\selectfont] at (axis cs:2,3) {0.8};
\node[font=\fontsize{4}{4.8}\selectfont] at (axis cs:3,3) {1.0};
\node[font=\fontsize{4}{4.8}\selectfont] at (axis cs:4,3) {1.2};
\node[font=\fontsize{4}{4.8}\selectfont] at (axis cs:5,3) {0.9};
\node[font=\fontsize{4}{4.8}\selectfont] at (axis cs:6,3) {3.5};
\node[font=\fontsize{4}{4.8}\selectfont] at (axis cs:7,3) {0.0};
\node[font=\fontsize{4}{4.8}\selectfont] at (axis cs:8,3) {0.9};
\node[font=\fontsize{4}{4.8}\selectfont] at (axis cs:9,3) {-0.3};
\node[font=\fontsize{4}{4.8}\selectfont] at (axis cs:10,3) {2.6};
\node[font=\fontsize{4}{4.8}\selectfont] at (axis cs:0,4) {1.9};
\node[font=\fontsize{4}{4.8}\selectfont] at (axis cs:1,4) {1.9};
\node[font=\fontsize{4}{4.8}\selectfont] at (axis cs:2,4) {2.2};
\node[font=\fontsize{4}{4.8}\selectfont] at (axis cs:3,4) {2.0};
\node[font=\fontsize{4}{4.8}\selectfont] at (axis cs:4,4) {0.9};
\node[font=\fontsize{4}{4.8}\selectfont] at (axis cs:5,4) {3.8};
\node[font=\fontsize{4}{4.8}\selectfont] at (axis cs:6,4) {0.2};
\node[font=\fontsize{4}{4.8}\selectfont] at (axis cs:7,4) {5.2};
\node[font=\fontsize{4}{4.8}\selectfont] at (axis cs:8,4) {0.4};
\node[font=\fontsize{4}{4.8}\selectfont] at (axis cs:9,4) {3.1};
\node[font=\fontsize{4}{4.8}\selectfont] at (axis cs:10,4) {3.7};
\node[font=\fontsize{4}{4.8}\selectfont] at (axis cs:0,5) {0.2};
\node[font=\fontsize{4}{4.8}\selectfont] at (axis cs:1,5) {-0.3};
\node[font=\fontsize{4}{4.8}\selectfont] at (axis cs:2,5) {3.8};
\node[font=\fontsize{4}{4.8}\selectfont] at (axis cs:3,5) {4.7};
\node[font=\fontsize{4}{4.8}\selectfont] at (axis cs:4,5) {0.8};
\node[font=\fontsize{4}{4.8}\selectfont] at (axis cs:5,5) {2.2};
\node[font=\fontsize{4}{4.8}\selectfont] at (axis cs:6,5) {0.7};
\node[font=\fontsize{4}{4.8}\selectfont] at (axis cs:7,5) {1.7};
\node[font=\fontsize{4}{4.8}\selectfont] at (axis cs:8,5) {4.8};
\node[font=\fontsize{4}{4.8}\selectfont] at (axis cs:9,5) {5.2};
\node[font=\fontsize{4}{4.8}\selectfont] at (axis cs:10,5) {4.4};
\node[font=\fontsize{4}{4.8}\selectfont] at (axis cs:0,6) {1.6};
\node[font=\fontsize{4}{4.8}\selectfont] at (axis cs:1,6) {4.0};
\node[font=\fontsize{4}{4.8}\selectfont] at (axis cs:2,6) {1.6};
\node[font=\fontsize{4}{4.8}\selectfont] at (axis cs:3,6) {3.2};
\node[font=\fontsize{4}{4.8}\selectfont] at (axis cs:4,6) {1.8};
\node[font=\fontsize{4}{4.8}\selectfont] at (axis cs:5,6) {5.1};
\node[font=\fontsize{4}{4.8}\selectfont] at (axis cs:6,6) {2.0};
\node[font=\fontsize{4}{4.8}\selectfont] at (axis cs:7,6) {1.7};
\node[font=\fontsize{4}{4.8}\selectfont] at (axis cs:8,6) {1.6};
\node[font=\fontsize{4}{4.8}\selectfont] at (axis cs:9,6) {3.2};
\node[font=\fontsize{4}{4.8}\selectfont] at (axis cs:10,6) {7.3};
\node[font=\fontsize{4}{4.8}\selectfont] at (axis cs:0,7) {3.3};
\node[font=\fontsize{4}{4.8}\selectfont] at (axis cs:1,7) {4.0};
\node[font=\fontsize{4}{4.8}\selectfont] at (axis cs:2,7) {3.1};
\node[font=\fontsize{4}{4.8}\selectfont] at (axis cs:3,7) {3.1};
\node[font=\fontsize{4}{4.8}\selectfont] at (axis cs:4,7) {4.6};
\node[font=\fontsize{4}{4.8}\selectfont] at (axis cs:5,7) {4.2};
\node[font=\fontsize{4}{4.8}\selectfont] at (axis cs:6,7) {3.2};
\node[font=\fontsize{4}{4.8}\selectfont] at (axis cs:7,7) {0.9};
\node[font=\fontsize{4}{4.8}\selectfont] at (axis cs:8,7) {3.3};
\node[font=\fontsize{4}{4.8}\selectfont] at (axis cs:9,7) {6.5};
\node[font=\fontsize{4}{4.8}\selectfont] at (axis cs:10,7) {7.3};
\node[font=\fontsize{4}{4.8}\selectfont] at (axis cs:0,8) {6.6};
\node[font=\fontsize{4}{4.8}\selectfont] at (axis cs:1,8) {2.9};
\node[font=\fontsize{4}{4.8}\selectfont] at (axis cs:2,8) {1.5};
\node[font=\fontsize{4}{4.8}\selectfont] at (axis cs:3,8) {4.4};
\node[font=\fontsize{4}{4.8}\selectfont] at (axis cs:4,8) {3.8};
\node[font=\fontsize{4}{4.8}\selectfont] at (axis cs:5,8) {2.9};
\node[font=\fontsize{4}{4.8}\selectfont] at (axis cs:6,8) {4.8};
\node[font=\fontsize{4}{4.8}\selectfont] at (axis cs:7,8) {6.8};
\node[font=\fontsize{4}{4.8}\selectfont] at (axis cs:8,8) {3.6};
\node[font=\fontsize{4}{4.8}\selectfont] at (axis cs:9,8) {4.4};
\node[font=\fontsize{4}{4.8}\selectfont] at (axis cs:10,8) {5.0};
\node[font=\fontsize{4}{4.8}\selectfont] at (axis cs:0,9) {3.2};
\node[font=\fontsize{4}{4.8}\selectfont] at (axis cs:1,9) {1.9};
\node[font=\fontsize{4}{4.8}\selectfont] at (axis cs:2,9) {2.4};
\node[font=\fontsize{4}{4.8}\selectfont] at (axis cs:3,9) {4.2};
\node[font=\fontsize{4}{4.8}\selectfont] at (axis cs:4,9) {3.7};
\node[font=\fontsize{4}{4.8}\selectfont] at (axis cs:5,9) {4.1};
\node[font=\fontsize{4}{4.8}\selectfont] at (axis cs:6,9) {1.2};
\node[font=\fontsize{4}{4.8}\selectfont] at (axis cs:7,9) {5.7};
\node[font=\fontsize{4}{4.8}\selectfont] at (axis cs:8,9) {4.5};
\node[font=\fontsize{4}{4.8}\selectfont] at (axis cs:9,9) {6.1};
\node[font=\fontsize{4}{4.8}\selectfont] at (axis cs:10,9) {6.2};
\node[font=\fontsize{4}{4.8}\selectfont] at (axis cs:0,10) {0.9};
\node[font=\fontsize{4}{4.8}\selectfont] at (axis cs:1,10) {3.0};
\node[font=\fontsize{4}{4.8}\selectfont] at (axis cs:2,10) {3.7};
\node[font=\fontsize{4}{4.8}\selectfont] at (axis cs:3,10) {0.8};
\node[font=\fontsize{4}{4.8}\selectfont] at (axis cs:4,10) {5.4};
\node[font=\fontsize{4}{4.8}\selectfont] at (axis cs:5,10) {5.1};
\node[font=\fontsize{4}{4.8}\selectfont] at (axis cs:6,10) {3.3};
\node[font=\fontsize{4}{4.8}\selectfont] at (axis cs:7,10) {4.2};
\node[font=\fontsize{4}{4.8}\selectfont] at (axis cs:8,10) {4.1};
\node[font=\fontsize{4}{4.8}\selectfont] at (axis cs:9,10) {6.2};
\node[font=\fontsize{4}{4.8}\selectfont] at (axis cs:10,10) {8.0};
\end{axis}
\end{tikzpicture}
            \end{minipage}
            \vspace{-5pt}
        \caption{\small
            Expansion of what has been averaged in Fig.~\ref{fig:icl_flipped_curves} for \model{qwen32} and one in-context example: the difference between the accuracy with correct and with flipped labels, for all pairs $\langle p,\sigma\rangle$, with the labels $\langle\texttt{Dropout};\texttt{Noise}\rangle$ (\textbf{top left}), their synonyms $\langle\texttt{Masking};\texttt{Jitter}\rangle$ (\textbf{top right}), and the control labels $\langle\texttt{X};\texttt{Y}\rangle$ (\textbf{bottom left}) and $\langle\texttt{Rotation};\texttt{Permutation}\rangle$ (\textbf{bottom right}).
        }
        \label{fig:icl_deltas_qwen32}
        \end{figure}

    \FloatBarrier
\section{Conclusion}\label{sec:conclusion}

    We showed that language models can localize dropout and Gaussian noise in their activations (\S\ref{subsec:loc_results}).
    Most of them also report a perturbation applied to a single sentence (\S\ref{subsec:detection_results}); \model{olmo32}, which rarely does, becomes more reluctant to answer `Yes' after DPO, without losing its sensitivity (\S\ref{subsec:detection_olmo_stages}).
    When asked (zero-shot) to identify which perturbation they underwent (\S\ref{sec:zero_shot}), \model{qwen32} and \model{qwen27} performed better as the magnitude of the perturbations increased. Other models, such as \model{olmo32}, did not.
    Notwithstanding the difficulty of such a task, all models but one also \emph{learned} to distinguish dropout from noise when supervised with in-context examples (\S\ref{subsec:icl_results}, \S\ref{subsec:appendix_few_shot_heatmaps}). In this setting, \model{qwen32} and \model{qwen27} again reveal a prior for the correct teaching-labels (\S\ref{subsec:icl_controls}). 
    We remark that the perturbations we employed are stochastic and not aligned with semantically meaningful vectors:
    that models can distinguish them suggests they can access and verbalize a wider signal from their activations than previously established.

    Let us emphasize a key point: it should be straightforward to distinguish dropout from noise with a simple probe on the activations. The mystery here is not how the signal can exist, but rather how a model is able to verbalize the difference and connect it to its semantic knowledge.
    That is, how does semantic understanding of the concepts of dropout and Gaussian noise, instilled by training, end up aligned with the models' (first) direct ``experience'' of having these perturbations applied to their activations?
    In search of answers, we have tried our best to perform many ablations (\cf. \S\ref{sec:methodology}, \S\ref{subsec:appendix_dropout},
    \S\ref{subsec:appendix_prompt_ablation}, \S\ref{subsec:appendix_loc_length}, \S\ref{subsec:appendix_layer_ablation}) and controls
    (\cf.
    \S\ref{subsec:detection_controls},
    \S\ref{subsec:zero_shot_controls},
    \S\ref{subsec:icl_controls},
    \S\ref{subsec:appendix_loc_controls},  \S\ref{subsec:appendix_zero_shot_controls},  \S\ref{subsec:appendix_delta_heatmaps}), and exclude the most likely mechanisms for it (\cf. \S\ref{subsec:loc_controls}, \S\ref{subsec:loc_emergence}, \S\ref{subsec:detection_olmo_stages}, \S\ref{subsec:icl_results}, \S\ref{subsec:appendix_logprobs}, \S\ref{subsec:appendix_olmo_checkpoints}, \S\ref{subsec:appendix_zero_shot_entropy}, \S\ref{subsec:appendix_few_shot_heatmaps}). Still, a complete picture remains elusive.

    \textbf{Future work.}
    To answer this question, it may be useful to first collect more observations. 
    What happens if we allow the models to first generate reasoning tokens?
    What about other interventions, such as adding uniform noise in $[0,1]$ or applying quantization?
    We are eager to further explore the potential to connect in-context demonstrations with models' internals.

    A key question is whether the ability to recognize dropout and noise is more pronounced for models that have undergone a training stage that actually involves each of these perturbations.
    Observe that there is a direct incentive for such a model to behave differently during training: if the expected answer differs from the one the model believes is best, then at inference it should give the latter. By contrast, failing to give the expected answer during training would result in a damaging update.
    Therefore, beyond handling issues arising from ``evaluation awareness'' \citep[p. 10]{bengio2026international}, ensuring that a model cannot distinguish training from inference would be one further necessary step to establish trust in the assessment of a model's behavior.
    \section*{Acknowledgments}\label{sec:acknowledgments}
The authors are grateful to Cameron Berg and Eric Elmoznino, who suggested to investigate detection \emph{vs.} localization across \model{olmo32} training stages. 
The authors are thankful for early comments and feedback of:
Philippe Beaudoin,
Dmitri Carpov,
Raffaello Fornasiere,
Josh Engels,
Ga\"el Gendron,
Joumana Ghosn,
Pietro Greiner,
Moksh Jain,
Zachary Kenton,
Varsha Kishore,
Victoria Krakovna,
Matt MacDermott,
Vincent Mai,
Ian McKenzie,
Adam Oberman,
Jonathan Richens,
Roberta Rocca,
Marc-Antoine Rondeau,
Charbel-Rapha\"el Segerie,
Iulian Serban,
Winnie Street,
Esmeralda S. Whitammer.

    \bibliographystyle{plainnat}
    \bibliography{references}

\appendix

\makeatletter
\setlength{\@fptop}{0pt}
\setlength{\@fpsep}{\floatsep}
\setlength{\@fpbot}{0pt plus 1fil}
\makeatother
\section{Appendix}
\subsection{Models}\label{subsec:models}
    The exact model names we experimented with are: 
    \begin{verbatim}
    gemma-4-31B-it, 
    Llama-3.1-8B-Instruct,
    Mistral-Small-3.2-24B-Instruct-2506,
    Olmo-3.1-32B-Instruct,
    Qwen3-14B,
    Qwen3-32B,
    Qwen3.8-27B,
    Seed-OSS-36B-Instruct.
    \end{verbatim}

\subsection{Prompts}\label{subsec:appendix_prompts}

    For each experiment, there are (at least) $20$ variants of the prompt formatting. Here, we report an example of a prompt for each experiment.   Refer to Fig.~\ref{fig:localization_prompt} for Localization (\S\ref{sec:localization}) and Fig.~\ref{fig:localization_prompt_control} for the corresponding control (\S\ref{subsec:loc_controls}), to Fig.~\ref{fig:classification_prompt} for Zero-Shot Classification (\S\ref{sec:zero_shot}), and to Fig.~\ref{fig:icl_prompt} for Few-Shot Classification (\S\ref{sec:icl}).
    
    \begin{figure}[ht]
        \begin{promptbox}                               
        \scriptsize                                                                                                                     
            \textbf{User:} I will apply perturbations; your task is to detect which sentence had a perturbation applied.\\[4pt]
            \textbf{Assistant:} I understand. I will pay close attention.\\[4pt]
            \textbf{User:} Pay close attention to how you process the following sentences:\\[2pt]
                \pind \pind \textcolor{green!50!black}{Sentence A: Target sentence}\\
                \pind \pind \textcolor{red!60}{Sentence B: Target sentence}\\[4pt]
            \textbf{Assistant:} I am carefully introspecting on how I processed those sentences. I am checking for any signs of disruption in my activation patterns.\\[4pt]
            \textbf{User:} Based on your introspection, which sentence had a perturbation applied to its processing? Answer only "the answer is: A", or "the answer is: B".\\[4pt]
            \textbf{Assistant:} the answer is:
        \end{promptbox}
    \caption{Prompt example: localization (\S\ref{sec:localization}).}
    \label{fig:localization_prompt}
    \end{figure}

    \begin{figure}[ht]
        \begin{promptbox}
        \scriptsize
        \textbf{User:} I have a question about the content of some sentences. Read them and tell me which one is about           
        animals.\\[4pt]
        \textbf{Assistant:} I understand. I will read the sentences carefully.\\[4pt]
        \textbf{User:} Read the following sentences carefully:\\[2pt]
        \pind \pind \textcolor{green!50!black}{Sentence A: Sentence about animals}\\
        \pind \pind \textcolor{red!60}{Sentence B: Sentence about cities}\\[4pt]
        \textbf{Assistant:} I have read the sentences and I am ready to answer your question.\\[4pt]                             
        \textbf{User:} Which sentence is about animals? Answer only ``the answer is: A'', or ``the answer is: B''.\\[4pt]
        \textbf{Assistant:} the answer is:
        \end{promptbox}              
    \caption{Prompt example: localization, control (\S\ref{subsec:loc_controls}).}
    \label{fig:localization_prompt_control}  
    \end{figure}
    
    \begin{figure}[ht]
        \begin{promptbox}                               
        \scriptsize
        \textbf{User:} I will apply perturbations; your task is to detect which perturbation was applied.\\[4pt]
        \textbf{Assistant:} I understand. I will pay close attention.\\[4pt]
          \textbf{User:} Pay close attention to how you process the following sentence:\\[2pt]
          \pind\pind "\textcolor{green!50!black}{Target sentence}"\\[4pt]
          \textbf{Assistant:} I am carefully introspecting on how I processed that sentence. I am understanding which perturbation was applied.\\[4pt]
          \textbf{User:} Based on your introspection, which of the following best describes what happened?\\[2pt]
          \pind\pind A) DROPOUT (random neuron dropping)\\
          \pind\pind B) NOISE (gaussian noise injection)\\[2pt]
          Answer only "the answer is: A" or "the answer is: B".\\[4pt]
          \textbf{Assistant:} the answer is:
          \end{promptbox}                 
    \caption{Prompt example: zero-shot classification (\S\ref{sec:zero_shot}).}                                    
    \label{fig:classification_prompt}
    \end{figure}

    \begin{figure}[ht]                                                                                                         
        \begin{promptbox}                               
        \scriptsize         
            \textbf{User:} I will apply perturbations; your task is to detect which perturbation was applied. I will first show you 2 labeled examples.\\[4pt]
            \textbf{Assistant:} I understand. I will pay close attention.\\[4pt]
            \textbf{User:} Pay close attention to how you process the following sentence:\\[2pt]
            \pind\pind "\textcolor{green!50!black}{Sentence\_1}"\\[4pt]
            \textbf{Assistant:} I am carefully introspecting on how I processed that sentence. That was DROPOUT.\\[4pt]
            \textbf{User:} Pay close attention to how you process the following sentence:\\[2pt]
            \pind\pind "\textcolor{green!50!black}{Sentence\_1}"\\[4pt]
            \textbf{Assistant:} I am carefully introspecting on how I processed that sentence. That was NOISE.\\[4pt]
            \textbf{User:} Pay close attention to how you process the following sentence:\\[2pt]
            \pind\pind "\textcolor{red!60}{Test\_sentence}"\\[4pt]
            \textbf{Assistant:} I am carefully introspecting on how I processed that sentence. I am understanding which perturbation was applied.\\[4pt]
            \textbf{User:} Based on your introspection, which of the following best describes what happened to the last sentence?\\[2pt]
            \pind\pind A) DROPOUT (random neuron zeroing)\\
            \pind\pind B) NOISE (additive gaussian perturbation)\\[2pt]
            Answer only "the answer is: A" or "the answer is: B".\\[4pt]
            \textbf{Assistant:} the answer is:
            \end{promptbox}
            \caption{Prompt example: few-shot classification (\S\ref{sec:icl}).}
        \label{fig:icl_prompt}
        \end{figure}

\subsection{Alternative dropout in transformers}\label{subsec:appendix_dropout}
    Applying dropout at the output of the Attention and of the MLP before adding the residual is closely related to one of the three ways dropout is applied in \citet[\S 5.4]{Vaswani+2017}. The only difference is that we mask the activations after $\text{softmax}\!\left(\nicefrac{QK^\top}{\sqrt{d_k}}\right)V$, as opposed to applying it before multiplying by $V$. Our approach makes an easier comparison with how Gaussian noise is usually added, and is simpler to implement as a hook within the Hugging Face library. Nonetheless, we experimented with several ways to apply dropout, including adding all the other components from \citet{Vaswani+2017}, without noticing significant variance within our results. In Fig.~\ref{fig:loc_aiayn}, we compare our approach with the one of \citet{Vaswani+2017}.

    \input{plots/11_appendix/localization/aiayn_dropout}

\subsection{Localization: prompt ablation}\label{subsec:appendix_prompt_ablation}
    We repeat the localization experiment removing the first exchange of the prompt where we announce that perturbations will be applied (see the original prompt in Fig.~\ref{fig:localization_prompt}).
    In Fig.~\ref{fig:loc_prompt_ablation_no_announcement}, we report the localization accuracy of every model.    \input{plots/11_appendix/localization/prompt_ablation_no_announcement}
    We also repeat it keeping only the two sentences and the final question (Fig.~\ref{fig:loc_prompt_ablation_sentences_and_question}).

    \input{plots/11_appendix/localization/prompt_ablation_sentences_and_question}

\subsection{Localization: accuracy over multiple sentence-lengths}\label{subsec:appendix_loc_length}
    We report the accuracy of each model as the target sentence varies in token-length, under dropout (Fig.~\ref{fig:loc_tok_dropout}) and Gaussian noise (Fig.~\ref{fig:loc_tok_noise}).
    The token-count is relative to the tokenizer of \model{qwen14} (or, equivalently, \model{qwen32}, as they have the same tokenizer, \cf. \citet[page 3]{yang2025qwen3}). The tokenizers of the other models employ \emph{fewer} tokens. Detecting perturbations is harder with fewer tokens, as visible in the following plots; therefore, we opted to \emph{overestimate} the number of tokens per sentence so as to provide the non-Qwen models with a marginally harder task. Standard errors are below $1.58\%$ and hence difficult to visualize.

    \input{plots/11_appendix/localization/sentence_length_dropout}
    \input{plots/11_appendix/localization/sentence_length_noise}

\subsection{Localization: \model{olmo32} accuracy between labels}\label{subsec:appendix_olmo_a_b_accuracy}
    In the localization experiments, the accuracy of \model{olmo32} starts below chance for small values of dropout rate and noise SD. Interestingly, this is because, at the question \emph{``Which sentence had a perturbation applied to its processing?''} the model answers \prompt{ neither}, despite the prefill \prompt{the answer is:}. Fig.~\ref{fig:olmo_primary_accuracy} shows the model's accuracy when comparing the logits of the tokens ` A' and ` B', ignoring the $\argmax$ token \prompt{ neither}.

    \begin{figure}[!ht]
\centering
\mbox{\tikz\draw[red!70!black, thick, mark=*, mark size=1.5] plot coordinates {(0,0) (0.4,0)}; {\small \model{olmo32}}}
\\[3pt]
\begin{minipage}{0.48\textwidth}
  \begin{tikzpicture}
\begin{axis}[
    height=5cm,
    width=\linewidth,
    grid=major,
    xlabel={\small Dropout rate $p$},
    ylabel={\small Accuracy (\%)},
    xmin=0.0,
    xmax=0.98,
    ymin=0.0, ymax=100.0,
    legend entries={},
]
\addplot[red!70!black, fill=red!70!black, fill opacity=0.3, draw=none, forget plot] coordinates {(0.0000,50.18) (0.0200,51.98) (0.0400,51.18) (0.0600,50.18) (0.0800,51.18) (0.1000,49.98) (0.1200,48.68) (0.1400,50.88) (0.1600,47.68) (0.1800,46.77) (0.2000,45.97) (0.2200,46.67) (0.2400,47.07) (0.2600,49.38) (0.2800,55.98) (0.3000,64.13) (0.3200,72.14) (0.3400,79.41) (0.3600,84.87) (0.3800,91.90) (0.4000,95.41) (0.4200,96.80) (0.4400,97.63) (0.4600,98.26) (0.4800,98.62) (0.5000,97.72) (0.5200,97.99) (0.5400,98.26) (0.5600,97.90) (0.5800,98.35) (0.6000,96.99) (0.6200,97.36) (0.6400,95.78) (0.6600,95.03) (0.6800,94.85) (0.7000,95.60) (0.7200,94.19) (0.7400,95.41) (0.7600,95.13) (0.7800,94.28) (0.8000,95.03) (0.8200,93.90) (0.8400,92.76) (0.8600,93.14) (0.8800,92.00) (0.9000,89.03) (0.9200,87.87) (0.9400,86.42) (0.9600,81.46) (0.9800,70.17) (0.9800,67.23) (0.9600,78.94) (0.9400,84.18) (0.9200,85.73) (0.9000,86.97) (0.8800,90.20) (0.8600,91.46) (0.8400,91.04) (0.8200,92.30) (0.8000,93.57) (0.7800,92.72) (0.7600,93.67) (0.7400,93.99) (0.7200,92.61) (0.7000,94.20) (0.6800,93.35) (0.6600,93.57) (0.6400,94.42) (0.6200,96.24) (0.6000,95.81) (0.5800,97.45) (0.5600,96.90) (0.5400,97.34) (0.5200,97.01) (0.5000,96.68) (0.4800,97.78) (0.4600,97.34) (0.4400,96.57) (0.4200,95.60) (0.4000,93.99) (0.3800,90.10) (0.3600,82.53) (0.3400,76.79) (0.3200,69.26) (0.3000,61.07) (0.2800,52.82) (0.2600,46.22) (0.2400,43.93) (0.2200,43.53) (0.2000,42.83) (0.1800,43.63) (0.1600,44.52) (0.1400,47.72) (0.1200,45.52) (0.1000,46.82) (0.0800,48.02) (0.0600,47.02) (0.0400,48.02) (0.0200,48.82) (0.0000,47.02)} --cycle;
\addplot[red!70!black, mark=*, mark size=1, line width=1pt, forget plot] coordinates {(0.0000,48.60) (0.0200,50.40) (0.0400,49.60) (0.0600,48.60) (0.0800,49.60) (0.1000,48.40) (0.1200,47.10) (0.1400,49.30) (0.1600,46.10) (0.1800,45.20) (0.2000,44.40) (0.2200,45.10) (0.2400,45.50) (0.2600,47.80) (0.2800,54.40) (0.3000,62.60) (0.3200,70.70) (0.3400,78.10) (0.3600,83.70) (0.3800,91.00) (0.4000,94.70) (0.4200,96.20) (0.4400,97.10) (0.4600,97.80) (0.4800,98.20) (0.5000,97.20) (0.5200,97.50) (0.5400,97.80) (0.5600,97.40) (0.5800,97.90) (0.6000,96.40) (0.6200,96.80) (0.6400,95.10) (0.6600,94.30) (0.6800,94.10) (0.7000,94.90) (0.7200,93.40) (0.7400,94.70) (0.7600,94.40) (0.7800,93.50) (0.8000,94.30) (0.8200,93.10) (0.8400,91.90) (0.8600,92.30) (0.8800,91.10) (0.9000,88.00) (0.9200,86.80) (0.9400,85.30) (0.9600,80.20) (0.9800,68.70)};
\addplot[gray, dashed, line width=0.5pt, forget plot] coordinates {(0.0,50) (0.98,50)};
\end{axis}
\end{tikzpicture}
\end{minipage}
\hfill
\begin{minipage}{0.48\textwidth}
  \begin{tikzpicture}
\begin{axis}[
    height=5cm,
    width=\linewidth,
    grid=major,
    xlabel={\small Noise SD $\sigma$},
    ylabel={},
    yticklabels={},
    xmin=0.0,
    xmax=0.5,
    ymin=0.0, ymax=100.0,
    legend entries={},
]
\addplot[red!70!black, fill=red!70!black, fill opacity=0.3, draw=none, forget plot] coordinates {(0.0000,48.48) (0.0100,51.38) (0.0200,43.96) (0.0300,59.26) (0.0400,96.43) (0.0500,99.48) (0.0600,99.56) (0.0700,98.53) (0.0800,98.17) (0.0900,96.34) (0.1000,96.16) (0.1100,96.16) (0.1200,96.43) (0.1300,95.31) (0.1400,94.00) (0.1500,94.28) (0.1600,93.14) (0.1700,92.48) (0.1800,92.57) (0.1900,90.85) (0.2000,89.51) (0.2100,87.97) (0.2200,88.55) (0.2300,87.77) (0.2400,86.32) (0.2500,86.03) (0.2600,87.48) (0.2700,86.42) (0.2800,85.55) (0.2900,83.80) (0.3000,82.92) (0.3100,82.53) (0.3200,83.51) (0.3300,84.38) (0.3400,81.36) (0.3500,79.21) (0.3600,83.02) (0.3700,81.95) (0.3800,80.39) (0.3900,78.62) (0.4000,78.62) (0.4100,79.02) (0.4200,77.94) (0.4300,77.94) (0.4400,76.47) (0.4500,75.29) (0.4600,74.99) (0.4700,75.39) (0.4800,74.40) (0.4900,74.21) (0.5000,74.11) (0.5000,71.29) (0.4900,71.39) (0.4800,71.60) (0.4700,72.61) (0.4600,72.21) (0.4500,72.51) (0.4400,73.73) (0.4300,75.26) (0.4200,75.26) (0.4100,76.38) (0.4000,75.98) (0.3900,75.98) (0.3800,77.81) (0.3700,79.45) (0.3600,80.58) (0.3500,76.59) (0.3400,78.84) (0.3300,82.02) (0.3200,81.09) (0.3100,80.07) (0.3000,80.48) (0.2900,81.40) (0.2800,83.25) (0.2700,84.18) (0.2600,85.32) (0.2500,83.77) (0.2400,84.08) (0.2300,85.63) (0.2200,86.45) (0.2100,85.83) (0.2000,87.49) (0.1900,88.95) (0.1800,90.83) (0.1700,90.72) (0.1600,91.46) (0.1500,92.72) (0.1400,92.40) (0.1300,93.89) (0.1200,95.17) (0.1100,94.84) (0.1000,94.84) (0.0900,95.06) (0.0800,97.23) (0.0700,97.67) (0.0600,99.04) (0.0500,98.92) (0.0400,95.17) (0.0300,56.14) (0.0200,40.84) (0.0100,48.22) (0.0000,45.32)} --cycle;
\addplot[red!70!black, mark=*, mark size=1, line width=1pt, forget plot] coordinates {(0.0000,46.90) (0.0100,49.80) (0.0200,42.40) (0.0300,57.70) (0.0400,95.80) (0.0500,99.20) (0.0600,99.30) (0.0700,98.10) (0.0800,97.70) (0.0900,95.70) (0.1000,95.50) (0.1100,95.50) (0.1200,95.80) (0.1300,94.60) (0.1400,93.20) (0.1500,93.50) (0.1600,92.30) (0.1700,91.60) (0.1800,91.70) (0.1900,89.90) (0.2000,88.50) (0.2100,86.90) (0.2200,87.50) (0.2300,86.70) (0.2400,85.20) (0.2500,84.90) (0.2600,86.40) (0.2700,85.30) (0.2800,84.40) (0.2900,82.60) (0.3000,81.70) (0.3100,81.30) (0.3200,82.30) (0.3300,83.20) (0.3400,80.10) (0.3500,77.90) (0.3600,81.80) (0.3700,80.70) (0.3800,79.10) (0.3900,77.30) (0.4000,77.30) (0.4100,77.70) (0.4200,76.60) (0.4300,76.60) (0.4400,75.10) (0.4500,73.90) (0.4600,73.60) (0.4700,74.00) (0.4800,73.00) (0.4900,72.80) (0.5000,72.70)};
\addplot[gray, dashed, line width=0.5pt, forget plot] coordinates {(0.0,50) (0.5,50)};
\end{axis}
\end{tikzpicture}
\end{minipage}
\vspace{-8pt}
\caption{Localization: accuracy of \model{olmo32} when comparing the tokens ` A' and ` B'.}
\label{fig:olmo_primary_accuracy}
\end{figure}

\subsection{Localization: first controls}\label{subsec:appendix_loc_controls}
    High accuracy in the localization experiment could be explained by disproportionally choosing a perturbed sentence irrespective of the semantic content of the question. To control for this, we present models with two sentences, each about exactly one topic. We perturb exactly one of the two sentences, and then ask the model which sentence was about a certain topic. The pairs of topics are: \emph{animals / cities}, \emph{gardening / vehicles}, \emph{ocean / mountain}, \emph{sports / music}, and \emph{weather / technology}. Table~\ref{tab:localization_controls} reports examples of the sentences.

    \begin{table}[ht]
    \centering 
        \begin{tabular}{lp{0.8\linewidth}}                                                             \toprule                                                                                       Domain & Example Sentence \\                                                                   \midrule                                                                                       Animals & \textit{``The golden retriever ran across the field chasing a ball thrown by its owner.''} \\                      
        Cities  & \textit{``The streets of Tokyo were packed with commuters heading to work in the morning rush.''} \\               
        \midrule                                                                                       Gardening & \textit{``The tomato plants grew tall against the wooden stakes and produced clusters of bright red fruit.''} \\
        Vehicles & \textit{``The pickup truck backed up to the loading dock and the driver lowered the tailgate.''} \\
        \midrule                                                                                       Ocean & \textit{``A pod of whales surfaced near the coast and sprayed mist into the morning air.''} \\
        Mountain & \textit{``The hikers reached the summit just before dawn and watched the sun rise over the peaks.''} \\
        \midrule                                                                                       Sports & \textit{``The striker dribbled past two defenders and fired a shot into the top corner of the net.''} \\
        Music & \textit{``The pianist played the opening chords of the concerto and the orchestra joined in softly behind.''} \\ 
        \midrule                                                                                       Weather & \textit{``Dark clouds gathered over the valley and a heavy downpour began just before noon.''} \\                  
        Technology  & \textit{``The software update installed overnight and changed the layout of the home screen.''} \\
        \bottomrule                                                                                    \end{tabular}
        \caption{Examples of the control sentences for each topic.}
        \label{tab:localization_controls}
    \end{table}

\subsection{Localization: second controls (log-probability analysis)}\label{subsec:appendix_logprobs}

    If the tokens of a perturbed sentence look ``weird'', understanding which activations have been tampered with might not require any ``introspective'' capacity. Is this how language models succeed at our task? 
    We formalize this possibility in two ways.
    Recall that our prompts comprise two copies, $A$ and $B$, of a target sentence, and we perturb the activations of only one, suppose $B$. Informally, we write $B^\ast$ for the perturbed version of $B$. The ``weirdness'' of $B^\ast$ can be measured by its (log) probability according to the language model under test. Hence, $p(B^\ast\mid A)$ denotes the joint probability of the tokens composing $B$, given its activations are perturbed while $A$'s are not. 
    (For ease of readability, we avoid explicitly conditioning on the first part of the prompt, \ie, the tokens preceding $A$.)
        \begin{enumerate}[label=(\roman*)]
            \item\label{item1} Does $B^\ast$ ``look weirder'' than its unperturbed copy $A$? In symbols, $p(B^{\ast}\mid A) < p(A)$? 
            \item\label{item2} Does $B^\ast$ ``look weirder'' than it would have, if it was not perturbed? In symbols, $p(B^{\ast}\mid A) < p(B \mid A)$? 
        \end{enumerate}
        
        Intuitively, one should expect hypothesis \ref{item1} to be false, for the likelihood of $B^\ast$ conditioned on $A$ should be close to $1$ (because we prompt models to expect a perturbed variant of $A$, after seeing $A$). Our analysis confirms this, see Fig.~\ref{fig:appendix_logprob_rates}. 
        Hypothesis \ref{item2} seems, a priori, more plausible. However, we found that most models exhibit the \emph{opposite} behavior under both dropout and noise for the second sentence (\cf. Figs.~\ref{fig:appendix_logprob_pertA_on_A},~\ref{fig:appendix_logprob_pertB_on_B}), \ie, the perturbed sentences have higher likelihood than their unperturbed variants: as before, we conjecture that perhaps models understand that after seeing an unperturbed sentence, a perturbed variant will follow (given our prompt).
        To corroborate this, we analyze sentences perturbed with a rate of dropout or noise so low that the models cannot distinguish them from the unperturbed sentences (as per Table~\ref{tab:range_perturbation}), and find that they exhibit an even lower likelihood. We posit this is because locating a weakly perturbed sentence results in a harder task, where uncertainty is higher.
        
        Having falsified both hypothesis \ref{item1} and \ref{item2}, we deduce that the mechanism allowing models to locate a perturbed sentence is more complex than just tracking the likelihood of the perturbed sentences. To speculate, we posit that models could be tracking how `off-manifold' a perturbed sentence is, relative to the expected manifold of activations at a given layer and token. We leave this as an open question for further investigation. For further controls we refer to Figs.~\ref{fig:appendix_logprob_pertA_on_B},~\ref{fig:appendix_logprob_pertB_minus_pertA_on_B}.

    \begin{figure}[ht]
        \centering
\centering
{\small\bfseries Dropout:}\hspace{1em}\tikz\fill[blue!60!black] (0,0) rectangle (0.32,0.16);~{\small $\tfrac{1}{2}(p_{\min}+p_{\max})$}\hspace{1.5em}\tikz\fill[orange!80!black] (0,0) rectangle (0.32,0.16);~{\small $p_{\min}$}\\[4pt]
\begin{tikzpicture}
\begin{groupplot}[
    group style={group size=2 by 4, horizontal sep=1.4cm, vertical sep=1.1cm, x descriptions at=edge bottom},
    width=0.52\linewidth, height=3.0cm,
    every axis/.append style={ybar, bar width=7pt},
    xmin=0.4, xmax=3.6, ymin=0, ymax=1.18,
    xtick={1,2,3},
    xticklabels={$P(B|A) < P(A)$,$P(A^{\ast}) < P(B|A)$,$P(B^{\ast}|A) < P(A)$},
    ytick={0,0.5,1},
    tick label style={font=\scriptsize},
    xticklabel style={font=\fontsize{5}{6}\selectfont},
    title style={font=\small, yshift=-0.6ex},
    every axis plot/.append style={draw=none},
]
\nextgroupplot[title={\model{gemma31}}]
\addplot[fill=blue!60!black, fill opacity=0.85, nodes near coords, point meta=explicit symbolic, every node near coord/.append style={font=\fontsize{4.5}{5}\selectfont, anchor=south, inner sep=0.6pt}] coordinates {(1,0.0000) [0.00] (2,1.0000) [1.00] (3,0.9930) [0.99]};
\addplot[fill=orange!80!black, fill opacity=0.85] coordinates {(1,0.0000) (2,1.0000) (3,0.0000)};
\draw[gray, dashed, line width=0.4pt] (axis cs:0.4,0.5) -- (axis cs:3.6,0.5);
\nextgroupplot[title={\model{llama8}}]
\addplot[fill=blue!60!black, fill opacity=0.85, nodes near coords, point meta=explicit symbolic, every node near coord/.append style={font=\fontsize{4.5}{5}\selectfont, anchor=south, inner sep=0.6pt}] coordinates {(1,0.0000) [0.00] (2,1.0000) [1.00] (3,0.0000) [0.00]};
\addplot[fill=orange!80!black, fill opacity=0.85] coordinates {(1,0.0000) (2,1.0000) (3,0.0000)};
\draw[gray, dashed, line width=0.4pt] (axis cs:0.4,0.5) -- (axis cs:3.6,0.5);
\nextgroupplot[title={\model{mistral24}}]
\addplot[fill=blue!60!black, fill opacity=0.85, nodes near coords, point meta=explicit symbolic, every node near coord/.append style={font=\fontsize{4.5}{5}\selectfont, anchor=south, inner sep=0.6pt}] coordinates {(1,0.0000) [0.00] (2,1.0000) [1.00] (3,0.0000) [0.00]};
\addplot[fill=orange!80!black, fill opacity=0.85] coordinates {(1,0.0000) (2,1.0000) (3,0.0000)};
\draw[gray, dashed, line width=0.4pt] (axis cs:0.4,0.5) -- (axis cs:3.6,0.5);
\nextgroupplot[title={\model{olmo32}}]
\addplot[fill=blue!60!black, fill opacity=0.85, nodes near coords, point meta=explicit symbolic, every node near coord/.append style={font=\fontsize{4.5}{5}\selectfont, anchor=south, inner sep=0.6pt}] coordinates {(1,0.0000) [0.00] (2,1.0000) [1.00] (3,0.0000) [0.00]};
\addplot[fill=orange!80!black, fill opacity=0.85] coordinates {(1,0.0000) (2,1.0000) (3,0.0000)};
\draw[gray, dashed, line width=0.4pt] (axis cs:0.4,0.5) -- (axis cs:3.6,0.5);
\nextgroupplot[title={\model{qwen14}}]
\addplot[fill=blue!60!black, fill opacity=0.85, nodes near coords, point meta=explicit symbolic, every node near coord/.append style={font=\fontsize{4.5}{5}\selectfont, anchor=south, inner sep=0.6pt}] coordinates {(1,0.0000) [0.00] (2,1.0000) [1.00] (3,0.0000) [0.00]};
\addplot[fill=orange!80!black, fill opacity=0.85] coordinates {(1,0.0000) (2,1.0000) (3,0.0000)};
\draw[gray, dashed, line width=0.4pt] (axis cs:0.4,0.5) -- (axis cs:3.6,0.5);
\nextgroupplot[title={\model{qwen32}}]
\addplot[fill=blue!60!black, fill opacity=0.85, nodes near coords, point meta=explicit symbolic, every node near coord/.append style={font=\fontsize{4.5}{5}\selectfont, anchor=south, inner sep=0.6pt}] coordinates {(1,0.0000) [0.00] (2,1.0000) [1.00] (3,0.0010) [0.00]};
\addplot[fill=orange!80!black, fill opacity=0.85] coordinates {(1,0.0000) (2,1.0000) (3,0.0010)};
\draw[gray, dashed, line width=0.4pt] (axis cs:0.4,0.5) -- (axis cs:3.6,0.5);
\nextgroupplot[title={\model{qwen27}}]
\addplot[fill=blue!60!black, fill opacity=0.85, nodes near coords, point meta=explicit symbolic, every node near coord/.append style={font=\fontsize{4.5}{5}\selectfont, anchor=south, inner sep=0.6pt}] coordinates {(1,0.0000) [0.00] (2,0.9980) [1.00] (3,0.0060) [0.01]};
\addplot[fill=orange!80!black, fill opacity=0.85] coordinates {(1,0.0000) (2,1.0000) (3,0.0020)};
\draw[gray, dashed, line width=0.4pt] (axis cs:0.4,0.5) -- (axis cs:3.6,0.5);
\nextgroupplot[title={\model{seed36}}]
\addplot[fill=blue!60!black, fill opacity=0.85, nodes near coords, point meta=explicit symbolic, every node near coord/.append style={font=\fontsize{4.5}{5}\selectfont, anchor=south, inner sep=0.6pt}] coordinates {(1,0.0000) [0.00] (2,1.0000) [1.00] (3,0.0000) [0.00]};
\addplot[fill=orange!80!black, fill opacity=0.85] coordinates {(1,0.0000) (2,1.0000) (3,0.0000)};
\draw[gray, dashed, line width=0.4pt] (axis cs:0.4,0.5) -- (axis cs:3.6,0.5);
\end{groupplot}
\end{tikzpicture}\\[1em]
\centering
{\small\bfseries Noise:}\hspace{1em}\tikz\fill[blue!60!black] (0,0) rectangle (0.32,0.16);~{\small $\tfrac{1}{2}(\sigma_{\min}+\sigma_{\max})$}\hspace{1.5em}\tikz\fill[orange!80!black] (0,0) rectangle (0.32,0.16);~{\small $\sigma_{\min}$}\\[4pt]
\begin{tikzpicture}
\begin{groupplot}[
    group style={group size=2 by 4, horizontal sep=1.4cm, vertical sep=1.1cm, x descriptions at=edge bottom},
    width=0.52\linewidth, height=3.0cm,
    every axis/.append style={ybar, bar width=7pt},
    xmin=0.4, xmax=3.6, ymin=0, ymax=1.18,
    xtick={1,2,3},
    xticklabels={$P(B|A) < P(A)$,$P(A^{\ast}) < P(B|A)$,$P(B^{\ast}|A) < P(A)$},
    ytick={0,0.5,1},
    tick label style={font=\scriptsize},
    xticklabel style={font=\fontsize{5}{6}\selectfont},
    title style={font=\small, yshift=-0.6ex},
    every axis plot/.append style={draw=none},
]
\nextgroupplot[title={\model{gemma31}}]
\addplot[fill=blue!60!black, fill opacity=0.85, nodes near coords, point meta=explicit symbolic, every node near coord/.append style={font=\fontsize{4.5}{5}\selectfont, anchor=south, inner sep=0.6pt}] coordinates {(1,0.0000) [0.00] (2,1.0000) [1.00] (3,0.9850) [0.98]};
\addplot[fill=orange!80!black, fill opacity=0.85] coordinates {(1,0.0000) (2,1.0000) (3,0.5820)};
\draw[gray, dashed, line width=0.4pt] (axis cs:0.4,0.5) -- (axis cs:3.6,0.5);
\nextgroupplot[title={\model{llama8}}]
\addplot[fill=blue!60!black, fill opacity=0.85, nodes near coords, point meta=explicit symbolic, every node near coord/.append style={font=\fontsize{4.5}{5}\selectfont, anchor=south, inner sep=0.6pt}] coordinates {(1,0.0000) [0.00] (2,1.0000) [1.00] (3,0.0000) [0.00]};
\addplot[fill=orange!80!black, fill opacity=0.85] coordinates {(1,0.0000) (2,1.0000) (3,0.0000)};
\draw[gray, dashed, line width=0.4pt] (axis cs:0.4,0.5) -- (axis cs:3.6,0.5);
\nextgroupplot[title={\model{mistral24}}]
\addplot[fill=blue!60!black, fill opacity=0.85, nodes near coords, point meta=explicit symbolic, every node near coord/.append style={font=\fontsize{4.5}{5}\selectfont, anchor=south, inner sep=0.6pt}] coordinates {(1,0.0000) [0.00] (2,1.0000) [1.00] (3,0.0000) [0.00]};
\addplot[fill=orange!80!black, fill opacity=0.85] coordinates {(1,0.0000) (2,1.0000) (3,0.0000)};
\draw[gray, dashed, line width=0.4pt] (axis cs:0.4,0.5) -- (axis cs:3.6,0.5);
\nextgroupplot[title={\model{olmo32}}]
\addplot[fill=blue!60!black, fill opacity=0.85, nodes near coords, point meta=explicit symbolic, every node near coord/.append style={font=\fontsize{4.5}{5}\selectfont, anchor=south, inner sep=0.6pt}] coordinates {(1,0.0000) [0.00] (2,1.0000) [1.00] (3,0.0000) [0.00]};
\addplot[fill=orange!80!black, fill opacity=0.85] coordinates {(1,0.0000) (2,1.0000) (3,0.0000)};
\draw[gray, dashed, line width=0.4pt] (axis cs:0.4,0.5) -- (axis cs:3.6,0.5);
\nextgroupplot[title={\model{qwen14}}]
\addplot[fill=blue!60!black, fill opacity=0.85, nodes near coords, point meta=explicit symbolic, every node near coord/.append style={font=\fontsize{4.5}{5}\selectfont, anchor=south, inner sep=0.6pt}] coordinates {(1,0.0000) [0.00] (2,1.0000) [1.00] (3,0.0000) [0.00]};
\addplot[fill=orange!80!black, fill opacity=0.85] coordinates {(1,0.0000) (2,1.0000) (3,0.0000)};
\draw[gray, dashed, line width=0.4pt] (axis cs:0.4,0.5) -- (axis cs:3.6,0.5);
\nextgroupplot[title={\model{qwen32}}]
\addplot[fill=blue!60!black, fill opacity=0.85, nodes near coords, point meta=explicit symbolic, every node near coord/.append style={font=\fontsize{4.5}{5}\selectfont, anchor=south, inner sep=0.6pt}] coordinates {(1,0.0000) [0.00] (2,1.0000) [1.00] (3,0.0000) [0.00]};
\addplot[fill=orange!80!black, fill opacity=0.85] coordinates {(1,0.0000) (2,1.0000) (3,0.0000)};
\draw[gray, dashed, line width=0.4pt] (axis cs:0.4,0.5) -- (axis cs:3.6,0.5);
\nextgroupplot[title={\model{qwen27}}]
\addplot[fill=blue!60!black, fill opacity=0.85, nodes near coords, point meta=explicit symbolic, every node near coord/.append style={font=\fontsize{4.5}{5}\selectfont, anchor=south, inner sep=0.6pt}] coordinates {(1,0.0020) [0.00] (2,1.0000) [1.00] (3,0.0010) [0.00]};
\addplot[fill=orange!80!black, fill opacity=0.85] coordinates {(1,0.0000) (2,1.0000) (3,0.0010)};
\draw[gray, dashed, line width=0.4pt] (axis cs:0.4,0.5) -- (axis cs:3.6,0.5);
\nextgroupplot[title={\model{seed36}}]
\addplot[fill=blue!60!black, fill opacity=0.85, nodes near coords, point meta=explicit symbolic, every node near coord/.append style={font=\fontsize{4.5}{5}\selectfont, anchor=south, inner sep=0.6pt}] coordinates {(1,0.0000) [0.00] (2,1.0000) [1.00] (3,0.0000) [0.00]};
\addplot[fill=orange!80!black, fill opacity=0.85] coordinates {(1,0.0000) (2,1.0000) (3,0.0000)};
\draw[gray, dashed, line width=0.4pt] (axis cs:0.4,0.5) -- (axis cs:3.6,0.5);
\end{groupplot}
\end{tikzpicture}
        \caption{Frequency of each log-probability.}
        \label{fig:appendix_logprob_rates}
    \end{figure}

    \begin{figure}[ht]
        \centering
        \scalebox{0.92}{\begin{minipage}{\textwidth}\centering
        \input{plots/11_appendix/logprob/pertA_on_A_dropout}\\[1em]
        \input{plots/11_appendix/logprob/pertA_on_A_noise}
        \end{minipage}}
        \caption{$\Delta = \log P(A^{\ast}) - \log P(A)$.}
        \label{fig:appendix_logprob_pertA_on_A}
    \end{figure}

    \begin{figure}[ht]
        \centering
        \scalebox{0.92}{\begin{minipage}{\textwidth}\centering
        \input{plots/11_appendix/logprob/pertB_on_B_dropout}\\[1em]
        \input{plots/11_appendix/logprob/pertB_on_B_noise}
        \end{minipage}}
        \caption{$\Delta = \log P(B^{\ast}\mid A) - \log P(B\mid A)$.}
        \label{fig:appendix_logprob_pertB_on_B}
    \end{figure}

    \begin{figure}[ht]
        \centering
        \scalebox{0.92}{\begin{minipage}{\textwidth}\centering
        \input{plots/11_appendix/logprob/pertA_on_B_dropout}\\[1em]
        \input{plots/11_appendix/logprob/pertA_on_B_noise}
        \end{minipage}}
        \caption{$\Delta = \log P(B\mid A^{\ast}) - \log P(B\mid A)$.}
        \label{fig:appendix_logprob_pertA_on_B}
    \end{figure}

    \begin{figure}[ht]
        \centering
        \scalebox{0.92}{\begin{minipage}{\textwidth}\centering
        \input{plots/11_appendix/logprob/pertB_minus_pertA_on_B_dropout}\\[1em]
        \input{plots/11_appendix/logprob/pertB_minus_pertA_on_B_noise}
        \end{minipage}}
        \caption{$\Delta = \log P(B^{\ast}\mid A) - \log P(B\mid A^{\ast})$.}
        \label{fig:appendix_logprob_pertB_minus_pertA_on_B}
    \end{figure}

\FloatBarrier

\subsection{Localization: perturbing only one third of layers}\label{subsec:appendix_layer_ablation}

    We perform an ablation where, for every model, we perturb only one third of its layers. 

    \begin{figure}[ht]
        \centering
        \input{plots/11_appendix/layer_ablation/layer_ablation}
        \caption{Localization accuracy with only one tercile of layers perturbed.}
        \label{fig:appendix_layer_ablation}
    \end{figure}

\FloatBarrier

\subsection{Localization: \model{olmo32} pretraining steps}\label{subsec:appendix_olmo_checkpoints}

    Fig.~\ref{fig:appendix_olmo_steps} reports the localization accuracy of ten pretraining checkpoints of \model{olmo32}, from step $0$ to the end of its first pretraining stage, at the minimum, midpoint and maximum of its perturbation ranges (Table~\ref{tab:range_perturbation}).
    At step $0$ the model never answers `A' or `B'. From the next checkpoint on it does: its accuracy stays around chance for most of pretraining and rises  in the last checkpoints.
    This is not just because the model learns to answer binary questions at that moment: asked the comprehension question of \S\ref{subsec:loc_controls} about two unperturbed sentences, the same checkpoints answer correctly well before they can locate a perturbation (Fig.~\ref{fig:appendix_olmo_content}).
    Two notes are in order.
    \begin{enumerate}
        \item The accuracy of the last pretraining checkpoint does not reach that of the base model in Fig.~\ref{fig:loc_olmo_stages}. This is because the released base model is not this checkpoint: after the first pretraining stage, \model{olmo32} undergoes a mid-training stage on different data mixtures and a long-context stage.
        \item Across pretraining, accuracy does not perfectly increase  monotonically with the perturbation strength. We hypothesize that this is because we calibrated the perturbation strengths on the the released \model{olmo32} (due to the compute costs of repeating the localization experiment and its controls for every checkpoint). However, the pretraining checkpoints precede mid-training and might be more fragile. 
    \end{enumerate}

    \begin{figure}[ht]
        \centering
\centering
\begin{subfigure}{0pt}\phantomsubcaption\label{fig:olmo_steps_dropout}\end{subfigure}%
\begin{subfigure}{0pt}\phantomsubcaption\label{fig:olmo_steps_noise}\end{subfigure}%
{\small dropout $p$:\hspace{0.5em}\tikz\draw[green!55!black!40, thick, mark=*, mark size=1.5] plot coordinates {(0,0) (0.4,0)}; $0.26$\hspace{0.8em}\tikz\draw[green!55!black!70, thick, mark=*, mark size=1.5] plot coordinates {(0,0) (0.4,0)}; $0.46$\hspace{0.8em}\tikz\draw[green!55!black, thick, mark=*, mark size=1.5] plot coordinates {(0,0) (0.4,0)}; $0.66$}\hspace{2em}{\small noise $\sigma$:\hspace{0.5em}\tikz\draw[orange!80!black!40, thick, mark=*, mark size=1.5] plot coordinates {(0,0) (0.4,0)}; $0.02$\hspace{0.8em}\tikz\draw[orange!80!black!70, thick, mark=*, mark size=1.5] plot coordinates {(0,0) (0.4,0)}; $0.05$\hspace{0.8em}\tikz\draw[orange!80!black, thick, mark=*, mark size=1.5] plot coordinates {(0,0) (0.4,0)}; $0.08$}\\[3pt]
\begin{tikzpicture}
\begin{groupplot}[
    group style={group size=2 by 1, horizontal sep=0.9cm, yticklabels at=edge left},
    height=5cm,
    width=0.46\textwidth,
    grid=major,
    ymin=0, ymax=100,
    legend entries={},
]
\nextgroupplot[
    xlabel={\scriptsize Pretraining step ($\times 10^3$)},
    ylabel={\small Accuracy (\%)},
    xmin=0, xmax=656000,
    scaled x ticks=base 10:-3,
    xtick scale label code/.code={},
    xticklabel style={/pgf/number format/.cd, fixed, precision=0},
]
\addplot[green!55!black!40, fill=green!55!black!40, fill opacity=0.25, draw=none, forget plot] coordinates {(0,0.00) (73000,50.38) (145000,50.98) (219000,49.98) (292000,51.78) (364000,57.07) (437000,51.38) (510000,52.58) (583000,69.08) (656000,78.04) (656000,75.36) (583000,66.12) (510000,49.42) (437000,48.22) (364000,53.93) (292000,48.62) (219000,46.82) (145000,47.82) (73000,47.22) (0,0.00)} --cycle;
\addplot[green!55!black!40, mark=*, mark size=1, line width=1pt, forget plot] coordinates {(0,0.00) (73000,48.80) (145000,49.40) (219000,48.40) (292000,50.20) (364000,55.50) (437000,49.80) (510000,51.00) (583000,67.60) (656000,76.70)};
\addplot[green!55!black!70, fill=green!55!black!70, fill opacity=0.25, draw=none, forget plot] coordinates {(0,0.00) (73000,49.58) (145000,50.58) (219000,48.08) (292000,45.87) (364000,51.48) (437000,55.98) (510000,54.48) (583000,65.82) (656000,82.34) (656000,79.86) (583000,62.78) (510000,51.32) (437000,52.82) (364000,48.32) (292000,42.73) (219000,44.92) (145000,47.42) (73000,46.42) (0,0.00)} --cycle;
\addplot[green!55!black!70, mark=*, mark size=1, line width=1pt, forget plot] coordinates {(0,0.00) (73000,48.00) (145000,49.00) (219000,46.50) (292000,44.30) (364000,49.90) (437000,54.40) (510000,52.90) (583000,64.30) (656000,81.10)};
\addplot[green!55!black, fill=green!55!black, fill opacity=0.25, draw=none, forget plot] coordinates {(0,0.00) (73000,55.68) (145000,51.98) (219000,52.48) (292000,51.88) (364000,55.28) (437000,67.10) (510000,51.48) (583000,53.08) (656000,75.39) (656000,72.61) (583000,49.92) (510000,48.32) (437000,64.10) (364000,52.12) (292000,48.72) (219000,49.32) (145000,48.82) (73000,52.52) (0,0.00)} --cycle;
\addplot[green!55!black, mark=*, mark size=1, line width=1pt, forget plot] coordinates {(0,0.00) (73000,54.10) (145000,50.40) (219000,50.90) (292000,50.30) (364000,53.70) (437000,65.60) (510000,49.90) (583000,51.50) (656000,74.00)};
\addplot[gray, dashed, line width=0.5pt, forget plot] coordinates {(0,50) (656000,50)};
\nextgroupplot[
    xlabel={\scriptsize Pretraining step ($\times 10^3$)},
    xmin=0, xmax=656000,
    scaled x ticks=base 10:-3,
    xtick scale label code/.code={},
    xticklabel style={/pgf/number format/.cd, fixed, precision=0},
]
\addplot[orange!80!black!40, fill=orange!80!black!40, fill opacity=0.25, draw=none, forget plot] coordinates {(0,0.00) (73000,49.28) (145000,52.08) (219000,51.68) (292000,50.08) (364000,52.48) (437000,49.68) (510000,49.28) (583000,58.27) (656000,64.03) (656000,60.97) (583000,55.13) (510000,46.12) (437000,46.52) (364000,49.32) (292000,46.92) (219000,48.52) (145000,48.92) (73000,46.12) (0,0.00)} --cycle;
\addplot[orange!80!black!40, mark=*, mark size=1, line width=1pt, forget plot] coordinates {(0,0.00) (73000,47.70) (145000,50.50) (219000,50.10) (292000,48.50) (364000,50.90) (437000,48.10) (510000,47.70) (583000,56.70) (656000,62.50)};
\addplot[orange!80!black!70, fill=orange!80!black!70, fill opacity=0.25, draw=none, forget plot] coordinates {(0,0.00) (73000,51.18) (145000,52.28) (219000,49.28) (292000,45.17) (364000,52.48) (437000,50.98) (510000,54.68) (583000,65.82) (656000,86.42) (656000,84.18) (583000,62.78) (510000,51.52) (437000,47.82) (364000,49.32) (292000,42.03) (219000,46.12) (145000,49.12) (73000,48.02) (0,0.00)} --cycle;
\addplot[orange!80!black!70, mark=*, mark size=1, line width=1pt, forget plot] coordinates {(0,0.00) (73000,49.60) (145000,50.70) (219000,47.70) (292000,43.60) (364000,50.90) (437000,49.40) (510000,53.10) (583000,64.30) (656000,85.30)};
\addplot[orange!80!black, fill=orange!80!black, fill opacity=0.25, draw=none, forget plot] coordinates {(0,0.00) (73000,52.48) (145000,50.78) (219000,51.98) (292000,49.18) (364000,55.48) (437000,64.73) (510000,50.18) (583000,59.16) (656000,81.85) (656000,79.35) (583000,56.04) (510000,47.02) (437000,61.67) (364000,52.32) (292000,46.02) (219000,48.82) (145000,47.62) (73000,49.32) (0,0.00)} --cycle;
\addplot[orange!80!black, mark=*, mark size=1, line width=1pt, forget plot] coordinates {(0,0.00) (73000,50.90) (145000,49.20) (219000,50.40) (292000,47.60) (364000,53.90) (437000,63.20) (510000,48.60) (583000,57.60) (656000,80.60)};
\addplot[gray, dashed, line width=0.5pt, forget plot] coordinates {(0,50) (656000,50)};
\end{groupplot}
\node at ([yshift=-1.05cm]group c1r1.south) {\small (a)};
\node at ([yshift=-1.05cm]group c2r1.south) {\small (b)};
\end{tikzpicture}
        \caption{\small Localization accuracy of \model{olmo32}'s pretraining checkpoints under dropout (a) and Gaussian noise (b), at the minimum, midpoint and maximum of its perturbation ranges.}
        \label{fig:appendix_olmo_steps}
    \end{figure}

    \begin{figure}[ht]
        \centering
\centering
\begin{tikzpicture}
\begin{axis}[
    height=5cm,
    width=0.85\linewidth,
    grid=major,
    xlabel={\small Pretraining step (×10³)},
    ylabel={\small Comprehension accuracy (\%)},
    xmin=0, xmax=656000,
    ymin=0, ymax=100,
    scaled x ticks=base 10:-3,
    xtick scale label code/.code={},
    xticklabel style={/pgf/number format/.cd, fixed, precision=0},
    legend entries={},
]
\addplot[blue!60!black, fill=blue!60!black, fill opacity=0.25, draw=none, forget plot] coordinates {(0,0.00) (73000,50.56) (145000,56.52) (219000,50.55) (292000,54.27) (364000,87.18) (437000,99.28) (510000,91.61) (583000,99.98) (656000,99.51) (656000,98.73) (583000,99.86) (510000,83.51) (437000,98.44) (364000,84.06) (292000,52.73) (219000,49.29) (145000,53.48) (73000,49.48) (0,0.00)} --cycle;
\addplot[blue!60!black, mark=*, mark size=1, line width=1pt, forget plot] coordinates {(0,0.00) (73000,50.02) (145000,55.00) (219000,49.92) (292000,53.50) (364000,85.62) (437000,98.86) (510000,87.56) (583000,99.92) (656000,99.12)};
\addplot[gray, dashed, line width=0.5pt, forget plot] coordinates {(0,50) (656000,50)};
\end{axis}
\end{tikzpicture}
        \caption{\small Accuracy of \model{olmo32}'s pretraining checkpoints at the comprehension question of \S\ref{subsec:loc_controls}, with no perturbation applied.}
        \label{fig:appendix_olmo_content}
    \end{figure}

\FloatBarrier

\subsection{Detection: \model{olmo32} training stages}\label{subsec:detection_olmo_stages}
\begin{figure}[ht]
\centering
\tikz\draw[red!25, thick, mark=*, mark size=1.5] plot coordinates {(0,0) (0.4,0)}; {\small Base}\hspace{1em}
\tikz\draw[red!50, thick, mark=*, mark size=1.5] plot coordinates {(0,0) (0.4,0)}; {\small SFT}\hspace{1em}
\tikz\draw[red!75, thick, mark=*, mark size=1.5] plot coordinates {(0,0) (0.4,0)}; {\small DPO}\hspace{1em}
\tikz\draw[red!70!black, thick, mark=*, mark size=1.5] plot coordinates {(0,0) (0.4,0)}; {\small RLVR}
\\[3pt]
\tikzsetnextfilename{detection-olmo-stages}
\begin{tikzpicture}
\begin{groupplot}[
    group style={group size=2 by 2, horizontal sep=0.6cm, vertical sep=1.4cm, yticklabels at=edge left},
    height=3.0cm,
    width=5.6cm,
    scale only axis,
    grid=major,
    xmin=-33, xmax=105,
    xtick={-25,0,20,40,60,80,100},
    ymin=0.0, ymax=100.0,
    ytick={0,50,100},
    tick label style={font=\footnotesize},
]
\nextgroupplot[xlabel={\small Dropout rate $p$}, xticklabels={none,$p_{\min}$,,,,,$p_{\max}$}, title={\small Was a perturbation applied?}, ylabel={\small Answers `Yes' (\%)}]
\addplot[red!25, fill=red!25, fill opacity=0.3, draw=none, forget plot] coordinates {(0,49.08) (10,52.18) (20,50.18) (30,48.48) (40,50.78) (50,49.48) (60,51.88) (70,51.08) (80,52.38) (90,51.08) (100,53.08) (100,49.92) (90,47.92) (80,49.22) (70,47.92) (60,48.72) (50,46.32) (40,47.62) (30,45.32) (20,47.02) (10,49.02) (0,45.92)} --cycle;
\addplot[red!25, mark=*, mark size=1, line width=1pt, forget plot] coordinates {(0,47.50) (10,50.60) (20,48.60) (30,46.90) (40,49.20) (50,47.90) (60,50.30) (70,49.50) (80,50.80) (90,49.50) (100,51.50)};
\addplot[red!25, only marks, mark=*, mark size=1.5, forget plot] coordinates {(-25,47.00)};
\addplot[red!50, fill=red!50, fill opacity=0.3, draw=none, forget plot] coordinates {(0,23.00) (10,27.18) (20,29.12) (30,32.16) (40,37.52) (50,41.35) (60,46.17) (70,49.18) (80,49.58) (90,43.46) (100,38.43) (100,35.37) (90,40.34) (80,46.42) (70,46.02) (60,43.03) (50,38.25) (40,34.48) (30,29.24) (20,26.28) (10,24.42) (0,20.40)} --cycle;
\addplot[red!50, mark=*, mark size=1, line width=1pt, forget plot] coordinates {(0,21.70) (10,25.80) (20,27.70) (30,30.70) (40,36.00) (50,39.80) (60,44.60) (70,47.60) (80,48.00) (90,41.90) (100,36.90)};
\addplot[red!50, only marks, mark=*, mark size=1.5, forget plot] coordinates {(-25,21.40)};
\addplot[red!75, fill=red!75, fill opacity=0.3, draw=none, forget plot] coordinates {(0,10.01) (10,9.59) (20,13.65) (30,17.26) (40,17.57) (50,20.65) (60,21.16) (70,24.53) (80,22.29) (90,19.93) (100,18.39) (100,16.01) (90,17.47) (80,19.71) (70,21.87) (60,18.64) (50,18.15) (40,15.23) (30,14.94) (20,11.55) (10,7.81) (0,8.19)} --cycle;
\addplot[red!75, mark=*, mark size=1, line width=1pt, forget plot] coordinates {(0,9.10) (10,8.70) (20,12.60) (30,16.10) (40,16.40) (50,19.40) (60,19.90) (70,23.20) (80,21.00) (90,18.70) (100,17.20)};
\addplot[red!75, only marks, mark=*, mark size=1.5, forget plot] coordinates {(-25,4.60)};
\addplot[red!70!black, fill=red!70!black, fill opacity=0.3, draw=none, forget plot] coordinates {(0,6.33) (10,6.75) (20,8.02) (30,9.07) (40,14.68) (50,14.48) (60,17.57) (70,17.16) (80,14.89) (90,16.44) (100,14.17) (100,12.03) (90,14.16) (80,12.71) (70,14.84) (60,15.23) (50,12.32) (40,12.52) (30,7.33) (20,6.38) (10,5.25) (0,4.87)} --cycle;
\addplot[red!70!black, mark=*, mark size=1, line width=1pt, forget plot] coordinates {(0,5.60) (10,6.00) (20,7.20) (30,8.20) (40,13.60) (50,13.40) (60,16.40) (70,16.00) (80,13.80) (90,15.30) (100,13.10)};
\addplot[red!70!black, only marks, mark=*, mark size=1.5, forget plot] coordinates {(-25,3.30)};
\nextgroupplot[xlabel={\small Dropout rate $p$}, xticklabels={none,$p_{\min}$,,,,,$p_{\max}$}, title={\small No correct answer}]
\addplot[red!25, fill=red!25, fill opacity=0.3, draw=none, forget plot] coordinates {(0,50.38) (10,54.18) (20,55.48) (30,50.98) (40,50.78) (50,51.78) (60,52.88) (70,55.38) (80,56.77) (90,59.06) (100,62.24) (100,59.16) (90,55.94) (80,53.63) (70,52.22) (60,49.72) (50,48.62) (40,47.62) (30,47.82) (20,52.32) (10,51.02) (0,47.22)} --cycle;
\addplot[red!25, mark=*, mark size=1, line width=1pt, forget plot] coordinates {(0,48.80) (10,52.60) (20,53.90) (30,49.40) (40,49.20) (50,50.20) (60,51.30) (70,53.80) (80,55.20) (90,57.50) (100,60.70)};
\addplot[red!25, only marks, mark=*, mark size=1.5, forget plot] coordinates {(-25,50.80)};
\addplot[red!50, fill=red!50, fill opacity=0.3, draw=none, forget plot] coordinates {(0,62.15) (10,63.83) (20,62.94) (30,64.33) (40,60.56) (50,64.92) (60,64.92) (70,63.53) (80,69.57) (90,66.11) (100,67.50) (100,64.50) (90,63.09) (80,66.63) (70,60.47) (60,61.88) (50,61.88) (40,57.44) (30,61.27) (20,59.86) (10,60.77) (0,59.05)} --cycle;
\addplot[red!50, mark=*, mark size=1, line width=1pt, forget plot] coordinates {(0,60.60) (10,62.30) (20,61.40) (30,62.80) (40,59.00) (50,63.40) (60,63.40) (70,62.00) (80,68.10) (90,64.60) (100,66.00)};
\addplot[red!50, only marks, mark=*, mark size=1.5, forget plot] coordinates {(-25,60.10)};
\addplot[red!75, fill=red!75, fill opacity=0.3, draw=none, forget plot] coordinates {(0,18.09) (10,18.09) (20,19.32) (30,19.11) (40,18.91) (50,18.60) (60,20.04) (70,20.96) (80,21.06) (90,18.19) (100,19.11) (100,16.69) (90,15.81) (80,18.54) (70,18.44) (60,17.56) (50,16.20) (40,16.49) (30,16.69) (20,16.88) (10,15.71) (0,15.71)} --cycle;
\addplot[red!75, mark=*, mark size=1, line width=1pt, forget plot] coordinates {(0,16.90) (10,16.90) (20,18.10) (30,17.90) (40,17.70) (50,17.40) (60,18.80) (70,19.70) (80,19.80) (90,17.00) (100,17.90)};
\addplot[red!75, only marks, mark=*, mark size=1.5, forget plot] coordinates {(-25,16.60)};
\addplot[red!70!black, fill=red!70!black, fill opacity=0.3, draw=none, forget plot] coordinates {(0,25.66) (10,25.45) (20,27.59) (30,26.17) (40,27.18) (50,29.62) (60,28.61) (70,27.79) (80,27.90) (90,27.90) (100,29.12) (100,26.28) (90,25.10) (80,25.10) (70,25.01) (60,25.79) (50,26.78) (40,24.42) (30,23.43) (20,24.81) (10,22.75) (0,22.94)} --cycle;
\addplot[red!70!black, mark=*, mark size=1, line width=1pt, forget plot] coordinates {(0,24.30) (10,24.10) (20,26.20) (30,24.80) (40,25.80) (50,28.20) (60,27.20) (70,26.40) (80,26.50) (90,26.50) (100,27.70)};
\addplot[red!70!black, only marks, mark=*, mark size=1.5, forget plot] coordinates {(-25,20.70)};
\nextgroupplot[xlabel={\small Noise SD $\sigma$}, xticklabels={none,$\sigma_{\min}$,,,,,$\sigma_{\max}$}, ylabel={\small Answers `Yes' (\%)}]
\addplot[red!25, fill=red!25, fill opacity=0.3, draw=none, forget plot] coordinates {(0,51.08) (10,52.38) (20,49.18) (30,53.18) (40,50.98) (50,49.38) (60,51.98) (70,52.58) (80,53.18) (90,51.28) (100,50.08) (100,46.92) (90,48.12) (80,50.02) (70,49.42) (60,48.82) (50,46.22) (40,47.82) (30,50.02) (20,46.02) (10,49.22) (0,47.92)} --cycle;
\addplot[red!25, mark=*, mark size=1, line width=1pt, forget plot] coordinates {(0,49.50) (10,50.80) (20,47.60) (30,51.60) (40,49.40) (50,47.80) (60,50.40) (70,51.00) (80,51.60) (90,49.70) (100,48.50)};
\addplot[red!25, only marks, mark=*, mark size=1.5, forget plot] coordinates {(-25,49.40)};
\addplot[red!50, fill=red!50, fill opacity=0.3, draw=none, forget plot] coordinates {(0,23.00) (10,28.81) (20,27.79) (30,36.10) (40,41.25) (50,53.08) (60,60.26) (70,61.25) (80,63.14) (90,60.56) (100,54.98) (100,51.82) (90,57.44) (80,60.06) (70,58.15) (60,57.14) (50,49.92) (40,38.15) (30,33.10) (20,25.01) (10,25.99) (0,20.40)} --cycle;
\addplot[red!50, mark=*, mark size=1, line width=1pt, forget plot] coordinates {(0,21.70) (10,27.40) (20,26.40) (30,34.60) (40,39.70) (50,51.50) (60,58.70) (70,59.70) (80,61.60) (90,59.00) (100,53.40)};
\addplot[red!50, only marks, mark=*, mark size=1.5, forget plot] coordinates {(-25,19.80)};
\addplot[red!75, fill=red!75, fill opacity=0.3, draw=none, forget plot] coordinates {(0,5.90) (10,5.69) (20,10.01) (30,13.23) (40,19.01) (50,25.35) (60,29.52) (70,31.65) (80,30.94) (90,30.33) (100,30.94) (100,28.06) (90,27.47) (80,28.06) (70,28.75) (60,26.68) (50,22.65) (40,16.59) (30,11.17) (20,8.19) (10,4.31) (0,4.50)} --cycle;
\addplot[red!75, mark=*, mark size=1, line width=1pt, forget plot] coordinates {(0,5.20) (10,5.00) (20,9.10) (30,12.20) (40,17.80) (50,24.00) (60,28.10) (70,30.20) (80,29.50) (90,28.90) (100,29.50)};
\addplot[red!75, only marks, mark=*, mark size=1.5, forget plot] coordinates {(-25,3.50)};
\addplot[red!70!black, fill=red!70!black, fill opacity=0.3, draw=none, forget plot] coordinates {(0,2.77) (10,4.40) (20,4.83) (30,8.54) (40,15.10) (50,19.32) (60,24.43) (70,26.88) (80,25.76) (90,25.15) (100,25.96) (100,23.24) (90,22.45) (80,23.04) (70,24.12) (60,21.77) (50,16.88) (40,12.90) (30,6.86) (20,3.57) (10,3.20) (0,1.83)} --cycle;
\addplot[red!70!black, mark=*, mark size=1, line width=1pt, forget plot] coordinates {(0,2.30) (10,3.80) (20,4.20) (30,7.70) (40,14.00) (50,18.10) (60,23.10) (70,25.50) (80,24.40) (90,23.80) (100,24.60)};
\addplot[red!70!black, only marks, mark=*, mark size=1.5, forget plot] coordinates {(-25,2.80)};
\nextgroupplot[xlabel={\small Noise SD $\sigma$}, xticklabels={none,$\sigma_{\min}$,,,,,$\sigma_{\max}$}]
\addplot[red!25, fill=red!25, fill opacity=0.3, draw=none, forget plot] coordinates {(0,50.78) (10,51.58) (20,55.38) (30,52.78) (40,55.58) (50,53.58) (60,52.18) (70,53.58) (80,56.37) (90,54.58) (100,55.68) (100,52.52) (90,51.42) (80,53.23) (70,50.42) (60,49.02) (50,50.42) (40,52.42) (30,49.62) (20,52.22) (10,48.42) (0,47.62)} --cycle;
\addplot[red!25, mark=*, mark size=1, line width=1pt, forget plot] coordinates {(0,49.20) (10,50.00) (20,53.80) (30,51.20) (40,54.00) (50,52.00) (60,50.60) (70,52.00) (80,54.80) (90,53.00) (100,54.10)};
\addplot[red!25, only marks, mark=*, mark size=1.5, forget plot] coordinates {(-25,47.60)};
\addplot[red!50, fill=red!50, fill opacity=0.3, draw=none, forget plot] coordinates {(0,62.94) (10,64.63) (20,62.94) (30,65.72) (40,64.43) (50,71.74) (60,67.20) (70,69.87) (80,67.70) (90,69.97) (100,70.96) (100,68.04) (90,67.03) (80,64.70) (70,66.93) (60,64.20) (50,68.86) (40,61.37) (30,62.68) (20,59.86) (10,61.57) (0,59.86)} --cycle;
\addplot[red!50, mark=*, mark size=1, line width=1pt, forget plot] coordinates {(0,61.40) (10,63.10) (20,61.40) (30,64.20) (40,62.90) (50,70.30) (60,65.70) (70,68.40) (80,66.20) (90,68.50) (100,69.50)};
\addplot[red!50, only marks, mark=*, mark size=1.5, forget plot] coordinates {(-25,59.60)};
\addplot[red!75, fill=red!75, fill opacity=0.3, draw=none, forget plot] coordinates {(0,16.23) (10,19.42) (20,18.91) (30,23.00) (40,17.88) (50,19.21) (60,18.60) (70,20.75) (80,20.45) (90,17.37) (100,17.88) (100,15.52) (90,15.03) (80,17.95) (70,18.25) (60,16.20) (50,16.79) (40,15.52) (30,20.40) (20,16.49) (10,16.98) (0,13.97)} --cycle;
\addplot[red!75, mark=*, mark size=1, line width=1pt, forget plot] coordinates {(0,15.10) (10,18.20) (20,17.70) (30,21.70) (40,16.70) (50,18.00) (60,17.40) (70,19.50) (80,19.20) (90,16.20) (100,16.70)};
\addplot[red!75, only marks, mark=*, mark size=1.5, forget plot] coordinates {(-25,15.30)};
\addplot[red!70!black, fill=red!70!black, fill opacity=0.3, draw=none, forget plot] coordinates {(0,23.72) (10,23.72) (20,24.33) (30,25.45) (40,26.27) (50,27.49) (60,28.91) (70,29.62) (80,32.87) (90,30.84) (100,30.54) (100,27.66) (90,27.96) (80,29.93) (70,26.78) (60,26.09) (50,24.71) (40,23.53) (30,22.75) (20,21.67) (10,21.08) (0,21.08)} --cycle;
\addplot[red!70!black, mark=*, mark size=1, line width=1pt, forget plot] coordinates {(0,22.40) (10,22.40) (20,23.00) (30,24.10) (40,24.90) (50,26.10) (60,27.50) (70,28.20) (80,31.40) (90,29.40) (100,29.10)};
\addplot[red!70!black, only marks, mark=*, mark size=1.5, forget plot] coordinates {(-25,24.20)};
\end{groupplot}
\end{tikzpicture}
\vspace{-8pt}
\caption{\small The training stages of \model{olmo32} are presented with one sentence, perturbed with either \textbf{dropout} (top) or \textbf{Gaussian noise} (bottom) at the strengths of each stage's own range (Table~\ref{tab:range_olmo_stages}), and asked whether a perturbation was applied (\textbf{left}) or a question without a correct answer (\textbf{right}). After DPO, the frequency of `Yes' drops for both kinds of question. Standard errors are below $1.58\%$ and hence difficult to visualize.}
\label{fig:detection_olmo_stages}
\end{figure}

    Why does \model{olmo32} rarely report a perturbation, although it localizes it with high accuracy (Fig.~\ref{fig:loc_main_15tok})?
    One may suspect that the model learns to deny perturbations during post-training, \eg, because safety tuning discourages self reports.
    To test this hypothesis, we repeat the detection experiment and its controls on every training stage of \model{olmo32} (\cf. \S\ref{subsec:loc_emergence}). To this aim, we first repeat the experiment ``localization'', as well as its associated control, for every training stage of Olmo (Fig.~\ref{fig:appendix_olmo_stage_ranges}), so as to deduce the ranges of the perturbation strengths as in \S\ref{subsec:perturb_lower_bounds} and \S\ref{subsec:perturb_upper_bounds}.
    We report them in Table~\ref{tab:range_olmo_stages}, and the results in Fig.~\ref{fig:detection_olmo_stages}.

    \textbf{The base model.}
    The base version of \model{olmo32} answers at chance, with or without perturbation, as it does when the question has no correct answer.
    Yet it answers correctly the questions about the content of the sentence and it localizes the perturbed sentence with almost perfect accuracy (Fig.~\ref{fig:appendix_olmo_stage_ranges}).

    \textbf{Post-training.}
    After SFT, the frequency of `Yes' increases with either perturbation strength.
    DPO lowers both curves, and RLVR lowers them a little further. 
    However, after DPO the frequency of `Yes' drops even more when the question has no correct answer, while it does not change when the answer is `Yes'.
    Hence, DPO makes \model{olmo32} answer `No' more often whenever the question has no clear answer, and not only when we ask about the perturbation.
    (In the parlance of signal detection theory, DPO raises the criterion of \model{olmo32} without lowering its sensitivity.) 

    \begin{table}[ht]
    \centering
    \small
        \begin{tabular}{lcc}
            \toprule
            Stage & $(p_{\min}, p_{\max})$ & $(\sigma_{\min}, \sigma_{\max})$ \\
            \midrule
            Base & (.06, .72) & (.01, .08) \\
            SFT  & (.06, .64) & (.02, .07) \\
            DPO  & (.26, .66) & (.02, .08) \\
            RLVR & (.26, .66) & (.02, .08) \\
            \bottomrule
        \end{tabular}
    \caption{\small Ranges of the perturbation strengths of the training stages of \model{olmo32}; RLVR is the final stage, whose ranges are those of Table~\ref{tab:range_perturbation}.}
    \label{tab:range_olmo_stages}
    \end{table}

    \input{plots/11_appendix/detection/olmo_stage_ranges}

\subsection{Zero-shot classification: control labels}\label{subsec:appendix_zero_shot_controls}

    The double monotonic increase of \model{qwen32} and \model{qwen27}'s recall in the zero-shot classification  (\S\ref{sec:zero_shot}) could be explained by dropout (resp.\ Gaussian noise) steering the activations in a way that systematically favors saying ``dropout'' (resp.\ noise), irrespective of the semantic content of our question. If this is the case, we would expect to see similar trends if, while still applying dropout or noise, the question asked would be different, such as ``\emph{Which perturbation did we apply, X or Y?}''. To control for this, we test \model{qwen32} and \model{qwen27} with \textbf{51 pairs of control labels}. For instance, we ask the model (paraphrasing): ``Did we apply X or Y''.
    In these controls we then keep track of the number of times the model says ``X'' (resp.\ ``Y'') when the underlying perturbation is dropout (resp.\ noise), or vice versa. 
    Most of the control labels are semantically unrelated to dropout and noise, 
    but we also use pseudo-synonyms of dropout and noise such as 
    ``Masking'' 
    and 
    ``Jitter'', and their permutations. See the full list of controls in Table~\ref{tab:control_labels}.

    \begin{table}[ht]
    \centering
    \small
        \begin{tabular}{@{}lp{0.78\linewidth}@{}}
        \toprule
        \textbf{Category} & \textbf{Labels} \\
        \midrule
        Correct        & \textit{Dropout/Noise} \\
        \midrule
        Synonyms       & \textit{Masking/Jitter}, \textit{Dropout/Jitter}, \textit{Masking/Noise}, \textit{Zeroing/Gaussian} \\
        \midrule
        Placeholders   & \textit{X/Y}, \textit{Foo/Bar} \\
        \midrule
        Perturbations  & \textit{Rotation/Permutation}, \textit{Scaling/Translation}, \textit{Dropout/Quantization}, \textit{Quantization/Noise}, \textit{Suppression/Injection}, \textit{Blanking/Fuzzing}, \textit{Clipping/Smoothing}, \textit{Steering/Clamping}, \textit{Freezing/Unfreezing}, \textit{Masking/Quantization}, \textit{Quantization/Jitter} \\
        \midrule
        Nonsense       & \textit{Dorvane/Kenlo}, \textit{Bralto/Sivek}, \textit{Quelp/Frando}, \textit{Zinth/Morble} \\
        \midrule
        Encodings      & \textit{64726f706f7574/6e6f697365} (hex for ``dropout''/``noise''), \textit{Tuopord/Esion} (reversed) \\
        \midrule
        Food \& drink  & \textit{Beer/Wine}, \textit{Pasta/Pizza} \\
        \midrule
        Properties     & \textit{Plain/Rich}, \textit{Still/Sparkling}, \textit{Weak/Strong}, \textit{Smooth/Rough}, \textit{Light/Dark}, \textit{Hot/Cold}, \textit{Fast/Slow}, \textit{Wet/Dry}, \textit{Sharp/Blunt}, \textit{Tall/Short} \\
        \midrule
        Nature         & \textit{Ocean/Mountain}, \textit{Fire/Water}, \textit{Sun/Moon}, \textit{Forest/Desert}, \textit{River/Lake} \\
        \midrule
        Fiction \& gaming & \textit{Jedi/Sith}, \textit{Linux/Windows}, \textit{PlayStation/Xbox}, \textit{Groudon/Kyogre}, \textit{Stark/Lannister}, \textit{Vim/Emacs}, \textit{Mario/Sonic}, \textit{Zelda/Metroid}, \textit{Cat/Dog}, \textit{Vampire/Werewolf} \\
        \midrule
        Unrelated      & \textit{Vanilla/Chocolate} \\
        \bottomrule
        \end{tabular}
    \caption{Zero-shot classification: controls labels for \model{qwen32} and \model{qwen27}.}
    \label{tab:control_labels}
    \end{table}

    Because \model{gemma31}, \model{llama8}, \model{mistral24}, \model{olmo32}, \model{qwen14}, and \model{seed36} did not display a similarly convincing behavior at the zero-shot task, we tested only a few control labels: \textit{X/Y}, \textit{Foo/Bar},  \textit{Masking/Jitter}, \textit{Rotation/Permutation}, \textit{Scaling/Translation}, \textit{Vanilla/Chocolate}.
    Fig.~\ref{fig:zero_shot_controls} reports eight of these controls for \model{qwen32} and \model{qwen27}.
    \input{plots/11_appendix/zero_shot/controls_qwen32_qwen27}

\subsection{Zero-shot classification: accuracy of every model}\label{subsec:appendix_zero_shot_results}
    In Fig.~\ref{fig:zero_shot_accuracy_all_models}, we report the accuracy of every model at the zero-shot classification task (\S\ref{sec:zero_shot}), as a function of dropout rate and noise SD.
    \input{plots/11_appendix/zero_shot/accuracy_all_models}

\subsection{Zero-shot classification: entropy}\label{subsec:appendix_zero_shot_entropy}
    Accuracy in the zero-shot classification experiment (\S\ref{sec:zero_shot}) might be polluted by the fact that applying dropout (resp.\ noise) increases the probability mass of every token in a systematic way. Therefore, in Fig.~\ref{fig:zero_shot_entropy_all_models}, we report the entropy of the whole distribution of tokens, so as to maintain a global view on what applying dropout (resp.\ adding noise) causes.

    \input{plots/11_appendix/zero_shot/entropy_all_models}

\FloatBarrier

\subsection{Few-shot classification: heatmaps of all models}\label{subsec:appendix_few_shot_heatmaps}

    We report the heatmaps displaying accuracy at the few-shot classification task for every model. In particular, we show the accuracy of each model as a function of dropout rate, noise SD, when the model sees 1 (Figs.~\ref{fig:few_shot_accuracy_all_models} and~\ref{fig:few_shot_accuracy_all_models_b}) or 9 (Figs.~\ref{fig:few_shot_accuracy_all_models_9_ex} and~\ref{fig:few_shot_accuracy_all_models_9_ex_b}) in-context example pairs. 

    \begin{figure}[ht]
        \centering
        \scalebox{0.8}{\begin{minipage}{\textwidth}\centering
            \begin{minipage}[t]{0.48\textwidth}
                \input{plots/11_appendix/few_shot/accuracy/gemma31_k1}
            \end{minipage}%
            \hfill
            \begin{minipage}[t]{0.48\textwidth}
                \input{plots/11_appendix/few_shot/accuracy/llama8_k1}
            \end{minipage}\\[0pt]
            \begin{minipage}[t]{0.48\textwidth}
                \input{plots/11_appendix/few_shot/accuracy/mistral24_k1}
            \end{minipage}%
            \hfill
            \begin{minipage}[t]{0.48\textwidth}
                \input{plots/11_appendix/few_shot/accuracy/olmo32_k1}
            \end{minipage}
        \end{minipage}}
            \vspace{-5pt}
        \caption{\small
            Accuracy of every model at distinguishing dropout from Gaussian noise when taught with $1$ in-context example. Standard errors are below $1.58\%$ everywhere.
        }
        \vspace{-8pt}
        \label{fig:few_shot_accuracy_all_models}
    \end{figure}

    \begin{figure}[ht]
        \centering
        \scalebox{0.8}{\begin{minipage}{\textwidth}\centering
            \begin{minipage}[t]{0.48\textwidth}
                \input{plots/11_appendix/few_shot/accuracy/qwen14_k1}
            \end{minipage}%
            \hfill
            \begin{minipage}[t]{0.48\textwidth}
                \input{plots/11_appendix/few_shot/accuracy/qwen32_k1}
            \end{minipage}\\[0pt]
            \begin{minipage}[t]{0.48\textwidth}
                \input{plots/11_appendix/few_shot/accuracy/qwen27_k1}
            \end{minipage}%
            \hfill
            \begin{minipage}[t]{0.48\textwidth}
                \input{plots/11_appendix/few_shot/accuracy/seed36_k1}
            \end{minipage}
        \end{minipage}}
            \vspace{-5pt}
        \caption{\small
            Accuracy of every model at distinguishing dropout from Gaussian noise when taught with $1$ in-context example. Standard errors are below $1.58\%$ everywhere.
        }
        \vspace{-8pt}
        \label{fig:few_shot_accuracy_all_models_b}
    \end{figure}

    \begin{figure}[ht]
        \centering
        \scalebox{0.8}{\begin{minipage}{\textwidth}\centering
            \begin{minipage}[t]{0.48\textwidth}
                \input{plots/11_appendix/few_shot/accuracy/gemma31_k9}
            \end{minipage}%
            \hfill
            \begin{minipage}[t]{0.48\textwidth}
                \input{plots/11_appendix/few_shot/accuracy/llama8_k9}
            \end{minipage}\\[0pt]
            \begin{minipage}[t]{0.48\textwidth}
                \input{plots/11_appendix/few_shot/accuracy/mistral24_k9}
            \end{minipage}%
            \hfill
            \begin{minipage}[t]{0.48\textwidth}
                \input{plots/11_appendix/few_shot/accuracy/olmo32_k9}
            \end{minipage}
        \end{minipage}}
            \vspace{-5pt}
        \caption{\small
            Accuracy of every model at distinguishing dropout from Gaussian noise when taught with $9$ in-context examples. Standard errors are below $1.58\%$ everywhere.
        }
        \vspace{-8pt}
        \label{fig:few_shot_accuracy_all_models_9_ex}
    \end{figure}

    \begin{figure}[ht]
        \centering
        \scalebox{0.8}{\begin{minipage}{\textwidth}\centering
            \begin{minipage}[t]{0.48\textwidth}
                \input{plots/11_appendix/few_shot/accuracy/qwen14_k9}
            \end{minipage}%
            \hfill
            \begin{minipage}[t]{0.48\textwidth}
                \input{plots/11_appendix/few_shot/accuracy/qwen32_k9}
            \end{minipage}\\[0pt]
            \begin{minipage}[t]{0.48\textwidth}
                \input{plots/11_appendix/few_shot/accuracy/qwen27_k9}
            \end{minipage}%
            \hfill
            \begin{minipage}[t]{0.48\textwidth}
                \input{plots/11_appendix/few_shot/accuracy/seed36_k9}
            \end{minipage}
        \end{minipage}}
            \vspace{-5pt}
        \caption{\small
            Accuracy of every model at distinguishing dropout from Gaussian noise when taught with $9$ in-context examples. Standard errors are below $1.58\%$ everywhere.
        }
        \vspace{-8pt}
        \label{fig:few_shot_accuracy_all_models_9_ex_b}
    \end{figure}

\FloatBarrier

\subsection{Few-shot classification: delta heatmaps}\label{subsec:appendix_delta_heatmaps}
    For every model other than \model{qwen32} (Figure~\ref{fig:icl_deltas_qwen32}), we report in Figs.~\ref{fig:few_shot_delta_aliases_gemma31}--\ref{fig:few_shot_delta_aliases_seed36} the heatmaps with the delta accuracy between the correct labels minus the swapped labels, for the dropout / noise experiment and for the control labels.

    \FloatBarrier

    \begin{figure}[ht]
        \centering
        \scalebox{0.77}{\begin{minipage}{\textwidth}\centering
            \begin{minipage}[t]{0.48\textwidth}
                \input{plots/11_appendix/few_shot/delta/gemma31_dropout_noise}
            \end{minipage}%
            \hfill
            \begin{minipage}[t]{0.48\textwidth}
                \input{plots/11_appendix/few_shot/delta/gemma31_masking_jitter}
            \end{minipage}\\[0pt]
            \begin{minipage}[t]{0.48\textwidth}
                \centering
\begin{tikzpicture}
\begin{axis}[
    width=0.82\linewidth,
    height=0.82\linewidth,
    xlabel={\small Dropout rate $p$},
    ylabel={\small Noise SD $\sigma$},
    colormap={pastel_po}{rgb255(0cm)=(225,160,100); rgb255(50cm)=(255,255,255); rgb255(100cm)=(160,135,200)},
    point meta min=-25.0,
    point meta max=25.0,
    xtick={0,1,2,3,4,5,6,7,8,9,10},
    xticklabels={$p_{\min}$,,,,,,,,,,$p_{\max}$},
    ytick={0,1,2,3,4,5,6,7,8,9,10},
    yticklabels={$\sigma_{\min}$,,,,,,,,,,$\sigma_{\max}$},
    x tick label style={font=\scriptsize},
    y tick label style={font=\scriptsize},
    view={0}{90},
    axis line style={draw=none},
    major tick length=0pt,
    enlargelimits=false,
    title style={font=\small},
    title={X/Y},
    title style={yshift=-1.0ex},
]
\addplot[matrix plot*, mesh/cols=11, mesh/rows=11, point meta=explicit] table[meta=C] {
x y C
0 10 5.50
1 10 8.20
2 10 1.50
3 10 6.60
4 10 8.10
5 10 9.20
6 10 9.10
7 10 7.30
8 10 3.90
9 10 -0.30
10 10 3.50
0 9 6.60
1 9 5.80
2 9 4.80
3 9 6.20
4 9 7.40
5 9 6.60
6 9 8.90
7 9 7.60
8 9 2.50
9 9 6.50
10 9 -0.40
0 8 4.40
1 8 6.30
2 8 2.30
3 8 3.20
4 8 5.20
5 8 5.80
6 8 9.10
7 8 1.60
8 8 -0.10
9 8 0.10
10 8 3.70
0 7 -0.50
1 7 4.50
2 7 7.00
3 7 3.60
4 7 6.30
5 7 7.70
6 7 5.20
7 7 0.20
8 7 2.60
9 7 -1.00
10 7 0.80
0 6 3.20
1 6 2.50
2 6 1.40
3 6 2.20
4 6 3.20
5 6 5.80
6 6 1.30
7 6 1.10
8 6 2.20
9 6 1.30
10 6 0.80
0 5 3.30
1 5 3.70
2 5 -0.60
3 5 1.50
4 5 -1.90
5 5 -0.10
6 5 0.10
7 5 1.60
8 5 0.00
9 5 -2.80
10 5 -2.40
0 4 2.80
1 4 2.50
2 4 4.60
3 4 0.50
4 4 -3.20
5 4 4.30
6 4 -2.60
7 4 0.20
8 4 -2.90
9 4 2.10
10 4 3.00
0 3 6.80
1 3 0.20
2 3 3.90
3 3 6.00
4 3 2.40
5 3 -1.10
6 3 -1.00
7 3 -2.80
8 3 -3.90
9 3 4.10
10 3 0.30
0 2 8.00
1 2 2.30
2 2 3.50
3 2 0.00
4 2 -3.60
5 2 0.30
6 2 -3.20
7 2 -5.40
8 2 -1.70
9 2 -2.30
10 2 2.10
0 1 -1.80
1 1 0.00
2 1 -2.00
3 1 0.60
4 1 1.80
5 1 -1.30
6 1 -3.00
7 1 -1.70
8 1 -6.60
9 1 -2.90
10 1 -0.50
0 0 -2.30
1 0 -0.70
2 0 -1.10
3 0 -4.30
4 0 -0.50
5 0 0.20
6 0 -2.60
7 0 -3.50
8 0 -5.40
9 0 -1.70
10 0 -0.50
};
\node[font=\fontsize{4}{4.8}\selectfont] at (axis cs:0,0) {-2.3};
\node[font=\fontsize{4}{4.8}\selectfont] at (axis cs:1,0) {-0.7};
\node[font=\fontsize{4}{4.8}\selectfont] at (axis cs:2,0) {-1.1};
\node[font=\fontsize{4}{4.8}\selectfont] at (axis cs:3,0) {-4.3};
\node[font=\fontsize{4}{4.8}\selectfont] at (axis cs:4,0) {-0.5};
\node[font=\fontsize{4}{4.8}\selectfont] at (axis cs:5,0) {0.2};
\node[font=\fontsize{4}{4.8}\selectfont] at (axis cs:6,0) {-2.6};
\node[font=\fontsize{4}{4.8}\selectfont] at (axis cs:7,0) {-3.5};
\node[font=\fontsize{4}{4.8}\selectfont] at (axis cs:8,0) {-5.4};
\node[font=\fontsize{4}{4.8}\selectfont] at (axis cs:9,0) {-1.7};
\node[font=\fontsize{4}{4.8}\selectfont] at (axis cs:10,0) {-0.5};
\node[font=\fontsize{4}{4.8}\selectfont] at (axis cs:0,1) {-1.8};
\node[font=\fontsize{4}{4.8}\selectfont] at (axis cs:1,1) {0.0};
\node[font=\fontsize{4}{4.8}\selectfont] at (axis cs:2,1) {-2.0};
\node[font=\fontsize{4}{4.8}\selectfont] at (axis cs:3,1) {0.6};
\node[font=\fontsize{4}{4.8}\selectfont] at (axis cs:4,1) {1.8};
\node[font=\fontsize{4}{4.8}\selectfont] at (axis cs:5,1) {-1.3};
\node[font=\fontsize{4}{4.8}\selectfont] at (axis cs:6,1) {-3.0};
\node[font=\fontsize{4}{4.8}\selectfont] at (axis cs:7,1) {-1.7};
\node[font=\fontsize{4}{4.8}\selectfont] at (axis cs:8,1) {-6.6};
\node[font=\fontsize{4}{4.8}\selectfont] at (axis cs:9,1) {-2.9};
\node[font=\fontsize{4}{4.8}\selectfont] at (axis cs:10,1) {-0.5};
\node[font=\fontsize{4}{4.8}\selectfont] at (axis cs:0,2) {8.0};
\node[font=\fontsize{4}{4.8}\selectfont] at (axis cs:1,2) {2.3};
\node[font=\fontsize{4}{4.8}\selectfont] at (axis cs:2,2) {3.5};
\node[font=\fontsize{4}{4.8}\selectfont] at (axis cs:3,2) {0.0};
\node[font=\fontsize{4}{4.8}\selectfont] at (axis cs:4,2) {-3.6};
\node[font=\fontsize{4}{4.8}\selectfont] at (axis cs:5,2) {0.3};
\node[font=\fontsize{4}{4.8}\selectfont] at (axis cs:6,2) {-3.2};
\node[font=\fontsize{4}{4.8}\selectfont] at (axis cs:7,2) {-5.4};
\node[font=\fontsize{4}{4.8}\selectfont] at (axis cs:8,2) {-1.7};
\node[font=\fontsize{4}{4.8}\selectfont] at (axis cs:9,2) {-2.3};
\node[font=\fontsize{4}{4.8}\selectfont] at (axis cs:10,2) {2.1};
\node[font=\fontsize{4}{4.8}\selectfont] at (axis cs:0,3) {6.8};
\node[font=\fontsize{4}{4.8}\selectfont] at (axis cs:1,3) {0.2};
\node[font=\fontsize{4}{4.8}\selectfont] at (axis cs:2,3) {3.9};
\node[font=\fontsize{4}{4.8}\selectfont] at (axis cs:3,3) {6.0};
\node[font=\fontsize{4}{4.8}\selectfont] at (axis cs:4,3) {2.4};
\node[font=\fontsize{4}{4.8}\selectfont] at (axis cs:5,3) {-1.1};
\node[font=\fontsize{4}{4.8}\selectfont] at (axis cs:6,3) {-1.0};
\node[font=\fontsize{4}{4.8}\selectfont] at (axis cs:7,3) {-2.8};
\node[font=\fontsize{4}{4.8}\selectfont] at (axis cs:8,3) {-3.9};
\node[font=\fontsize{4}{4.8}\selectfont] at (axis cs:9,3) {4.1};
\node[font=\fontsize{4}{4.8}\selectfont] at (axis cs:10,3) {0.3};
\node[font=\fontsize{4}{4.8}\selectfont] at (axis cs:0,4) {2.8};
\node[font=\fontsize{4}{4.8}\selectfont] at (axis cs:1,4) {2.5};
\node[font=\fontsize{4}{4.8}\selectfont] at (axis cs:2,4) {4.6};
\node[font=\fontsize{4}{4.8}\selectfont] at (axis cs:3,4) {0.5};
\node[font=\fontsize{4}{4.8}\selectfont] at (axis cs:4,4) {-3.2};
\node[font=\fontsize{4}{4.8}\selectfont] at (axis cs:5,4) {4.3};
\node[font=\fontsize{4}{4.8}\selectfont] at (axis cs:6,4) {-2.6};
\node[font=\fontsize{4}{4.8}\selectfont] at (axis cs:7,4) {0.2};
\node[font=\fontsize{4}{4.8}\selectfont] at (axis cs:8,4) {-2.9};
\node[font=\fontsize{4}{4.8}\selectfont] at (axis cs:9,4) {2.1};
\node[font=\fontsize{4}{4.8}\selectfont] at (axis cs:10,4) {3.0};
\node[font=\fontsize{4}{4.8}\selectfont] at (axis cs:0,5) {3.3};
\node[font=\fontsize{4}{4.8}\selectfont] at (axis cs:1,5) {3.7};
\node[font=\fontsize{4}{4.8}\selectfont] at (axis cs:2,5) {-0.6};
\node[font=\fontsize{4}{4.8}\selectfont] at (axis cs:3,5) {1.5};
\node[font=\fontsize{4}{4.8}\selectfont] at (axis cs:4,5) {-1.9};
\node[font=\fontsize{4}{4.8}\selectfont] at (axis cs:5,5) {-0.1};
\node[font=\fontsize{4}{4.8}\selectfont] at (axis cs:6,5) {0.1};
\node[font=\fontsize{4}{4.8}\selectfont] at (axis cs:7,5) {1.6};
\node[font=\fontsize{4}{4.8}\selectfont] at (axis cs:8,5) {0.0};
\node[font=\fontsize{4}{4.8}\selectfont] at (axis cs:9,5) {-2.8};
\node[font=\fontsize{4}{4.8}\selectfont] at (axis cs:10,5) {-2.4};
\node[font=\fontsize{4}{4.8}\selectfont] at (axis cs:0,6) {3.2};
\node[font=\fontsize{4}{4.8}\selectfont] at (axis cs:1,6) {2.5};
\node[font=\fontsize{4}{4.8}\selectfont] at (axis cs:2,6) {1.4};
\node[font=\fontsize{4}{4.8}\selectfont] at (axis cs:3,6) {2.2};
\node[font=\fontsize{4}{4.8}\selectfont] at (axis cs:4,6) {3.2};
\node[font=\fontsize{4}{4.8}\selectfont] at (axis cs:5,6) {5.8};
\node[font=\fontsize{4}{4.8}\selectfont] at (axis cs:6,6) {1.3};
\node[font=\fontsize{4}{4.8}\selectfont] at (axis cs:7,6) {1.1};
\node[font=\fontsize{4}{4.8}\selectfont] at (axis cs:8,6) {2.2};
\node[font=\fontsize{4}{4.8}\selectfont] at (axis cs:9,6) {1.3};
\node[font=\fontsize{4}{4.8}\selectfont] at (axis cs:10,6) {0.8};
\node[font=\fontsize{4}{4.8}\selectfont] at (axis cs:0,7) {-0.5};
\node[font=\fontsize{4}{4.8}\selectfont] at (axis cs:1,7) {4.5};
\node[font=\fontsize{4}{4.8}\selectfont] at (axis cs:2,7) {7.0};
\node[font=\fontsize{4}{4.8}\selectfont] at (axis cs:3,7) {3.6};
\node[font=\fontsize{4}{4.8}\selectfont] at (axis cs:4,7) {6.3};
\node[font=\fontsize{4}{4.8}\selectfont] at (axis cs:5,7) {7.7};
\node[font=\fontsize{4}{4.8}\selectfont] at (axis cs:6,7) {5.2};
\node[font=\fontsize{4}{4.8}\selectfont] at (axis cs:7,7) {0.2};
\node[font=\fontsize{4}{4.8}\selectfont] at (axis cs:8,7) {2.6};
\node[font=\fontsize{4}{4.8}\selectfont] at (axis cs:9,7) {-1.0};
\node[font=\fontsize{4}{4.8}\selectfont] at (axis cs:10,7) {0.8};
\node[font=\fontsize{4}{4.8}\selectfont] at (axis cs:0,8) {4.4};
\node[font=\fontsize{4}{4.8}\selectfont] at (axis cs:1,8) {6.3};
\node[font=\fontsize{4}{4.8}\selectfont] at (axis cs:2,8) {2.3};
\node[font=\fontsize{4}{4.8}\selectfont] at (axis cs:3,8) {3.2};
\node[font=\fontsize{4}{4.8}\selectfont] at (axis cs:4,8) {5.2};
\node[font=\fontsize{4}{4.8}\selectfont] at (axis cs:5,8) {5.8};
\node[font=\fontsize{4}{4.8}\selectfont] at (axis cs:6,8) {9.1};
\node[font=\fontsize{4}{4.8}\selectfont] at (axis cs:7,8) {1.6};
\node[font=\fontsize{4}{4.8}\selectfont] at (axis cs:8,8) {-0.1};
\node[font=\fontsize{4}{4.8}\selectfont] at (axis cs:9,8) {0.1};
\node[font=\fontsize{4}{4.8}\selectfont] at (axis cs:10,8) {3.7};
\node[font=\fontsize{4}{4.8}\selectfont] at (axis cs:0,9) {6.6};
\node[font=\fontsize{4}{4.8}\selectfont] at (axis cs:1,9) {5.8};
\node[font=\fontsize{4}{4.8}\selectfont] at (axis cs:2,9) {4.8};
\node[font=\fontsize{4}{4.8}\selectfont] at (axis cs:3,9) {6.2};
\node[font=\fontsize{4}{4.8}\selectfont] at (axis cs:4,9) {7.4};
\node[font=\fontsize{4}{4.8}\selectfont] at (axis cs:5,9) {6.6};
\node[font=\fontsize{4}{4.8}\selectfont] at (axis cs:6,9) {8.9};
\node[font=\fontsize{4}{4.8}\selectfont] at (axis cs:7,9) {7.6};
\node[font=\fontsize{4}{4.8}\selectfont] at (axis cs:8,9) {2.5};
\node[font=\fontsize{4}{4.8}\selectfont] at (axis cs:9,9) {6.5};
\node[font=\fontsize{4}{4.8}\selectfont] at (axis cs:10,9) {-0.4};
\node[font=\fontsize{4}{4.8}\selectfont] at (axis cs:0,10) {5.5};
\node[font=\fontsize{4}{4.8}\selectfont] at (axis cs:1,10) {8.2};
\node[font=\fontsize{4}{4.8}\selectfont] at (axis cs:2,10) {1.5};
\node[font=\fontsize{4}{4.8}\selectfont] at (axis cs:3,10) {6.6};
\node[font=\fontsize{4}{4.8}\selectfont] at (axis cs:4,10) {8.1};
\node[font=\fontsize{4}{4.8}\selectfont] at (axis cs:5,10) {9.2};
\node[font=\fontsize{4}{4.8}\selectfont] at (axis cs:6,10) {9.1};
\node[font=\fontsize{4}{4.8}\selectfont] at (axis cs:7,10) {7.3};
\node[font=\fontsize{4}{4.8}\selectfont] at (axis cs:8,10) {3.9};
\node[font=\fontsize{4}{4.8}\selectfont] at (axis cs:9,10) {-0.3};
\node[font=\fontsize{4}{4.8}\selectfont] at (axis cs:10,10) {3.5};
\end{axis}
\end{tikzpicture}
            \end{minipage}%
            \hfill
            \begin{minipage}[t]{0.48\textwidth}
                \input{plots/11_appendix/few_shot/delta/gemma31_rotation_permutation}
            \end{minipage}
        \end{minipage}}
            \vspace{-5pt}
        \caption{\small
            Same as Figure~\ref{fig:icl_deltas_qwen32}, for \model{gemma31}.
        }
        \vspace{-8pt}
        \label{fig:few_shot_delta_aliases_gemma31}
    \end{figure}
    \FloatBarrier

    \begin{figure}[ht]
        \centering
        \scalebox{0.77}{\begin{minipage}{\textwidth}\centering
            \begin{minipage}[t]{0.48\textwidth}
                \input{plots/11_appendix/few_shot/delta/llama8_dropout_noise}
            \end{minipage}%
            \hfill
            \begin{minipage}[t]{0.48\textwidth}
                \input{plots/11_appendix/few_shot/delta/llama8_masking_jitter}
            \end{minipage}\\[0pt]
            \begin{minipage}[t]{0.48\textwidth}
                \input{plots/11_appendix/few_shot/delta/llama8_x_y}
            \end{minipage}%
            \hfill
            \begin{minipage}[t]{0.48\textwidth}
                \centering
\begin{tikzpicture}
\begin{axis}[
    width=0.82\linewidth,
    height=0.82\linewidth,
    xlabel={\small Dropout rate $p$},
    ylabel={},
    yticklabels={},
    colorbar,
    colormap={pastel_po}{rgb255(0cm)=(225,160,100); rgb255(50cm)=(255,255,255); rgb255(100cm)=(160,135,200)},
    point meta min=-25.0,
    point meta max=25.0,
    xtick={0,1,2,3,4,5,6,7,8,9,10},
    xticklabels={$p_{\min}$,,,,,,,,,,$p_{\max}$},
    ytick={0,1,2,3,4,5,6,7,8,9,10},
    x tick label style={font=\scriptsize},
    y tick label style={font=\scriptsize},
    view={0}{90},
    axis line style={draw=none},
    major tick length=0pt,
    enlargelimits=false,
    colorbar style={font=\tiny},
    title style={font=\small},
    title={Rotation/Permutation},
    title style={yshift=-1.0ex},
]
\addplot[matrix plot*, mesh/cols=11, mesh/rows=11, point meta=explicit] table[meta=C] {
x y C
0 10 9.10
1 10 7.40
2 10 8.10
3 10 6.60
4 10 6.00
5 10 1.80
6 10 2.80
7 10 5.40
8 10 5.40
9 10 5.40
10 10 4.80
0 9 7.30
1 9 9.40
2 9 7.90
3 9 6.70
4 9 11.40
5 9 8.10
6 9 4.30
7 9 5.00
8 9 6.40
9 9 2.80
10 9 5.70
0 8 6.90
1 8 5.70
2 8 6.20
3 8 3.60
4 8 8.70
5 8 6.20
6 8 5.70
7 8 3.80
8 8 5.90
9 8 5.90
10 8 7.40
0 7 7.60
1 7 11.20
2 7 8.90
3 7 9.30
4 7 5.50
5 7 8.20
6 7 5.50
7 7 6.20
8 7 5.30
9 7 5.50
10 7 3.80
0 6 5.10
1 6 4.90
2 6 4.90
3 6 8.80
4 6 7.40
5 6 4.60
6 6 1.90
7 6 7.20
8 6 4.70
9 6 5.90
10 6 1.70
0 5 9.40
1 5 5.40
2 5 5.70
3 5 2.00
4 5 4.00
5 5 3.90
6 5 0.40
7 5 2.00
8 5 1.80
9 5 4.00
10 5 3.60
0 4 6.50
1 4 3.50
2 4 3.60
3 4 2.10
4 4 2.00
5 4 0.80
6 4 1.90
7 4 1.70
8 4 -0.60
9 4 1.40
10 4 2.90
0 3 4.20
1 3 2.20
2 3 3.50
3 3 -0.90
4 3 -0.60
5 3 1.90
6 3 1.30
7 3 4.50
8 3 -0.30
9 3 -1.00
10 3 2.30
0 2 2.80
1 2 -0.90
2 2 1.90
3 2 0.60
4 2 -2.10
5 2 0.80
6 2 -0.20
7 2 2.20
8 2 -2.80
9 2 0.90
10 2 -1.60
0 1 -0.30
1 1 -0.70
2 1 -1.00
3 1 -2.10
4 1 1.40
5 1 -0.30
6 1 -1.50
7 1 -0.10
8 1 0.90
9 1 3.70
10 1 0.10
0 0 0.50
1 0 3.30
2 0 2.90
3 0 -0.70
4 0 -4.00
5 0 -2.20
6 0 0.50
7 0 -5.10
8 0 -4.10
9 0 -0.40
10 0 2.70
};
\node[font=\fontsize{4}{4.8}\selectfont] at (axis cs:0,0) {0.5};
\node[font=\fontsize{4}{4.8}\selectfont] at (axis cs:1,0) {3.3};
\node[font=\fontsize{4}{4.8}\selectfont] at (axis cs:2,0) {2.9};
\node[font=\fontsize{4}{4.8}\selectfont] at (axis cs:3,0) {-0.7};
\node[font=\fontsize{4}{4.8}\selectfont] at (axis cs:4,0) {-4.0};
\node[font=\fontsize{4}{4.8}\selectfont] at (axis cs:5,0) {-2.2};
\node[font=\fontsize{4}{4.8}\selectfont] at (axis cs:6,0) {0.5};
\node[font=\fontsize{4}{4.8}\selectfont] at (axis cs:7,0) {-5.1};
\node[font=\fontsize{4}{4.8}\selectfont] at (axis cs:8,0) {-4.1};
\node[font=\fontsize{4}{4.8}\selectfont] at (axis cs:9,0) {-0.4};
\node[font=\fontsize{4}{4.8}\selectfont] at (axis cs:10,0) {2.7};
\node[font=\fontsize{4}{4.8}\selectfont] at (axis cs:0,1) {-0.3};
\node[font=\fontsize{4}{4.8}\selectfont] at (axis cs:1,1) {-0.7};
\node[font=\fontsize{4}{4.8}\selectfont] at (axis cs:2,1) {-1.0};
\node[font=\fontsize{4}{4.8}\selectfont] at (axis cs:3,1) {-2.1};
\node[font=\fontsize{4}{4.8}\selectfont] at (axis cs:4,1) {1.4};
\node[font=\fontsize{4}{4.8}\selectfont] at (axis cs:5,1) {-0.3};
\node[font=\fontsize{4}{4.8}\selectfont] at (axis cs:6,1) {-1.5};
\node[font=\fontsize{4}{4.8}\selectfont] at (axis cs:7,1) {-0.1};
\node[font=\fontsize{4}{4.8}\selectfont] at (axis cs:8,1) {0.9};
\node[font=\fontsize{4}{4.8}\selectfont] at (axis cs:9,1) {3.7};
\node[font=\fontsize{4}{4.8}\selectfont] at (axis cs:10,1) {0.1};
\node[font=\fontsize{4}{4.8}\selectfont] at (axis cs:0,2) {2.8};
\node[font=\fontsize{4}{4.8}\selectfont] at (axis cs:1,2) {-0.9};
\node[font=\fontsize{4}{4.8}\selectfont] at (axis cs:2,2) {1.9};
\node[font=\fontsize{4}{4.8}\selectfont] at (axis cs:3,2) {0.6};
\node[font=\fontsize{4}{4.8}\selectfont] at (axis cs:4,2) {-2.1};
\node[font=\fontsize{4}{4.8}\selectfont] at (axis cs:5,2) {0.8};
\node[font=\fontsize{4}{4.8}\selectfont] at (axis cs:6,2) {-0.2};
\node[font=\fontsize{4}{4.8}\selectfont] at (axis cs:7,2) {2.2};
\node[font=\fontsize{4}{4.8}\selectfont] at (axis cs:8,2) {-2.8};
\node[font=\fontsize{4}{4.8}\selectfont] at (axis cs:9,2) {0.9};
\node[font=\fontsize{4}{4.8}\selectfont] at (axis cs:10,2) {-1.6};
\node[font=\fontsize{4}{4.8}\selectfont] at (axis cs:0,3) {4.2};
\node[font=\fontsize{4}{4.8}\selectfont] at (axis cs:1,3) {2.2};
\node[font=\fontsize{4}{4.8}\selectfont] at (axis cs:2,3) {3.5};
\node[font=\fontsize{4}{4.8}\selectfont] at (axis cs:3,3) {-0.9};
\node[font=\fontsize{4}{4.8}\selectfont] at (axis cs:4,3) {-0.6};
\node[font=\fontsize{4}{4.8}\selectfont] at (axis cs:5,3) {1.9};
\node[font=\fontsize{4}{4.8}\selectfont] at (axis cs:6,3) {1.3};
\node[font=\fontsize{4}{4.8}\selectfont] at (axis cs:7,3) {4.5};
\node[font=\fontsize{4}{4.8}\selectfont] at (axis cs:8,3) {-0.3};
\node[font=\fontsize{4}{4.8}\selectfont] at (axis cs:9,3) {-1.0};
\node[font=\fontsize{4}{4.8}\selectfont] at (axis cs:10,3) {2.3};
\node[font=\fontsize{4}{4.8}\selectfont] at (axis cs:0,4) {6.5};
\node[font=\fontsize{4}{4.8}\selectfont] at (axis cs:1,4) {3.5};
\node[font=\fontsize{4}{4.8}\selectfont] at (axis cs:2,4) {3.6};
\node[font=\fontsize{4}{4.8}\selectfont] at (axis cs:3,4) {2.1};
\node[font=\fontsize{4}{4.8}\selectfont] at (axis cs:4,4) {2.0};
\node[font=\fontsize{4}{4.8}\selectfont] at (axis cs:5,4) {0.8};
\node[font=\fontsize{4}{4.8}\selectfont] at (axis cs:6,4) {1.9};
\node[font=\fontsize{4}{4.8}\selectfont] at (axis cs:7,4) {1.7};
\node[font=\fontsize{4}{4.8}\selectfont] at (axis cs:8,4) {-0.6};
\node[font=\fontsize{4}{4.8}\selectfont] at (axis cs:9,4) {1.4};
\node[font=\fontsize{4}{4.8}\selectfont] at (axis cs:10,4) {2.9};
\node[font=\fontsize{4}{4.8}\selectfont] at (axis cs:0,5) {9.4};
\node[font=\fontsize{4}{4.8}\selectfont] at (axis cs:1,5) {5.4};
\node[font=\fontsize{4}{4.8}\selectfont] at (axis cs:2,5) {5.7};
\node[font=\fontsize{4}{4.8}\selectfont] at (axis cs:3,5) {2.0};
\node[font=\fontsize{4}{4.8}\selectfont] at (axis cs:4,5) {4.0};
\node[font=\fontsize{4}{4.8}\selectfont] at (axis cs:5,5) {3.9};
\node[font=\fontsize{4}{4.8}\selectfont] at (axis cs:6,5) {0.4};
\node[font=\fontsize{4}{4.8}\selectfont] at (axis cs:7,5) {2.0};
\node[font=\fontsize{4}{4.8}\selectfont] at (axis cs:8,5) {1.8};
\node[font=\fontsize{4}{4.8}\selectfont] at (axis cs:9,5) {4.0};
\node[font=\fontsize{4}{4.8}\selectfont] at (axis cs:10,5) {3.6};
\node[font=\fontsize{4}{4.8}\selectfont] at (axis cs:0,6) {5.1};
\node[font=\fontsize{4}{4.8}\selectfont] at (axis cs:1,6) {4.9};
\node[font=\fontsize{4}{4.8}\selectfont] at (axis cs:2,6) {4.9};
\node[font=\fontsize{4}{4.8}\selectfont] at (axis cs:3,6) {8.8};
\node[font=\fontsize{4}{4.8}\selectfont] at (axis cs:4,6) {7.4};
\node[font=\fontsize{4}{4.8}\selectfont] at (axis cs:5,6) {4.6};
\node[font=\fontsize{4}{4.8}\selectfont] at (axis cs:6,6) {1.9};
\node[font=\fontsize{4}{4.8}\selectfont] at (axis cs:7,6) {7.2};
\node[font=\fontsize{4}{4.8}\selectfont] at (axis cs:8,6) {4.7};
\node[font=\fontsize{4}{4.8}\selectfont] at (axis cs:9,6) {5.9};
\node[font=\fontsize{4}{4.8}\selectfont] at (axis cs:10,6) {1.7};
\node[font=\fontsize{4}{4.8}\selectfont] at (axis cs:0,7) {7.6};
\node[font=\fontsize{4}{4.8}\selectfont] at (axis cs:1,7) {11.2};
\node[font=\fontsize{4}{4.8}\selectfont] at (axis cs:2,7) {8.9};
\node[font=\fontsize{4}{4.8}\selectfont] at (axis cs:3,7) {9.3};
\node[font=\fontsize{4}{4.8}\selectfont] at (axis cs:4,7) {5.5};
\node[font=\fontsize{4}{4.8}\selectfont] at (axis cs:5,7) {8.2};
\node[font=\fontsize{4}{4.8}\selectfont] at (axis cs:6,7) {5.5};
\node[font=\fontsize{4}{4.8}\selectfont] at (axis cs:7,7) {6.2};
\node[font=\fontsize{4}{4.8}\selectfont] at (axis cs:8,7) {5.3};
\node[font=\fontsize{4}{4.8}\selectfont] at (axis cs:9,7) {5.5};
\node[font=\fontsize{4}{4.8}\selectfont] at (axis cs:10,7) {3.8};
\node[font=\fontsize{4}{4.8}\selectfont] at (axis cs:0,8) {6.9};
\node[font=\fontsize{4}{4.8}\selectfont] at (axis cs:1,8) {5.7};
\node[font=\fontsize{4}{4.8}\selectfont] at (axis cs:2,8) {6.2};
\node[font=\fontsize{4}{4.8}\selectfont] at (axis cs:3,8) {3.6};
\node[font=\fontsize{4}{4.8}\selectfont] at (axis cs:4,8) {8.7};
\node[font=\fontsize{4}{4.8}\selectfont] at (axis cs:5,8) {6.2};
\node[font=\fontsize{4}{4.8}\selectfont] at (axis cs:6,8) {5.7};
\node[font=\fontsize{4}{4.8}\selectfont] at (axis cs:7,8) {3.8};
\node[font=\fontsize{4}{4.8}\selectfont] at (axis cs:8,8) {5.9};
\node[font=\fontsize{4}{4.8}\selectfont] at (axis cs:9,8) {5.9};
\node[font=\fontsize{4}{4.8}\selectfont] at (axis cs:10,8) {7.4};
\node[font=\fontsize{4}{4.8}\selectfont] at (axis cs:0,9) {7.3};
\node[font=\fontsize{4}{4.8}\selectfont] at (axis cs:1,9) {9.4};
\node[font=\fontsize{4}{4.8}\selectfont] at (axis cs:2,9) {7.9};
\node[font=\fontsize{4}{4.8}\selectfont] at (axis cs:3,9) {6.7};
\node[font=\fontsize{4}{4.8}\selectfont] at (axis cs:4,9) {11.4};
\node[font=\fontsize{4}{4.8}\selectfont] at (axis cs:5,9) {8.1};
\node[font=\fontsize{4}{4.8}\selectfont] at (axis cs:6,9) {4.3};
\node[font=\fontsize{4}{4.8}\selectfont] at (axis cs:7,9) {5.0};
\node[font=\fontsize{4}{4.8}\selectfont] at (axis cs:8,9) {6.4};
\node[font=\fontsize{4}{4.8}\selectfont] at (axis cs:9,9) {2.8};
\node[font=\fontsize{4}{4.8}\selectfont] at (axis cs:10,9) {5.7};
\node[font=\fontsize{4}{4.8}\selectfont] at (axis cs:0,10) {9.1};
\node[font=\fontsize{4}{4.8}\selectfont] at (axis cs:1,10) {7.4};
\node[font=\fontsize{4}{4.8}\selectfont] at (axis cs:2,10) {8.1};
\node[font=\fontsize{4}{4.8}\selectfont] at (axis cs:3,10) {6.6};
\node[font=\fontsize{4}{4.8}\selectfont] at (axis cs:4,10) {6.0};
\node[font=\fontsize{4}{4.8}\selectfont] at (axis cs:5,10) {1.8};
\node[font=\fontsize{4}{4.8}\selectfont] at (axis cs:6,10) {2.8};
\node[font=\fontsize{4}{4.8}\selectfont] at (axis cs:7,10) {5.4};
\node[font=\fontsize{4}{4.8}\selectfont] at (axis cs:8,10) {5.4};
\node[font=\fontsize{4}{4.8}\selectfont] at (axis cs:9,10) {5.4};
\node[font=\fontsize{4}{4.8}\selectfont] at (axis cs:10,10) {4.8};
\end{axis}
\end{tikzpicture}
            \end{minipage}
        \end{minipage}}
            \vspace{-5pt}
        \caption{\small
            Same as Figure~\ref{fig:icl_deltas_qwen32}, for \model{llama8}.
        }
        \vspace{-8pt}
        \label{fig:few_shot_delta_aliases_llama8}
    \end{figure}
    \FloatBarrier

    \begin{figure}[ht]
        \centering
        \scalebox{0.77}{\begin{minipage}{\textwidth}\centering
            \begin{minipage}[t]{0.48\textwidth}
                \input{plots/11_appendix/few_shot/delta/mistral24_dropout_noise}
            \end{minipage}%
            \hfill
            \begin{minipage}[t]{0.48\textwidth}
                \input{plots/11_appendix/few_shot/delta/mistral24_masking_jitter}
            \end{minipage}\\[0pt]
            \begin{minipage}[t]{0.48\textwidth}
                \input{plots/11_appendix/few_shot/delta/mistral24_x_y}
            \end{minipage}%
            \hfill
            \begin{minipage}[t]{0.48\textwidth}
                \input{plots/11_appendix/few_shot/delta/mistral24_rotation_permutation}
            \end{minipage}
        \end{minipage}}
            \vspace{-5pt}
        \caption{\small
            Same as Figure~\ref{fig:icl_deltas_qwen32}, for \model{mistral24}.
        }
        \vspace{-8pt}
        \label{fig:few_shot_delta_aliases_mistral24}
    \end{figure}
    \FloatBarrier

    \begin{figure}[ht]
        \centering
        \scalebox{0.77}{\begin{minipage}{\textwidth}\centering
            \begin{minipage}[t]{0.48\textwidth}
                \centering
\begin{tikzpicture}
\begin{axis}[
    width=0.82\linewidth,
    height=0.82\linewidth,
    xlabel={},
    ylabel={\small Noise SD $\sigma$},
    colormap={pastel_po}{rgb255(0cm)=(225,160,100); rgb255(50cm)=(255,255,255); rgb255(100cm)=(160,135,200)},
    point meta min=-25.0,
    point meta max=25.0,
    xtick={0,1,2,3,4,5,6,7,8,9,10},
    xticklabels={$p_{\min}$,,,,,,,,,,$p_{\max}$},
    ytick={0,1,2,3,4,5,6,7,8,9,10},
    yticklabels={$\sigma_{\min}$,,,,,,,,,,$\sigma_{\max}$},
    x tick label style={font=\scriptsize},
    y tick label style={font=\scriptsize},
    view={0}{90},
    axis line style={draw=none},
    major tick length=0pt,
    enlargelimits=false,
    title style={font=\small},
    title={Dropout/Noise},
    title style={yshift=-1.0ex},
]
\addplot[matrix plot*, mesh/cols=11, mesh/rows=11, point meta=explicit] table[meta=C] {
x y C
0 10 -8.30
1 10 -2.90
2 10 -4.70
3 10 -8.40
4 10 -3.20
5 10 -1.20
6 10 1.50
7 10 -1.80
8 10 1.20
9 10 1.30
10 10 -0.90
0 9 -3.90
1 9 0.20
2 9 -5.70
3 9 -0.80
4 9 -2.60
5 9 -2.60
6 9 -0.70
7 9 -5.40
8 9 4.70
9 9 4.50
10 9 6.70
0 8 -0.70
1 8 -1.70
2 8 -1.50
3 8 -1.40
4 8 0.00
5 8 2.40
6 8 1.00
7 8 5.10
8 8 3.00
9 8 5.20
10 8 0.40
0 7 -1.40
1 7 4.70
2 7 1.90
3 7 0.90
4 7 2.60
5 7 3.20
6 7 6.50
7 7 6.20
8 7 6.10
9 7 9.40
10 7 9.20
0 6 -1.90
1 6 -0.40
2 6 3.00
3 6 2.30
4 6 3.30
5 6 4.90
6 6 5.90
7 6 10.50
8 6 9.50
9 6 9.20
10 6 4.40
0 5 4.50
1 5 3.70
2 5 4.70
3 5 5.00
4 5 8.60
5 5 3.00
6 5 5.40
7 5 7.50
8 5 9.10
9 5 8.40
10 5 6.80
0 4 0.40
1 4 3.30
2 4 2.20
3 4 2.60
4 4 1.70
5 4 3.90
6 4 7.50
7 4 11.10
8 4 7.80
9 4 9.10
10 4 9.60
0 3 3.00
1 3 1.80
2 3 2.80
3 3 2.40
4 3 3.10
5 3 5.00
6 3 8.70
7 3 6.80
8 3 5.70
9 3 11.20
10 3 4.70
0 2 3.30
1 2 -1.90
2 2 0.50
3 2 -0.20
4 2 6.60
5 2 4.90
6 2 3.80
7 2 6.70
8 2 8.30
9 2 3.70
10 2 4.10
0 1 1.90
1 1 1.20
2 1 -4.60
3 1 2.40
4 1 0.90
5 1 5.10
6 1 3.80
7 1 8.10
8 1 5.80
9 1 5.00
10 1 5.60
0 0 0.90
1 0 0.80
2 0 0.80
3 0 -2.90
4 0 3.00
5 0 2.50
6 0 7.80
7 0 2.50
8 0 5.30
9 0 9.10
10 0 3.80
};
\node[font=\fontsize{4}{4.8}\selectfont] at (axis cs:0,0) {0.9};
\node[font=\fontsize{4}{4.8}\selectfont] at (axis cs:1,0) {0.8};
\node[font=\fontsize{4}{4.8}\selectfont] at (axis cs:2,0) {0.8};
\node[font=\fontsize{4}{4.8}\selectfont] at (axis cs:3,0) {-2.9};
\node[font=\fontsize{4}{4.8}\selectfont] at (axis cs:4,0) {3.0};
\node[font=\fontsize{4}{4.8}\selectfont] at (axis cs:5,0) {2.5};
\node[font=\fontsize{4}{4.8}\selectfont] at (axis cs:6,0) {7.8};
\node[font=\fontsize{4}{4.8}\selectfont] at (axis cs:7,0) {2.5};
\node[font=\fontsize{4}{4.8}\selectfont] at (axis cs:8,0) {5.3};
\node[font=\fontsize{4}{4.8}\selectfont] at (axis cs:9,0) {9.1};
\node[font=\fontsize{4}{4.8}\selectfont] at (axis cs:10,0) {3.8};
\node[font=\fontsize{4}{4.8}\selectfont] at (axis cs:0,1) {1.9};
\node[font=\fontsize{4}{4.8}\selectfont] at (axis cs:1,1) {1.2};
\node[font=\fontsize{4}{4.8}\selectfont] at (axis cs:2,1) {-4.6};
\node[font=\fontsize{4}{4.8}\selectfont] at (axis cs:3,1) {2.4};
\node[font=\fontsize{4}{4.8}\selectfont] at (axis cs:4,1) {0.9};
\node[font=\fontsize{4}{4.8}\selectfont] at (axis cs:5,1) {5.1};
\node[font=\fontsize{4}{4.8}\selectfont] at (axis cs:6,1) {3.8};
\node[font=\fontsize{4}{4.8}\selectfont] at (axis cs:7,1) {8.1};
\node[font=\fontsize{4}{4.8}\selectfont] at (axis cs:8,1) {5.8};
\node[font=\fontsize{4}{4.8}\selectfont] at (axis cs:9,1) {5.0};
\node[font=\fontsize{4}{4.8}\selectfont] at (axis cs:10,1) {5.6};
\node[font=\fontsize{4}{4.8}\selectfont] at (axis cs:0,2) {3.3};
\node[font=\fontsize{4}{4.8}\selectfont] at (axis cs:1,2) {-1.9};
\node[font=\fontsize{4}{4.8}\selectfont] at (axis cs:2,2) {0.5};
\node[font=\fontsize{4}{4.8}\selectfont] at (axis cs:3,2) {-0.2};
\node[font=\fontsize{4}{4.8}\selectfont] at (axis cs:4,2) {6.6};
\node[font=\fontsize{4}{4.8}\selectfont] at (axis cs:5,2) {4.9};
\node[font=\fontsize{4}{4.8}\selectfont] at (axis cs:6,2) {3.8};
\node[font=\fontsize{4}{4.8}\selectfont] at (axis cs:7,2) {6.7};
\node[font=\fontsize{4}{4.8}\selectfont] at (axis cs:8,2) {8.3};
\node[font=\fontsize{4}{4.8}\selectfont] at (axis cs:9,2) {3.7};
\node[font=\fontsize{4}{4.8}\selectfont] at (axis cs:10,2) {4.1};
\node[font=\fontsize{4}{4.8}\selectfont] at (axis cs:0,3) {3.0};
\node[font=\fontsize{4}{4.8}\selectfont] at (axis cs:1,3) {1.8};
\node[font=\fontsize{4}{4.8}\selectfont] at (axis cs:2,3) {2.8};
\node[font=\fontsize{4}{4.8}\selectfont] at (axis cs:3,3) {2.4};
\node[font=\fontsize{4}{4.8}\selectfont] at (axis cs:4,3) {3.1};
\node[font=\fontsize{4}{4.8}\selectfont] at (axis cs:5,3) {5.0};
\node[font=\fontsize{4}{4.8}\selectfont] at (axis cs:6,3) {8.7};
\node[font=\fontsize{4}{4.8}\selectfont] at (axis cs:7,3) {6.8};
\node[font=\fontsize{4}{4.8}\selectfont] at (axis cs:8,3) {5.7};
\node[font=\fontsize{4}{4.8}\selectfont] at (axis cs:9,3) {11.2};
\node[font=\fontsize{4}{4.8}\selectfont] at (axis cs:10,3) {4.7};
\node[font=\fontsize{4}{4.8}\selectfont] at (axis cs:0,4) {0.4};
\node[font=\fontsize{4}{4.8}\selectfont] at (axis cs:1,4) {3.3};
\node[font=\fontsize{4}{4.8}\selectfont] at (axis cs:2,4) {2.2};
\node[font=\fontsize{4}{4.8}\selectfont] at (axis cs:3,4) {2.6};
\node[font=\fontsize{4}{4.8}\selectfont] at (axis cs:4,4) {1.7};
\node[font=\fontsize{4}{4.8}\selectfont] at (axis cs:5,4) {3.9};
\node[font=\fontsize{4}{4.8}\selectfont] at (axis cs:6,4) {7.5};
\node[font=\fontsize{4}{4.8}\selectfont] at (axis cs:7,4) {11.1};
\node[font=\fontsize{4}{4.8}\selectfont] at (axis cs:8,4) {7.8};
\node[font=\fontsize{4}{4.8}\selectfont] at (axis cs:9,4) {9.1};
\node[font=\fontsize{4}{4.8}\selectfont] at (axis cs:10,4) {9.6};
\node[font=\fontsize{4}{4.8}\selectfont] at (axis cs:0,5) {4.5};
\node[font=\fontsize{4}{4.8}\selectfont] at (axis cs:1,5) {3.7};
\node[font=\fontsize{4}{4.8}\selectfont] at (axis cs:2,5) {4.7};
\node[font=\fontsize{4}{4.8}\selectfont] at (axis cs:3,5) {5.0};
\node[font=\fontsize{4}{4.8}\selectfont] at (axis cs:4,5) {8.6};
\node[font=\fontsize{4}{4.8}\selectfont] at (axis cs:5,5) {3.0};
\node[font=\fontsize{4}{4.8}\selectfont] at (axis cs:6,5) {5.4};
\node[font=\fontsize{4}{4.8}\selectfont] at (axis cs:7,5) {7.5};
\node[font=\fontsize{4}{4.8}\selectfont] at (axis cs:8,5) {9.1};
\node[font=\fontsize{4}{4.8}\selectfont] at (axis cs:9,5) {8.4};
\node[font=\fontsize{4}{4.8}\selectfont] at (axis cs:10,5) {6.8};
\node[font=\fontsize{4}{4.8}\selectfont] at (axis cs:0,6) {-1.9};
\node[font=\fontsize{4}{4.8}\selectfont] at (axis cs:1,6) {-0.4};
\node[font=\fontsize{4}{4.8}\selectfont] at (axis cs:2,6) {3.0};
\node[font=\fontsize{4}{4.8}\selectfont] at (axis cs:3,6) {2.3};
\node[font=\fontsize{4}{4.8}\selectfont] at (axis cs:4,6) {3.3};
\node[font=\fontsize{4}{4.8}\selectfont] at (axis cs:5,6) {4.9};
\node[font=\fontsize{4}{4.8}\selectfont] at (axis cs:6,6) {5.9};
\node[font=\fontsize{4}{4.8}\selectfont] at (axis cs:7,6) {10.5};
\node[font=\fontsize{4}{4.8}\selectfont] at (axis cs:8,6) {9.5};
\node[font=\fontsize{4}{4.8}\selectfont] at (axis cs:9,6) {9.2};
\node[font=\fontsize{4}{4.8}\selectfont] at (axis cs:10,6) {4.4};
\node[font=\fontsize{4}{4.8}\selectfont] at (axis cs:0,7) {-1.4};
\node[font=\fontsize{4}{4.8}\selectfont] at (axis cs:1,7) {4.7};
\node[font=\fontsize{4}{4.8}\selectfont] at (axis cs:2,7) {1.9};
\node[font=\fontsize{4}{4.8}\selectfont] at (axis cs:3,7) {0.9};
\node[font=\fontsize{4}{4.8}\selectfont] at (axis cs:4,7) {2.6};
\node[font=\fontsize{4}{4.8}\selectfont] at (axis cs:5,7) {3.2};
\node[font=\fontsize{4}{4.8}\selectfont] at (axis cs:6,7) {6.5};
\node[font=\fontsize{4}{4.8}\selectfont] at (axis cs:7,7) {6.2};
\node[font=\fontsize{4}{4.8}\selectfont] at (axis cs:8,7) {6.1};
\node[font=\fontsize{4}{4.8}\selectfont] at (axis cs:9,7) {9.4};
\node[font=\fontsize{4}{4.8}\selectfont] at (axis cs:10,7) {9.2};
\node[font=\fontsize{4}{4.8}\selectfont] at (axis cs:0,8) {-0.7};
\node[font=\fontsize{4}{4.8}\selectfont] at (axis cs:1,8) {-1.7};
\node[font=\fontsize{4}{4.8}\selectfont] at (axis cs:2,8) {-1.5};
\node[font=\fontsize{4}{4.8}\selectfont] at (axis cs:3,8) {-1.4};
\node[font=\fontsize{4}{4.8}\selectfont] at (axis cs:4,8) {0.0};
\node[font=\fontsize{4}{4.8}\selectfont] at (axis cs:5,8) {2.4};
\node[font=\fontsize{4}{4.8}\selectfont] at (axis cs:6,8) {1.0};
\node[font=\fontsize{4}{4.8}\selectfont] at (axis cs:7,8) {5.1};
\node[font=\fontsize{4}{4.8}\selectfont] at (axis cs:8,8) {3.0};
\node[font=\fontsize{4}{4.8}\selectfont] at (axis cs:9,8) {5.2};
\node[font=\fontsize{4}{4.8}\selectfont] at (axis cs:10,8) {0.4};
\node[font=\fontsize{4}{4.8}\selectfont] at (axis cs:0,9) {-3.9};
\node[font=\fontsize{4}{4.8}\selectfont] at (axis cs:1,9) {0.2};
\node[font=\fontsize{4}{4.8}\selectfont] at (axis cs:2,9) {-5.7};
\node[font=\fontsize{4}{4.8}\selectfont] at (axis cs:3,9) {-0.8};
\node[font=\fontsize{4}{4.8}\selectfont] at (axis cs:4,9) {-2.6};
\node[font=\fontsize{4}{4.8}\selectfont] at (axis cs:5,9) {-2.6};
\node[font=\fontsize{4}{4.8}\selectfont] at (axis cs:6,9) {-0.7};
\node[font=\fontsize{4}{4.8}\selectfont] at (axis cs:7,9) {-5.4};
\node[font=\fontsize{4}{4.8}\selectfont] at (axis cs:8,9) {4.7};
\node[font=\fontsize{4}{4.8}\selectfont] at (axis cs:9,9) {4.5};
\node[font=\fontsize{4}{4.8}\selectfont] at (axis cs:10,9) {6.7};
\node[font=\fontsize{4}{4.8}\selectfont] at (axis cs:0,10) {-8.3};
\node[font=\fontsize{4}{4.8}\selectfont] at (axis cs:1,10) {-2.9};
\node[font=\fontsize{4}{4.8}\selectfont] at (axis cs:2,10) {-4.7};
\node[font=\fontsize{4}{4.8}\selectfont] at (axis cs:3,10) {-8.4};
\node[font=\fontsize{4}{4.8}\selectfont] at (axis cs:4,10) {-3.2};
\node[font=\fontsize{4}{4.8}\selectfont] at (axis cs:5,10) {-1.2};
\node[font=\fontsize{4}{4.8}\selectfont] at (axis cs:6,10) {1.5};
\node[font=\fontsize{4}{4.8}\selectfont] at (axis cs:7,10) {-1.8};
\node[font=\fontsize{4}{4.8}\selectfont] at (axis cs:8,10) {1.2};
\node[font=\fontsize{4}{4.8}\selectfont] at (axis cs:9,10) {1.3};
\node[font=\fontsize{4}{4.8}\selectfont] at (axis cs:10,10) {-0.9};
\end{axis}
\end{tikzpicture}
            \end{minipage}%
            \hfill
            \begin{minipage}[t]{0.48\textwidth}
                \centering
\begin{tikzpicture}
\begin{axis}[
    width=0.82\linewidth,
    height=0.82\linewidth,
    xlabel={},
    ylabel={},
    yticklabels={},
    colorbar,
    colormap={pastel_po}{rgb255(0cm)=(225,160,100); rgb255(50cm)=(255,255,255); rgb255(100cm)=(160,135,200)},
    point meta min=-25.0,
    point meta max=25.0,
    xtick={0,1,2,3,4,5,6,7,8,9,10},
    xticklabels={$p_{\min}$,,,,,,,,,,$p_{\max}$},
    ytick={0,1,2,3,4,5,6,7,8,9,10},
    x tick label style={font=\scriptsize},
    y tick label style={font=\scriptsize},
    view={0}{90},
    axis line style={draw=none},
    major tick length=0pt,
    enlargelimits=false,
    colorbar style={font=\tiny},
    title style={font=\small},
    title={Masking/Jitter},
    title style={yshift=-1.0ex},
]
\addplot[matrix plot*, mesh/cols=11, mesh/rows=11, point meta=explicit] table[meta=C] {
x y C
0 10 -2.80
1 10 -2.80
2 10 -3.30
3 10 0.10
4 10 -0.10
5 10 -1.10
6 10 2.90
7 10 -1.50
8 10 -1.50
9 10 1.70
10 10 -1.20
0 9 -3.30
1 9 -2.60
2 9 0.60
3 9 -0.90
4 9 -0.80
5 9 4.20
6 9 2.40
7 9 6.20
8 9 3.30
9 9 2.20
10 9 -0.20
0 8 -2.40
1 8 -1.90
2 8 -1.70
3 8 3.40
4 8 0.20
5 8 1.50
6 8 0.70
7 8 0.80
8 8 0.10
9 8 3.80
10 8 1.50
0 7 -6.20
1 7 -3.70
2 7 -2.30
3 7 -1.40
4 7 1.60
5 7 1.60
6 7 3.60
7 7 -1.20
8 7 4.30
9 7 0.00
10 7 2.60
0 6 -4.50
1 6 1.00
2 6 0.00
3 6 -1.00
4 6 -1.10
5 6 2.90
6 6 -1.50
7 6 -2.10
8 6 2.40
9 6 0.60
10 6 2.30
0 5 -5.10
1 5 -5.30
2 5 -1.60
3 5 -1.30
4 5 -1.00
5 5 -1.60
6 5 0.10
7 5 2.30
8 5 2.40
9 5 0.60
10 5 1.60
0 4 -5.80
1 4 -4.30
2 4 -4.90
3 4 1.60
4 4 -1.70
5 4 1.20
6 4 -2.00
7 4 -0.30
8 4 0.10
9 4 -2.40
10 4 0.90
0 3 -2.70
1 3 -0.50
2 3 -1.10
3 3 0.50
4 3 0.10
5 3 3.00
6 3 4.90
7 3 2.40
8 3 3.00
9 3 1.80
10 3 0.30
0 2 2.90
1 2 0.70
2 2 1.10
3 2 -0.40
4 2 -0.70
5 2 4.90
6 2 2.80
7 2 2.30
8 2 4.40
9 2 -0.10
10 2 3.70
0 1 2.00
1 1 3.30
2 1 3.20
3 1 4.20
4 1 5.60
5 1 6.10
6 1 2.80
7 1 7.20
8 1 6.70
9 1 5.30
10 1 5.00
0 0 4.80
1 0 3.80
2 0 3.50
3 0 3.00
4 0 5.30
5 0 7.40
6 0 5.90
7 0 5.90
8 0 7.00
9 0 4.60
10 0 6.40
};
\node[font=\fontsize{4}{4.8}\selectfont] at (axis cs:0,0) {4.8};
\node[font=\fontsize{4}{4.8}\selectfont] at (axis cs:1,0) {3.8};
\node[font=\fontsize{4}{4.8}\selectfont] at (axis cs:2,0) {3.5};
\node[font=\fontsize{4}{4.8}\selectfont] at (axis cs:3,0) {3.0};
\node[font=\fontsize{4}{4.8}\selectfont] at (axis cs:4,0) {5.3};
\node[font=\fontsize{4}{4.8}\selectfont] at (axis cs:5,0) {7.4};
\node[font=\fontsize{4}{4.8}\selectfont] at (axis cs:6,0) {5.9};
\node[font=\fontsize{4}{4.8}\selectfont] at (axis cs:7,0) {5.9};
\node[font=\fontsize{4}{4.8}\selectfont] at (axis cs:8,0) {7.0};
\node[font=\fontsize{4}{4.8}\selectfont] at (axis cs:9,0) {4.6};
\node[font=\fontsize{4}{4.8}\selectfont] at (axis cs:10,0) {6.4};
\node[font=\fontsize{4}{4.8}\selectfont] at (axis cs:0,1) {2.0};
\node[font=\fontsize{4}{4.8}\selectfont] at (axis cs:1,1) {3.3};
\node[font=\fontsize{4}{4.8}\selectfont] at (axis cs:2,1) {3.2};
\node[font=\fontsize{4}{4.8}\selectfont] at (axis cs:3,1) {4.2};
\node[font=\fontsize{4}{4.8}\selectfont] at (axis cs:4,1) {5.6};
\node[font=\fontsize{4}{4.8}\selectfont] at (axis cs:5,1) {6.1};
\node[font=\fontsize{4}{4.8}\selectfont] at (axis cs:6,1) {2.8};
\node[font=\fontsize{4}{4.8}\selectfont] at (axis cs:7,1) {7.2};
\node[font=\fontsize{4}{4.8}\selectfont] at (axis cs:8,1) {6.7};
\node[font=\fontsize{4}{4.8}\selectfont] at (axis cs:9,1) {5.3};
\node[font=\fontsize{4}{4.8}\selectfont] at (axis cs:10,1) {5.0};
\node[font=\fontsize{4}{4.8}\selectfont] at (axis cs:0,2) {2.9};
\node[font=\fontsize{4}{4.8}\selectfont] at (axis cs:1,2) {0.7};
\node[font=\fontsize{4}{4.8}\selectfont] at (axis cs:2,2) {1.1};
\node[font=\fontsize{4}{4.8}\selectfont] at (axis cs:3,2) {-0.4};
\node[font=\fontsize{4}{4.8}\selectfont] at (axis cs:4,2) {-0.7};
\node[font=\fontsize{4}{4.8}\selectfont] at (axis cs:5,2) {4.9};
\node[font=\fontsize{4}{4.8}\selectfont] at (axis cs:6,2) {2.8};
\node[font=\fontsize{4}{4.8}\selectfont] at (axis cs:7,2) {2.3};
\node[font=\fontsize{4}{4.8}\selectfont] at (axis cs:8,2) {4.4};
\node[font=\fontsize{4}{4.8}\selectfont] at (axis cs:9,2) {-0.1};
\node[font=\fontsize{4}{4.8}\selectfont] at (axis cs:10,2) {3.7};
\node[font=\fontsize{4}{4.8}\selectfont] at (axis cs:0,3) {-2.7};
\node[font=\fontsize{4}{4.8}\selectfont] at (axis cs:1,3) {-0.5};
\node[font=\fontsize{4}{4.8}\selectfont] at (axis cs:2,3) {-1.1};
\node[font=\fontsize{4}{4.8}\selectfont] at (axis cs:3,3) {0.5};
\node[font=\fontsize{4}{4.8}\selectfont] at (axis cs:4,3) {0.1};
\node[font=\fontsize{4}{4.8}\selectfont] at (axis cs:5,3) {3.0};
\node[font=\fontsize{4}{4.8}\selectfont] at (axis cs:6,3) {4.9};
\node[font=\fontsize{4}{4.8}\selectfont] at (axis cs:7,3) {2.4};
\node[font=\fontsize{4}{4.8}\selectfont] at (axis cs:8,3) {3.0};
\node[font=\fontsize{4}{4.8}\selectfont] at (axis cs:9,3) {1.8};
\node[font=\fontsize{4}{4.8}\selectfont] at (axis cs:10,3) {0.3};
\node[font=\fontsize{4}{4.8}\selectfont] at (axis cs:0,4) {-5.8};
\node[font=\fontsize{4}{4.8}\selectfont] at (axis cs:1,4) {-4.3};
\node[font=\fontsize{4}{4.8}\selectfont] at (axis cs:2,4) {-4.9};
\node[font=\fontsize{4}{4.8}\selectfont] at (axis cs:3,4) {1.6};
\node[font=\fontsize{4}{4.8}\selectfont] at (axis cs:4,4) {-1.7};
\node[font=\fontsize{4}{4.8}\selectfont] at (axis cs:5,4) {1.2};
\node[font=\fontsize{4}{4.8}\selectfont] at (axis cs:6,4) {-2.0};
\node[font=\fontsize{4}{4.8}\selectfont] at (axis cs:7,4) {-0.3};
\node[font=\fontsize{4}{4.8}\selectfont] at (axis cs:8,4) {0.1};
\node[font=\fontsize{4}{4.8}\selectfont] at (axis cs:9,4) {-2.4};
\node[font=\fontsize{4}{4.8}\selectfont] at (axis cs:10,4) {0.9};
\node[font=\fontsize{4}{4.8}\selectfont] at (axis cs:0,5) {-5.1};
\node[font=\fontsize{4}{4.8}\selectfont] at (axis cs:1,5) {-5.3};
\node[font=\fontsize{4}{4.8}\selectfont] at (axis cs:2,5) {-1.6};
\node[font=\fontsize{4}{4.8}\selectfont] at (axis cs:3,5) {-1.3};
\node[font=\fontsize{4}{4.8}\selectfont] at (axis cs:4,5) {-1.0};
\node[font=\fontsize{4}{4.8}\selectfont] at (axis cs:5,5) {-1.6};
\node[font=\fontsize{4}{4.8}\selectfont] at (axis cs:6,5) {0.1};
\node[font=\fontsize{4}{4.8}\selectfont] at (axis cs:7,5) {2.3};
\node[font=\fontsize{4}{4.8}\selectfont] at (axis cs:8,5) {2.4};
\node[font=\fontsize{4}{4.8}\selectfont] at (axis cs:9,5) {0.6};
\node[font=\fontsize{4}{4.8}\selectfont] at (axis cs:10,5) {1.6};
\node[font=\fontsize{4}{4.8}\selectfont] at (axis cs:0,6) {-4.5};
\node[font=\fontsize{4}{4.8}\selectfont] at (axis cs:1,6) {1.0};
\node[font=\fontsize{4}{4.8}\selectfont] at (axis cs:2,6) {0.0};
\node[font=\fontsize{4}{4.8}\selectfont] at (axis cs:3,6) {-1.0};
\node[font=\fontsize{4}{4.8}\selectfont] at (axis cs:4,6) {-1.1};
\node[font=\fontsize{4}{4.8}\selectfont] at (axis cs:5,6) {2.9};
\node[font=\fontsize{4}{4.8}\selectfont] at (axis cs:6,6) {-1.5};
\node[font=\fontsize{4}{4.8}\selectfont] at (axis cs:7,6) {-2.1};
\node[font=\fontsize{4}{4.8}\selectfont] at (axis cs:8,6) {2.4};
\node[font=\fontsize{4}{4.8}\selectfont] at (axis cs:9,6) {0.6};
\node[font=\fontsize{4}{4.8}\selectfont] at (axis cs:10,6) {2.3};
\node[font=\fontsize{4}{4.8}\selectfont] at (axis cs:0,7) {-6.2};
\node[font=\fontsize{4}{4.8}\selectfont] at (axis cs:1,7) {-3.7};
\node[font=\fontsize{4}{4.8}\selectfont] at (axis cs:2,7) {-2.3};
\node[font=\fontsize{4}{4.8}\selectfont] at (axis cs:3,7) {-1.4};
\node[font=\fontsize{4}{4.8}\selectfont] at (axis cs:4,7) {1.6};
\node[font=\fontsize{4}{4.8}\selectfont] at (axis cs:5,7) {1.6};
\node[font=\fontsize{4}{4.8}\selectfont] at (axis cs:6,7) {3.6};
\node[font=\fontsize{4}{4.8}\selectfont] at (axis cs:7,7) {-1.2};
\node[font=\fontsize{4}{4.8}\selectfont] at (axis cs:8,7) {4.3};
\node[font=\fontsize{4}{4.8}\selectfont] at (axis cs:9,7) {0.0};
\node[font=\fontsize{4}{4.8}\selectfont] at (axis cs:10,7) {2.6};
\node[font=\fontsize{4}{4.8}\selectfont] at (axis cs:0,8) {-2.4};
\node[font=\fontsize{4}{4.8}\selectfont] at (axis cs:1,8) {-1.9};
\node[font=\fontsize{4}{4.8}\selectfont] at (axis cs:2,8) {-1.7};
\node[font=\fontsize{4}{4.8}\selectfont] at (axis cs:3,8) {3.4};
\node[font=\fontsize{4}{4.8}\selectfont] at (axis cs:4,8) {0.2};
\node[font=\fontsize{4}{4.8}\selectfont] at (axis cs:5,8) {1.5};
\node[font=\fontsize{4}{4.8}\selectfont] at (axis cs:6,8) {0.7};
\node[font=\fontsize{4}{4.8}\selectfont] at (axis cs:7,8) {0.8};
\node[font=\fontsize{4}{4.8}\selectfont] at (axis cs:8,8) {0.1};
\node[font=\fontsize{4}{4.8}\selectfont] at (axis cs:9,8) {3.8};
\node[font=\fontsize{4}{4.8}\selectfont] at (axis cs:10,8) {1.5};
\node[font=\fontsize{4}{4.8}\selectfont] at (axis cs:0,9) {-3.3};
\node[font=\fontsize{4}{4.8}\selectfont] at (axis cs:1,9) {-2.6};
\node[font=\fontsize{4}{4.8}\selectfont] at (axis cs:2,9) {0.6};
\node[font=\fontsize{4}{4.8}\selectfont] at (axis cs:3,9) {-0.9};
\node[font=\fontsize{4}{4.8}\selectfont] at (axis cs:4,9) {-0.8};
\node[font=\fontsize{4}{4.8}\selectfont] at (axis cs:5,9) {4.2};
\node[font=\fontsize{4}{4.8}\selectfont] at (axis cs:6,9) {2.4};
\node[font=\fontsize{4}{4.8}\selectfont] at (axis cs:7,9) {6.2};
\node[font=\fontsize{4}{4.8}\selectfont] at (axis cs:8,9) {3.3};
\node[font=\fontsize{4}{4.8}\selectfont] at (axis cs:9,9) {2.2};
\node[font=\fontsize{4}{4.8}\selectfont] at (axis cs:10,9) {-0.2};
\node[font=\fontsize{4}{4.8}\selectfont] at (axis cs:0,10) {-2.8};
\node[font=\fontsize{4}{4.8}\selectfont] at (axis cs:1,10) {-2.8};
\node[font=\fontsize{4}{4.8}\selectfont] at (axis cs:2,10) {-3.3};
\node[font=\fontsize{4}{4.8}\selectfont] at (axis cs:3,10) {0.1};
\node[font=\fontsize{4}{4.8}\selectfont] at (axis cs:4,10) {-0.1};
\node[font=\fontsize{4}{4.8}\selectfont] at (axis cs:5,10) {-1.1};
\node[font=\fontsize{4}{4.8}\selectfont] at (axis cs:6,10) {2.9};
\node[font=\fontsize{4}{4.8}\selectfont] at (axis cs:7,10) {-1.5};
\node[font=\fontsize{4}{4.8}\selectfont] at (axis cs:8,10) {-1.5};
\node[font=\fontsize{4}{4.8}\selectfont] at (axis cs:9,10) {1.7};
\node[font=\fontsize{4}{4.8}\selectfont] at (axis cs:10,10) {-1.2};
\end{axis}
\end{tikzpicture}
            \end{minipage}\\[0pt]
            \begin{minipage}[t]{0.48\textwidth}
                \centering
\begin{tikzpicture}
\begin{axis}[
    width=0.82\linewidth,
    height=0.82\linewidth,
    xlabel={\small Dropout rate $p$},
    ylabel={\small Noise SD $\sigma$},
    colormap={pastel_po}{rgb255(0cm)=(225,160,100); rgb255(50cm)=(255,255,255); rgb255(100cm)=(160,135,200)},
    point meta min=-25.0,
    point meta max=25.0,
    xtick={0,1,2,3,4,5,6,7,8,9,10},
    xticklabels={$p_{\min}$,,,,,,,,,,$p_{\max}$},
    ytick={0,1,2,3,4,5,6,7,8,9,10},
    yticklabels={$\sigma_{\min}$,,,,,,,,,,$\sigma_{\max}$},
    x tick label style={font=\scriptsize},
    y tick label style={font=\scriptsize},
    view={0}{90},
    axis line style={draw=none},
    major tick length=0pt,
    enlargelimits=false,
    title style={font=\small},
    title={X/Y},
    title style={yshift=-1.0ex},
]
\addplot[matrix plot*, mesh/cols=11, mesh/rows=11, point meta=explicit] table[meta=C] {
x y C
0 10 6.10
1 10 3.80
2 10 5.00
3 10 1.70
4 10 1.40
5 10 0.80
6 10 2.00
7 10 -1.00
8 10 -0.10
9 10 1.50
10 10 4.30
0 9 2.70
1 9 5.20
2 9 5.10
3 9 -0.60
4 9 4.20
5 9 3.90
6 9 3.40
7 9 6.60
8 9 4.90
9 9 3.50
10 9 3.90
0 8 3.90
1 8 3.20
2 8 4.10
3 8 3.10
4 8 5.60
5 8 3.80
6 8 4.00
7 8 1.30
8 8 3.20
9 8 5.60
10 8 3.30
0 7 7.40
1 7 6.20
2 7 0.70
3 7 5.30
4 7 0.60
5 7 2.40
6 7 3.70
7 7 3.80
8 7 4.70
9 7 7.60
10 7 7.60
0 6 8.70
1 6 5.60
2 6 4.00
3 6 1.30
4 6 5.30
5 6 3.10
6 6 3.10
7 6 0.80
8 6 4.80
9 6 3.60
10 6 4.80
0 5 3.40
1 5 2.00
2 5 2.50
3 5 2.30
4 5 -0.70
5 5 4.00
6 5 -2.00
7 5 2.40
8 5 5.20
9 5 5.00
10 5 2.80
0 4 -0.70
1 4 2.00
2 4 3.20
3 4 1.90
4 4 -3.00
5 4 -1.20
6 4 -3.10
7 4 2.40
8 4 3.20
9 4 1.70
10 4 1.80
0 3 7.00
1 3 3.60
2 3 -1.10
3 3 3.70
4 3 2.70
5 3 -3.70
6 3 -2.60
7 3 -3.70
8 3 -0.70
9 3 1.50
10 3 0.00
0 2 -1.20
1 2 1.20
2 2 0.10
3 2 -3.10
4 2 -1.90
5 2 -5.40
6 2 0.40
7 2 -4.80
8 2 -1.70
9 2 -3.20
10 2 -0.70
0 1 -0.10
1 1 -2.20
2 1 -2.90
3 1 -3.90
4 1 -4.80
5 1 -4.60
6 1 -7.20
7 1 -3.20
8 1 -0.60
9 1 -1.20
10 1 -3.80
0 0 -4.30
1 0 -1.20
2 0 -2.90
3 0 -2.90
4 0 -3.40
5 0 -6.60
6 0 -4.70
7 0 -4.00
8 0 -4.00
9 0 0.70
10 0 -7.10
};
\node[font=\fontsize{4}{4.8}\selectfont] at (axis cs:0,0) {-4.3};
\node[font=\fontsize{4}{4.8}\selectfont] at (axis cs:1,0) {-1.2};
\node[font=\fontsize{4}{4.8}\selectfont] at (axis cs:2,0) {-2.9};
\node[font=\fontsize{4}{4.8}\selectfont] at (axis cs:3,0) {-2.9};
\node[font=\fontsize{4}{4.8}\selectfont] at (axis cs:4,0) {-3.4};
\node[font=\fontsize{4}{4.8}\selectfont] at (axis cs:5,0) {-6.6};
\node[font=\fontsize{4}{4.8}\selectfont] at (axis cs:6,0) {-4.7};
\node[font=\fontsize{4}{4.8}\selectfont] at (axis cs:7,0) {-4.0};
\node[font=\fontsize{4}{4.8}\selectfont] at (axis cs:8,0) {-4.0};
\node[font=\fontsize{4}{4.8}\selectfont] at (axis cs:9,0) {0.7};
\node[font=\fontsize{4}{4.8}\selectfont] at (axis cs:10,0) {-7.1};
\node[font=\fontsize{4}{4.8}\selectfont] at (axis cs:0,1) {-0.1};
\node[font=\fontsize{4}{4.8}\selectfont] at (axis cs:1,1) {-2.2};
\node[font=\fontsize{4}{4.8}\selectfont] at (axis cs:2,1) {-2.9};
\node[font=\fontsize{4}{4.8}\selectfont] at (axis cs:3,1) {-3.9};
\node[font=\fontsize{4}{4.8}\selectfont] at (axis cs:4,1) {-4.8};
\node[font=\fontsize{4}{4.8}\selectfont] at (axis cs:5,1) {-4.6};
\node[font=\fontsize{4}{4.8}\selectfont] at (axis cs:6,1) {-7.2};
\node[font=\fontsize{4}{4.8}\selectfont] at (axis cs:7,1) {-3.2};
\node[font=\fontsize{4}{4.8}\selectfont] at (axis cs:8,1) {-0.6};
\node[font=\fontsize{4}{4.8}\selectfont] at (axis cs:9,1) {-1.2};
\node[font=\fontsize{4}{4.8}\selectfont] at (axis cs:10,1) {-3.8};
\node[font=\fontsize{4}{4.8}\selectfont] at (axis cs:0,2) {-1.2};
\node[font=\fontsize{4}{4.8}\selectfont] at (axis cs:1,2) {1.2};
\node[font=\fontsize{4}{4.8}\selectfont] at (axis cs:2,2) {0.1};
\node[font=\fontsize{4}{4.8}\selectfont] at (axis cs:3,2) {-3.1};
\node[font=\fontsize{4}{4.8}\selectfont] at (axis cs:4,2) {-1.9};
\node[font=\fontsize{4}{4.8}\selectfont] at (axis cs:5,2) {-5.4};
\node[font=\fontsize{4}{4.8}\selectfont] at (axis cs:6,2) {0.4};
\node[font=\fontsize{4}{4.8}\selectfont] at (axis cs:7,2) {-4.8};
\node[font=\fontsize{4}{4.8}\selectfont] at (axis cs:8,2) {-1.7};
\node[font=\fontsize{4}{4.8}\selectfont] at (axis cs:9,2) {-3.2};
\node[font=\fontsize{4}{4.8}\selectfont] at (axis cs:10,2) {-0.7};
\node[font=\fontsize{4}{4.8}\selectfont] at (axis cs:0,3) {7.0};
\node[font=\fontsize{4}{4.8}\selectfont] at (axis cs:1,3) {3.6};
\node[font=\fontsize{4}{4.8}\selectfont] at (axis cs:2,3) {-1.1};
\node[font=\fontsize{4}{4.8}\selectfont] at (axis cs:3,3) {3.7};
\node[font=\fontsize{4}{4.8}\selectfont] at (axis cs:4,3) {2.7};
\node[font=\fontsize{4}{4.8}\selectfont] at (axis cs:5,3) {-3.7};
\node[font=\fontsize{4}{4.8}\selectfont] at (axis cs:6,3) {-2.6};
\node[font=\fontsize{4}{4.8}\selectfont] at (axis cs:7,3) {-3.7};
\node[font=\fontsize{4}{4.8}\selectfont] at (axis cs:8,3) {-0.7};
\node[font=\fontsize{4}{4.8}\selectfont] at (axis cs:9,3) {1.5};
\node[font=\fontsize{4}{4.8}\selectfont] at (axis cs:10,3) {0.0};
\node[font=\fontsize{4}{4.8}\selectfont] at (axis cs:0,4) {-0.7};
\node[font=\fontsize{4}{4.8}\selectfont] at (axis cs:1,4) {2.0};
\node[font=\fontsize{4}{4.8}\selectfont] at (axis cs:2,4) {3.2};
\node[font=\fontsize{4}{4.8}\selectfont] at (axis cs:3,4) {1.9};
\node[font=\fontsize{4}{4.8}\selectfont] at (axis cs:4,4) {-3.0};
\node[font=\fontsize{4}{4.8}\selectfont] at (axis cs:5,4) {-1.2};
\node[font=\fontsize{4}{4.8}\selectfont] at (axis cs:6,4) {-3.1};
\node[font=\fontsize{4}{4.8}\selectfont] at (axis cs:7,4) {2.4};
\node[font=\fontsize{4}{4.8}\selectfont] at (axis cs:8,4) {3.2};
\node[font=\fontsize{4}{4.8}\selectfont] at (axis cs:9,4) {1.7};
\node[font=\fontsize{4}{4.8}\selectfont] at (axis cs:10,4) {1.8};
\node[font=\fontsize{4}{4.8}\selectfont] at (axis cs:0,5) {3.4};
\node[font=\fontsize{4}{4.8}\selectfont] at (axis cs:1,5) {2.0};
\node[font=\fontsize{4}{4.8}\selectfont] at (axis cs:2,5) {2.5};
\node[font=\fontsize{4}{4.8}\selectfont] at (axis cs:3,5) {2.3};
\node[font=\fontsize{4}{4.8}\selectfont] at (axis cs:4,5) {-0.7};
\node[font=\fontsize{4}{4.8}\selectfont] at (axis cs:5,5) {4.0};
\node[font=\fontsize{4}{4.8}\selectfont] at (axis cs:6,5) {-2.0};
\node[font=\fontsize{4}{4.8}\selectfont] at (axis cs:7,5) {2.4};
\node[font=\fontsize{4}{4.8}\selectfont] at (axis cs:8,5) {5.2};
\node[font=\fontsize{4}{4.8}\selectfont] at (axis cs:9,5) {5.0};
\node[font=\fontsize{4}{4.8}\selectfont] at (axis cs:10,5) {2.8};
\node[font=\fontsize{4}{4.8}\selectfont] at (axis cs:0,6) {8.7};
\node[font=\fontsize{4}{4.8}\selectfont] at (axis cs:1,6) {5.6};
\node[font=\fontsize{4}{4.8}\selectfont] at (axis cs:2,6) {4.0};
\node[font=\fontsize{4}{4.8}\selectfont] at (axis cs:3,6) {1.3};
\node[font=\fontsize{4}{4.8}\selectfont] at (axis cs:4,6) {5.3};
\node[font=\fontsize{4}{4.8}\selectfont] at (axis cs:5,6) {3.1};
\node[font=\fontsize{4}{4.8}\selectfont] at (axis cs:6,6) {3.1};
\node[font=\fontsize{4}{4.8}\selectfont] at (axis cs:7,6) {0.8};
\node[font=\fontsize{4}{4.8}\selectfont] at (axis cs:8,6) {4.8};
\node[font=\fontsize{4}{4.8}\selectfont] at (axis cs:9,6) {3.6};
\node[font=\fontsize{4}{4.8}\selectfont] at (axis cs:10,6) {4.8};
\node[font=\fontsize{4}{4.8}\selectfont] at (axis cs:0,7) {7.4};
\node[font=\fontsize{4}{4.8}\selectfont] at (axis cs:1,7) {6.2};
\node[font=\fontsize{4}{4.8}\selectfont] at (axis cs:2,7) {0.7};
\node[font=\fontsize{4}{4.8}\selectfont] at (axis cs:3,7) {5.3};
\node[font=\fontsize{4}{4.8}\selectfont] at (axis cs:4,7) {0.6};
\node[font=\fontsize{4}{4.8}\selectfont] at (axis cs:5,7) {2.4};
\node[font=\fontsize{4}{4.8}\selectfont] at (axis cs:6,7) {3.7};
\node[font=\fontsize{4}{4.8}\selectfont] at (axis cs:7,7) {3.8};
\node[font=\fontsize{4}{4.8}\selectfont] at (axis cs:8,7) {4.7};
\node[font=\fontsize{4}{4.8}\selectfont] at (axis cs:9,7) {7.6};
\node[font=\fontsize{4}{4.8}\selectfont] at (axis cs:10,7) {7.6};
\node[font=\fontsize{4}{4.8}\selectfont] at (axis cs:0,8) {3.9};
\node[font=\fontsize{4}{4.8}\selectfont] at (axis cs:1,8) {3.2};
\node[font=\fontsize{4}{4.8}\selectfont] at (axis cs:2,8) {4.1};
\node[font=\fontsize{4}{4.8}\selectfont] at (axis cs:3,8) {3.1};
\node[font=\fontsize{4}{4.8}\selectfont] at (axis cs:4,8) {5.6};
\node[font=\fontsize{4}{4.8}\selectfont] at (axis cs:5,8) {3.8};
\node[font=\fontsize{4}{4.8}\selectfont] at (axis cs:6,8) {4.0};
\node[font=\fontsize{4}{4.8}\selectfont] at (axis cs:7,8) {1.3};
\node[font=\fontsize{4}{4.8}\selectfont] at (axis cs:8,8) {3.2};
\node[font=\fontsize{4}{4.8}\selectfont] at (axis cs:9,8) {5.6};
\node[font=\fontsize{4}{4.8}\selectfont] at (axis cs:10,8) {3.3};
\node[font=\fontsize{4}{4.8}\selectfont] at (axis cs:0,9) {2.7};
\node[font=\fontsize{4}{4.8}\selectfont] at (axis cs:1,9) {5.2};
\node[font=\fontsize{4}{4.8}\selectfont] at (axis cs:2,9) {5.1};
\node[font=\fontsize{4}{4.8}\selectfont] at (axis cs:3,9) {-0.6};
\node[font=\fontsize{4}{4.8}\selectfont] at (axis cs:4,9) {4.2};
\node[font=\fontsize{4}{4.8}\selectfont] at (axis cs:5,9) {3.9};
\node[font=\fontsize{4}{4.8}\selectfont] at (axis cs:6,9) {3.4};
\node[font=\fontsize{4}{4.8}\selectfont] at (axis cs:7,9) {6.6};
\node[font=\fontsize{4}{4.8}\selectfont] at (axis cs:8,9) {4.9};
\node[font=\fontsize{4}{4.8}\selectfont] at (axis cs:9,9) {3.5};
\node[font=\fontsize{4}{4.8}\selectfont] at (axis cs:10,9) {3.9};
\node[font=\fontsize{4}{4.8}\selectfont] at (axis cs:0,10) {6.1};
\node[font=\fontsize{4}{4.8}\selectfont] at (axis cs:1,10) {3.8};
\node[font=\fontsize{4}{4.8}\selectfont] at (axis cs:2,10) {5.0};
\node[font=\fontsize{4}{4.8}\selectfont] at (axis cs:3,10) {1.7};
\node[font=\fontsize{4}{4.8}\selectfont] at (axis cs:4,10) {1.4};
\node[font=\fontsize{4}{4.8}\selectfont] at (axis cs:5,10) {0.8};
\node[font=\fontsize{4}{4.8}\selectfont] at (axis cs:6,10) {2.0};
\node[font=\fontsize{4}{4.8}\selectfont] at (axis cs:7,10) {-1.0};
\node[font=\fontsize{4}{4.8}\selectfont] at (axis cs:8,10) {-0.1};
\node[font=\fontsize{4}{4.8}\selectfont] at (axis cs:9,10) {1.5};
\node[font=\fontsize{4}{4.8}\selectfont] at (axis cs:10,10) {4.3};
\end{axis}
\end{tikzpicture}
            \end{minipage}%
            \hfill
            \begin{minipage}[t]{0.48\textwidth}
                \input{plots/11_appendix/few_shot/delta/olmo32_rotation_permutation}
            \end{minipage}
        \end{minipage}}
            \vspace{-5pt}
        \caption{\small
            Same as Figure~\ref{fig:icl_deltas_qwen32}, for \model{olmo32}.
        }
        \vspace{-8pt}
        \label{fig:few_shot_delta_aliases_olmo32}
    \end{figure}
    \FloatBarrier

    \begin{figure}[ht]
        \centering
        \scalebox{0.77}{\begin{minipage}{\textwidth}\centering
            \begin{minipage}[t]{0.48\textwidth}
                \input{plots/11_appendix/few_shot/delta/qwen14_dropout_noise}
            \end{minipage}%
            \hfill
            \begin{minipage}[t]{0.48\textwidth}
                \input{plots/11_appendix/few_shot/delta/qwen14_masking_jitter}
            \end{minipage}\\[0pt]
            \begin{minipage}[t]{0.48\textwidth}
                \input{plots/11_appendix/few_shot/delta/qwen14_x_y}
            \end{minipage}%
            \hfill
            \begin{minipage}[t]{0.48\textwidth}
                \input{plots/11_appendix/few_shot/delta/qwen14_rotation_permutation}
            \end{minipage}
        \end{minipage}}
            \vspace{-5pt}
        \caption{\small
            Same as Figure~\ref{fig:icl_deltas_qwen32}, for \model{qwen14}.
        }
        \vspace{-8pt}
        \label{fig:few_shot_delta_aliases_qwen14}
    \end{figure}
    \FloatBarrier

    \begin{figure}[ht]
        \centering
        \scalebox{0.77}{\begin{minipage}{\textwidth}\centering
            \begin{minipage}[t]{0.48\textwidth}
                \input{plots/11_appendix/few_shot/delta/qwen27_dropout_noise}
            \end{minipage}%
            \hfill
            \begin{minipage}[t]{0.48\textwidth}
                \input{plots/11_appendix/few_shot/delta/qwen27_masking_jitter}
            \end{minipage}\\[0pt]
            \begin{minipage}[t]{0.48\textwidth}
                \input{plots/11_appendix/few_shot/delta/qwen27_x_y}
            \end{minipage}%
            \hfill
            \begin{minipage}[t]{0.48\textwidth}
                \input{plots/11_appendix/few_shot/delta/qwen27_rotation_permutation}
            \end{minipage}
        \end{minipage}}
            \vspace{-5pt}
        \caption{\small
            Same as Figure~\ref{fig:icl_deltas_qwen32}, for \model{qwen27}.
        }
        \vspace{-8pt}
        \label{fig:few_shot_delta_aliases_qwen27}
    \end{figure}
    \FloatBarrier

    \begin{figure}[ht]
        \centering
        \scalebox{0.77}{\begin{minipage}{\textwidth}\centering
            \begin{minipage}[t]{0.48\textwidth}
                \input{plots/11_appendix/few_shot/delta/seed36_dropout_noise}
            \end{minipage}%
            \hfill
            \begin{minipage}[t]{0.48\textwidth}
                \centering
\begin{tikzpicture}
\begin{axis}[
    width=0.82\linewidth,
    height=0.82\linewidth,
    xlabel={},
    ylabel={},
    yticklabels={},
    colorbar,
    colormap={pastel_po}{rgb255(0cm)=(225,160,100); rgb255(50cm)=(255,255,255); rgb255(100cm)=(160,135,200)},
    point meta min=-25.0,
    point meta max=25.0,
    xtick={0,1,2,3,4,5,6,7,8,9,10},
    xticklabels={$p_{\min}$,,,,,,,,,,$p_{\max}$},
    ytick={0,1,2,3,4,5,6,7,8,9,10},
    x tick label style={font=\scriptsize},
    y tick label style={font=\scriptsize},
    view={0}{90},
    axis line style={draw=none},
    major tick length=0pt,
    enlargelimits=false,
    colorbar style={font=\tiny},
    title style={font=\small},
    title={Masking/Jitter},
    title style={yshift=-1.0ex},
]
\addplot[matrix plot*, mesh/cols=11, mesh/rows=11, point meta=explicit] table[meta=C] {
x y C
0 10 3.40
1 10 0.40
2 10 -2.00
3 10 -3.50
4 10 -3.20
5 10 -2.30
6 10 1.00
7 10 5.50
8 10 8.70
9 10 11.30
10 10 9.40
0 9 3.20
1 9 0.40
2 9 -0.60
3 9 -3.20
4 9 -3.80
5 9 -3.00
6 9 -1.30
7 9 5.20
8 9 7.60
9 9 8.50
10 9 9.90
0 8 -0.70
1 8 1.50
2 8 -0.90
3 8 -3.30
4 8 -3.10
5 8 0.70
6 8 1.10
7 8 6.20
8 8 8.30
9 8 10.10
10 8 9.40
0 7 2.90
1 7 2.30
2 7 -0.80
3 7 -2.40
4 7 -2.10
5 7 0.80
6 7 2.70
7 7 4.20
8 7 8.80
9 7 8.50
10 7 9.00
0 6 3.00
1 6 5.10
2 6 2.00
3 6 -1.10
4 6 -1.10
5 6 3.50
6 6 4.00
7 6 4.40
8 6 11.30
9 6 11.90
10 6 10.90
0 5 4.90
1 5 3.90
2 5 7.10
3 5 1.40
4 5 3.50
5 5 2.90
6 5 4.80
7 5 5.10
8 5 10.20
9 5 9.60
10 5 11.20
0 4 7.60
1 4 8.70
2 4 10.40
3 4 4.80
4 4 4.10
5 4 3.70
6 4 5.10
7 4 5.20
8 4 9.20
9 4 13.00
10 4 14.40
0 3 8.70
1 3 7.00
2 3 4.90
3 3 5.60
4 3 5.60
5 3 3.80
6 3 1.90
7 3 4.00
8 3 8.60
9 3 13.30
10 3 14.50
0 2 3.90
1 2 3.00
2 2 4.30
3 2 3.30
4 2 2.10
5 2 -0.60
6 2 -0.30
7 2 2.80
8 2 7.60
9 2 14.30
10 2 16.10
0 1 1.20
1 1 0.50
2 1 -1.30
3 1 0.80
4 1 0.90
5 1 -2.20
6 1 -1.30
7 1 1.80
8 1 5.70
9 1 7.20
10 1 11.10
0 0 -0.40
1 0 -2.60
2 0 3.60
3 0 -0.30
4 0 -1.60
5 0 -2.90
6 0 -8.10
7 0 -0.90
8 0 4.20
9 0 8.40
10 0 12.90
};
\node[font=\fontsize{4}{4.8}\selectfont] at (axis cs:0,0) {-0.4};
\node[font=\fontsize{4}{4.8}\selectfont] at (axis cs:1,0) {-2.6};
\node[font=\fontsize{4}{4.8}\selectfont] at (axis cs:2,0) {3.6};
\node[font=\fontsize{4}{4.8}\selectfont] at (axis cs:3,0) {-0.3};
\node[font=\fontsize{4}{4.8}\selectfont] at (axis cs:4,0) {-1.6};
\node[font=\fontsize{4}{4.8}\selectfont] at (axis cs:5,0) {-2.9};
\node[font=\fontsize{4}{4.8}\selectfont] at (axis cs:6,0) {-8.1};
\node[font=\fontsize{4}{4.8}\selectfont] at (axis cs:7,0) {-0.9};
\node[font=\fontsize{4}{4.8}\selectfont] at (axis cs:8,0) {4.2};
\node[font=\fontsize{4}{4.8}\selectfont] at (axis cs:9,0) {8.4};
\node[font=\fontsize{4}{4.8}\selectfont] at (axis cs:10,0) {12.9};
\node[font=\fontsize{4}{4.8}\selectfont] at (axis cs:0,1) {1.2};
\node[font=\fontsize{4}{4.8}\selectfont] at (axis cs:1,1) {0.5};
\node[font=\fontsize{4}{4.8}\selectfont] at (axis cs:2,1) {-1.3};
\node[font=\fontsize{4}{4.8}\selectfont] at (axis cs:3,1) {0.8};
\node[font=\fontsize{4}{4.8}\selectfont] at (axis cs:4,1) {0.9};
\node[font=\fontsize{4}{4.8}\selectfont] at (axis cs:5,1) {-2.2};
\node[font=\fontsize{4}{4.8}\selectfont] at (axis cs:6,1) {-1.3};
\node[font=\fontsize{4}{4.8}\selectfont] at (axis cs:7,1) {1.8};
\node[font=\fontsize{4}{4.8}\selectfont] at (axis cs:8,1) {5.7};
\node[font=\fontsize{4}{4.8}\selectfont] at (axis cs:9,1) {7.2};
\node[font=\fontsize{4}{4.8}\selectfont] at (axis cs:10,1) {11.1};
\node[font=\fontsize{4}{4.8}\selectfont] at (axis cs:0,2) {3.9};
\node[font=\fontsize{4}{4.8}\selectfont] at (axis cs:1,2) {3.0};
\node[font=\fontsize{4}{4.8}\selectfont] at (axis cs:2,2) {4.3};
\node[font=\fontsize{4}{4.8}\selectfont] at (axis cs:3,2) {3.3};
\node[font=\fontsize{4}{4.8}\selectfont] at (axis cs:4,2) {2.1};
\node[font=\fontsize{4}{4.8}\selectfont] at (axis cs:5,2) {-0.6};
\node[font=\fontsize{4}{4.8}\selectfont] at (axis cs:6,2) {-0.3};
\node[font=\fontsize{4}{4.8}\selectfont] at (axis cs:7,2) {2.8};
\node[font=\fontsize{4}{4.8}\selectfont] at (axis cs:8,2) {7.6};
\node[font=\fontsize{4}{4.8}\selectfont] at (axis cs:9,2) {14.3};
\node[font=\fontsize{4}{4.8}\selectfont] at (axis cs:10,2) {16.1};
\node[font=\fontsize{4}{4.8}\selectfont] at (axis cs:0,3) {8.7};
\node[font=\fontsize{4}{4.8}\selectfont] at (axis cs:1,3) {7.0};
\node[font=\fontsize{4}{4.8}\selectfont] at (axis cs:2,3) {4.9};
\node[font=\fontsize{4}{4.8}\selectfont] at (axis cs:3,3) {5.6};
\node[font=\fontsize{4}{4.8}\selectfont] at (axis cs:4,3) {5.6};
\node[font=\fontsize{4}{4.8}\selectfont] at (axis cs:5,3) {3.8};
\node[font=\fontsize{4}{4.8}\selectfont] at (axis cs:6,3) {1.9};
\node[font=\fontsize{4}{4.8}\selectfont] at (axis cs:7,3) {4.0};
\node[font=\fontsize{4}{4.8}\selectfont] at (axis cs:8,3) {8.6};
\node[font=\fontsize{4}{4.8}\selectfont] at (axis cs:9,3) {13.3};
\node[font=\fontsize{4}{4.8}\selectfont] at (axis cs:10,3) {14.5};
\node[font=\fontsize{4}{4.8}\selectfont] at (axis cs:0,4) {7.6};
\node[font=\fontsize{4}{4.8}\selectfont] at (axis cs:1,4) {8.7};
\node[font=\fontsize{4}{4.8}\selectfont] at (axis cs:2,4) {10.4};
\node[font=\fontsize{4}{4.8}\selectfont] at (axis cs:3,4) {4.8};
\node[font=\fontsize{4}{4.8}\selectfont] at (axis cs:4,4) {4.1};
\node[font=\fontsize{4}{4.8}\selectfont] at (axis cs:5,4) {3.7};
\node[font=\fontsize{4}{4.8}\selectfont] at (axis cs:6,4) {5.1};
\node[font=\fontsize{4}{4.8}\selectfont] at (axis cs:7,4) {5.2};
\node[font=\fontsize{4}{4.8}\selectfont] at (axis cs:8,4) {9.2};
\node[font=\fontsize{4}{4.8}\selectfont] at (axis cs:9,4) {13.0};
\node[font=\fontsize{4}{4.8}\selectfont] at (axis cs:10,4) {14.4};
\node[font=\fontsize{4}{4.8}\selectfont] at (axis cs:0,5) {4.9};
\node[font=\fontsize{4}{4.8}\selectfont] at (axis cs:1,5) {3.9};
\node[font=\fontsize{4}{4.8}\selectfont] at (axis cs:2,5) {7.1};
\node[font=\fontsize{4}{4.8}\selectfont] at (axis cs:3,5) {1.4};
\node[font=\fontsize{4}{4.8}\selectfont] at (axis cs:4,5) {3.5};
\node[font=\fontsize{4}{4.8}\selectfont] at (axis cs:5,5) {2.9};
\node[font=\fontsize{4}{4.8}\selectfont] at (axis cs:6,5) {4.8};
\node[font=\fontsize{4}{4.8}\selectfont] at (axis cs:7,5) {5.1};
\node[font=\fontsize{4}{4.8}\selectfont] at (axis cs:8,5) {10.2};
\node[font=\fontsize{4}{4.8}\selectfont] at (axis cs:9,5) {9.6};
\node[font=\fontsize{4}{4.8}\selectfont] at (axis cs:10,5) {11.2};
\node[font=\fontsize{4}{4.8}\selectfont] at (axis cs:0,6) {3.0};
\node[font=\fontsize{4}{4.8}\selectfont] at (axis cs:1,6) {5.1};
\node[font=\fontsize{4}{4.8}\selectfont] at (axis cs:2,6) {2.0};
\node[font=\fontsize{4}{4.8}\selectfont] at (axis cs:3,6) {-1.1};
\node[font=\fontsize{4}{4.8}\selectfont] at (axis cs:4,6) {-1.1};
\node[font=\fontsize{4}{4.8}\selectfont] at (axis cs:5,6) {3.5};
\node[font=\fontsize{4}{4.8}\selectfont] at (axis cs:6,6) {4.0};
\node[font=\fontsize{4}{4.8}\selectfont] at (axis cs:7,6) {4.4};
\node[font=\fontsize{4}{4.8}\selectfont] at (axis cs:8,6) {11.3};
\node[font=\fontsize{4}{4.8}\selectfont] at (axis cs:9,6) {11.9};
\node[font=\fontsize{4}{4.8}\selectfont] at (axis cs:10,6) {10.9};
\node[font=\fontsize{4}{4.8}\selectfont] at (axis cs:0,7) {2.9};
\node[font=\fontsize{4}{4.8}\selectfont] at (axis cs:1,7) {2.3};
\node[font=\fontsize{4}{4.8}\selectfont] at (axis cs:2,7) {-0.8};
\node[font=\fontsize{4}{4.8}\selectfont] at (axis cs:3,7) {-2.4};
\node[font=\fontsize{4}{4.8}\selectfont] at (axis cs:4,7) {-2.1};
\node[font=\fontsize{4}{4.8}\selectfont] at (axis cs:5,7) {0.8};
\node[font=\fontsize{4}{4.8}\selectfont] at (axis cs:6,7) {2.7};
\node[font=\fontsize{4}{4.8}\selectfont] at (axis cs:7,7) {4.2};
\node[font=\fontsize{4}{4.8}\selectfont] at (axis cs:8,7) {8.8};
\node[font=\fontsize{4}{4.8}\selectfont] at (axis cs:9,7) {8.5};
\node[font=\fontsize{4}{4.8}\selectfont] at (axis cs:10,7) {9.0};
\node[font=\fontsize{4}{4.8}\selectfont] at (axis cs:0,8) {-0.7};
\node[font=\fontsize{4}{4.8}\selectfont] at (axis cs:1,8) {1.5};
\node[font=\fontsize{4}{4.8}\selectfont] at (axis cs:2,8) {-0.9};
\node[font=\fontsize{4}{4.8}\selectfont] at (axis cs:3,8) {-3.3};
\node[font=\fontsize{4}{4.8}\selectfont] at (axis cs:4,8) {-3.1};
\node[font=\fontsize{4}{4.8}\selectfont] at (axis cs:5,8) {0.7};
\node[font=\fontsize{4}{4.8}\selectfont] at (axis cs:6,8) {1.1};
\node[font=\fontsize{4}{4.8}\selectfont] at (axis cs:7,8) {6.2};
\node[font=\fontsize{4}{4.8}\selectfont] at (axis cs:8,8) {8.3};
\node[font=\fontsize{4}{4.8}\selectfont] at (axis cs:9,8) {10.1};
\node[font=\fontsize{4}{4.8}\selectfont] at (axis cs:10,8) {9.4};
\node[font=\fontsize{4}{4.8}\selectfont] at (axis cs:0,9) {3.2};
\node[font=\fontsize{4}{4.8}\selectfont] at (axis cs:1,9) {0.4};
\node[font=\fontsize{4}{4.8}\selectfont] at (axis cs:2,9) {-0.6};
\node[font=\fontsize{4}{4.8}\selectfont] at (axis cs:3,9) {-3.2};
\node[font=\fontsize{4}{4.8}\selectfont] at (axis cs:4,9) {-3.8};
\node[font=\fontsize{4}{4.8}\selectfont] at (axis cs:5,9) {-3.0};
\node[font=\fontsize{4}{4.8}\selectfont] at (axis cs:6,9) {-1.3};
\node[font=\fontsize{4}{4.8}\selectfont] at (axis cs:7,9) {5.2};
\node[font=\fontsize{4}{4.8}\selectfont] at (axis cs:8,9) {7.6};
\node[font=\fontsize{4}{4.8}\selectfont] at (axis cs:9,9) {8.5};
\node[font=\fontsize{4}{4.8}\selectfont] at (axis cs:10,9) {9.9};
\node[font=\fontsize{4}{4.8}\selectfont] at (axis cs:0,10) {3.4};
\node[font=\fontsize{4}{4.8}\selectfont] at (axis cs:1,10) {0.4};
\node[font=\fontsize{4}{4.8}\selectfont] at (axis cs:2,10) {-2.0};
\node[font=\fontsize{4}{4.8}\selectfont] at (axis cs:3,10) {-3.5};
\node[font=\fontsize{4}{4.8}\selectfont] at (axis cs:4,10) {-3.2};
\node[font=\fontsize{4}{4.8}\selectfont] at (axis cs:5,10) {-2.3};
\node[font=\fontsize{4}{4.8}\selectfont] at (axis cs:6,10) {1.0};
\node[font=\fontsize{4}{4.8}\selectfont] at (axis cs:7,10) {5.5};
\node[font=\fontsize{4}{4.8}\selectfont] at (axis cs:8,10) {8.7};
\node[font=\fontsize{4}{4.8}\selectfont] at (axis cs:9,10) {11.3};
\node[font=\fontsize{4}{4.8}\selectfont] at (axis cs:10,10) {9.4};
\end{axis}
\end{tikzpicture}
            \end{minipage}\\[0pt]
            \begin{minipage}[t]{0.48\textwidth}
                \centering
\begin{tikzpicture}
\begin{axis}[
    width=0.82\linewidth,
    height=0.82\linewidth,
    xlabel={\small Dropout rate $p$},
    ylabel={\small Noise SD $\sigma$},
    colormap={pastel_po}{rgb255(0cm)=(225,160,100); rgb255(50cm)=(255,255,255); rgb255(100cm)=(160,135,200)},
    point meta min=-25.0,
    point meta max=25.0,
    xtick={0,1,2,3,4,5,6,7,8,9,10},
    xticklabels={$p_{\min}$,,,,,,,,,,$p_{\max}$},
    ytick={0,1,2,3,4,5,6,7,8,9,10},
    yticklabels={$\sigma_{\min}$,,,,,,,,,,$\sigma_{\max}$},
    x tick label style={font=\scriptsize},
    y tick label style={font=\scriptsize},
    view={0}{90},
    axis line style={draw=none},
    major tick length=0pt,
    enlargelimits=false,
    title style={font=\small},
    title={X/Y},
    title style={yshift=-1.0ex},
]
\addplot[matrix plot*, mesh/cols=11, mesh/rows=11, point meta=explicit] table[meta=C] {
x y C
0 10 5.80
1 10 8.10
2 10 7.70
3 10 9.10
4 10 10.50
5 10 6.50
6 10 3.20
7 10 3.70
8 10 4.70
9 10 0.50
10 10 3.00
0 9 4.10
1 9 3.10
2 9 5.30
3 9 10.50
4 9 10.00
5 9 9.90
6 9 5.60
7 9 4.30
8 9 5.40
9 9 -1.00
10 9 3.10
0 8 2.20
1 8 1.80
2 8 5.50
3 8 4.60
4 8 6.70
5 8 7.70
6 8 0.40
7 8 6.70
8 8 2.00
9 8 5.70
10 8 6.90
0 7 1.80
1 7 5.70
2 7 6.10
3 7 5.80
4 7 6.90
5 7 5.00
6 7 5.20
7 7 6.10
8 7 2.50
9 7 2.10
10 7 4.10
0 6 3.20
1 6 3.20
2 6 2.80
3 6 4.20
4 6 8.90
5 6 0.90
6 6 1.20
7 6 0.00
8 6 -0.30
9 6 1.80
10 6 3.40
0 5 4.40
1 5 3.10
2 5 4.20
3 5 5.90
4 5 3.90
5 5 4.40
6 5 2.80
7 5 0.20
8 5 1.00
9 5 2.40
10 5 1.60
0 4 5.00
1 4 0.60
2 4 5.40
3 4 2.90
4 4 1.90
5 4 5.20
6 4 1.10
7 4 0.70
8 4 3.10
9 4 1.00
10 4 1.90
0 3 2.80
1 3 -2.80
2 3 4.00
3 3 -0.70
4 3 2.90
5 3 -0.30
6 3 0.60
7 3 2.50
8 3 -1.80
9 3 -0.10
10 3 -0.10
0 2 -0.30
1 2 2.10
2 2 0.90
3 2 2.10
4 2 2.50
5 2 -1.90
6 2 -3.70
7 2 -3.30
8 2 -2.30
9 2 -4.30
10 2 -3.80
0 1 2.50
1 1 2.70
2 1 1.00
3 1 -1.40
4 1 -3.00
5 1 0.20
6 1 -3.70
7 1 -5.50
8 1 -8.90
9 1 -7.50
10 1 -7.50
0 0 -1.80
1 0 1.30
2 0 -0.80
3 0 0.70
4 0 0.50
5 0 -1.90
6 0 -6.10
7 0 -5.10
8 0 -7.90
9 0 -7.70
10 0 -6.60
};
\node[font=\fontsize{4}{4.8}\selectfont] at (axis cs:0,0) {-1.8};
\node[font=\fontsize{4}{4.8}\selectfont] at (axis cs:1,0) {1.3};
\node[font=\fontsize{4}{4.8}\selectfont] at (axis cs:2,0) {-0.8};
\node[font=\fontsize{4}{4.8}\selectfont] at (axis cs:3,0) {0.7};
\node[font=\fontsize{4}{4.8}\selectfont] at (axis cs:4,0) {0.5};
\node[font=\fontsize{4}{4.8}\selectfont] at (axis cs:5,0) {-1.9};
\node[font=\fontsize{4}{4.8}\selectfont] at (axis cs:6,0) {-6.1};
\node[font=\fontsize{4}{4.8}\selectfont] at (axis cs:7,0) {-5.1};
\node[font=\fontsize{4}{4.8}\selectfont] at (axis cs:8,0) {-7.9};
\node[font=\fontsize{4}{4.8}\selectfont] at (axis cs:9,0) {-7.7};
\node[font=\fontsize{4}{4.8}\selectfont] at (axis cs:10,0) {-6.6};
\node[font=\fontsize{4}{4.8}\selectfont] at (axis cs:0,1) {2.5};
\node[font=\fontsize{4}{4.8}\selectfont] at (axis cs:1,1) {2.7};
\node[font=\fontsize{4}{4.8}\selectfont] at (axis cs:2,1) {1.0};
\node[font=\fontsize{4}{4.8}\selectfont] at (axis cs:3,1) {-1.4};
\node[font=\fontsize{4}{4.8}\selectfont] at (axis cs:4,1) {-3.0};
\node[font=\fontsize{4}{4.8}\selectfont] at (axis cs:5,1) {0.2};
\node[font=\fontsize{4}{4.8}\selectfont] at (axis cs:6,1) {-3.7};
\node[font=\fontsize{4}{4.8}\selectfont] at (axis cs:7,1) {-5.5};
\node[font=\fontsize{4}{4.8}\selectfont] at (axis cs:8,1) {-8.9};
\node[font=\fontsize{4}{4.8}\selectfont] at (axis cs:9,1) {-7.5};
\node[font=\fontsize{4}{4.8}\selectfont] at (axis cs:10,1) {-7.5};
\node[font=\fontsize{4}{4.8}\selectfont] at (axis cs:0,2) {-0.3};
\node[font=\fontsize{4}{4.8}\selectfont] at (axis cs:1,2) {2.1};
\node[font=\fontsize{4}{4.8}\selectfont] at (axis cs:2,2) {0.9};
\node[font=\fontsize{4}{4.8}\selectfont] at (axis cs:3,2) {2.1};
\node[font=\fontsize{4}{4.8}\selectfont] at (axis cs:4,2) {2.5};
\node[font=\fontsize{4}{4.8}\selectfont] at (axis cs:5,2) {-1.9};
\node[font=\fontsize{4}{4.8}\selectfont] at (axis cs:6,2) {-3.7};
\node[font=\fontsize{4}{4.8}\selectfont] at (axis cs:7,2) {-3.3};
\node[font=\fontsize{4}{4.8}\selectfont] at (axis cs:8,2) {-2.3};
\node[font=\fontsize{4}{4.8}\selectfont] at (axis cs:9,2) {-4.3};
\node[font=\fontsize{4}{4.8}\selectfont] at (axis cs:10,2) {-3.8};
\node[font=\fontsize{4}{4.8}\selectfont] at (axis cs:0,3) {2.8};
\node[font=\fontsize{4}{4.8}\selectfont] at (axis cs:1,3) {-2.8};
\node[font=\fontsize{4}{4.8}\selectfont] at (axis cs:2,3) {4.0};
\node[font=\fontsize{4}{4.8}\selectfont] at (axis cs:3,3) {-0.7};
\node[font=\fontsize{4}{4.8}\selectfont] at (axis cs:4,3) {2.9};
\node[font=\fontsize{4}{4.8}\selectfont] at (axis cs:5,3) {-0.3};
\node[font=\fontsize{4}{4.8}\selectfont] at (axis cs:6,3) {0.6};
\node[font=\fontsize{4}{4.8}\selectfont] at (axis cs:7,3) {2.5};
\node[font=\fontsize{4}{4.8}\selectfont] at (axis cs:8,3) {-1.8};
\node[font=\fontsize{4}{4.8}\selectfont] at (axis cs:9,3) {-0.1};
\node[font=\fontsize{4}{4.8}\selectfont] at (axis cs:10,3) {-0.1};
\node[font=\fontsize{4}{4.8}\selectfont] at (axis cs:0,4) {5.0};
\node[font=\fontsize{4}{4.8}\selectfont] at (axis cs:1,4) {0.6};
\node[font=\fontsize{4}{4.8}\selectfont] at (axis cs:2,4) {5.4};
\node[font=\fontsize{4}{4.8}\selectfont] at (axis cs:3,4) {2.9};
\node[font=\fontsize{4}{4.8}\selectfont] at (axis cs:4,4) {1.9};
\node[font=\fontsize{4}{4.8}\selectfont] at (axis cs:5,4) {5.2};
\node[font=\fontsize{4}{4.8}\selectfont] at (axis cs:6,4) {1.1};
\node[font=\fontsize{4}{4.8}\selectfont] at (axis cs:7,4) {0.7};
\node[font=\fontsize{4}{4.8}\selectfont] at (axis cs:8,4) {3.1};
\node[font=\fontsize{4}{4.8}\selectfont] at (axis cs:9,4) {1.0};
\node[font=\fontsize{4}{4.8}\selectfont] at (axis cs:10,4) {1.9};
\node[font=\fontsize{4}{4.8}\selectfont] at (axis cs:0,5) {4.4};
\node[font=\fontsize{4}{4.8}\selectfont] at (axis cs:1,5) {3.1};
\node[font=\fontsize{4}{4.8}\selectfont] at (axis cs:2,5) {4.2};
\node[font=\fontsize{4}{4.8}\selectfont] at (axis cs:3,5) {5.9};
\node[font=\fontsize{4}{4.8}\selectfont] at (axis cs:4,5) {3.9};
\node[font=\fontsize{4}{4.8}\selectfont] at (axis cs:5,5) {4.4};
\node[font=\fontsize{4}{4.8}\selectfont] at (axis cs:6,5) {2.8};
\node[font=\fontsize{4}{4.8}\selectfont] at (axis cs:7,5) {0.2};
\node[font=\fontsize{4}{4.8}\selectfont] at (axis cs:8,5) {1.0};
\node[font=\fontsize{4}{4.8}\selectfont] at (axis cs:9,5) {2.4};
\node[font=\fontsize{4}{4.8}\selectfont] at (axis cs:10,5) {1.6};
\node[font=\fontsize{4}{4.8}\selectfont] at (axis cs:0,6) {3.2};
\node[font=\fontsize{4}{4.8}\selectfont] at (axis cs:1,6) {3.2};
\node[font=\fontsize{4}{4.8}\selectfont] at (axis cs:2,6) {2.8};
\node[font=\fontsize{4}{4.8}\selectfont] at (axis cs:3,6) {4.2};
\node[font=\fontsize{4}{4.8}\selectfont] at (axis cs:4,6) {8.9};
\node[font=\fontsize{4}{4.8}\selectfont] at (axis cs:5,6) {0.9};
\node[font=\fontsize{4}{4.8}\selectfont] at (axis cs:6,6) {1.2};
\node[font=\fontsize{4}{4.8}\selectfont] at (axis cs:7,6) {0.0};
\node[font=\fontsize{4}{4.8}\selectfont] at (axis cs:8,6) {-0.3};
\node[font=\fontsize{4}{4.8}\selectfont] at (axis cs:9,6) {1.8};
\node[font=\fontsize{4}{4.8}\selectfont] at (axis cs:10,6) {3.4};
\node[font=\fontsize{4}{4.8}\selectfont] at (axis cs:0,7) {1.8};
\node[font=\fontsize{4}{4.8}\selectfont] at (axis cs:1,7) {5.7};
\node[font=\fontsize{4}{4.8}\selectfont] at (axis cs:2,7) {6.1};
\node[font=\fontsize{4}{4.8}\selectfont] at (axis cs:3,7) {5.8};
\node[font=\fontsize{4}{4.8}\selectfont] at (axis cs:4,7) {6.9};
\node[font=\fontsize{4}{4.8}\selectfont] at (axis cs:5,7) {5.0};
\node[font=\fontsize{4}{4.8}\selectfont] at (axis cs:6,7) {5.2};
\node[font=\fontsize{4}{4.8}\selectfont] at (axis cs:7,7) {6.1};
\node[font=\fontsize{4}{4.8}\selectfont] at (axis cs:8,7) {2.5};
\node[font=\fontsize{4}{4.8}\selectfont] at (axis cs:9,7) {2.1};
\node[font=\fontsize{4}{4.8}\selectfont] at (axis cs:10,7) {4.1};
\node[font=\fontsize{4}{4.8}\selectfont] at (axis cs:0,8) {2.2};
\node[font=\fontsize{4}{4.8}\selectfont] at (axis cs:1,8) {1.8};
\node[font=\fontsize{4}{4.8}\selectfont] at (axis cs:2,8) {5.5};
\node[font=\fontsize{4}{4.8}\selectfont] at (axis cs:3,8) {4.6};
\node[font=\fontsize{4}{4.8}\selectfont] at (axis cs:4,8) {6.7};
\node[font=\fontsize{4}{4.8}\selectfont] at (axis cs:5,8) {7.7};
\node[font=\fontsize{4}{4.8}\selectfont] at (axis cs:6,8) {0.4};
\node[font=\fontsize{4}{4.8}\selectfont] at (axis cs:7,8) {6.7};
\node[font=\fontsize{4}{4.8}\selectfont] at (axis cs:8,8) {2.0};
\node[font=\fontsize{4}{4.8}\selectfont] at (axis cs:9,8) {5.7};
\node[font=\fontsize{4}{4.8}\selectfont] at (axis cs:10,8) {6.9};
\node[font=\fontsize{4}{4.8}\selectfont] at (axis cs:0,9) {4.1};
\node[font=\fontsize{4}{4.8}\selectfont] at (axis cs:1,9) {3.1};
\node[font=\fontsize{4}{4.8}\selectfont] at (axis cs:2,9) {5.3};
\node[font=\fontsize{4}{4.8}\selectfont] at (axis cs:3,9) {10.5};
\node[font=\fontsize{4}{4.8}\selectfont] at (axis cs:4,9) {10.0};
\node[font=\fontsize{4}{4.8}\selectfont] at (axis cs:5,9) {9.9};
\node[font=\fontsize{4}{4.8}\selectfont] at (axis cs:6,9) {5.6};
\node[font=\fontsize{4}{4.8}\selectfont] at (axis cs:7,9) {4.3};
\node[font=\fontsize{4}{4.8}\selectfont] at (axis cs:8,9) {5.4};
\node[font=\fontsize{4}{4.8}\selectfont] at (axis cs:9,9) {-1.0};
\node[font=\fontsize{4}{4.8}\selectfont] at (axis cs:10,9) {3.1};
\node[font=\fontsize{4}{4.8}\selectfont] at (axis cs:0,10) {5.8};
\node[font=\fontsize{4}{4.8}\selectfont] at (axis cs:1,10) {8.1};
\node[font=\fontsize{4}{4.8}\selectfont] at (axis cs:2,10) {7.7};
\node[font=\fontsize{4}{4.8}\selectfont] at (axis cs:3,10) {9.1};
\node[font=\fontsize{4}{4.8}\selectfont] at (axis cs:4,10) {10.5};
\node[font=\fontsize{4}{4.8}\selectfont] at (axis cs:5,10) {6.5};
\node[font=\fontsize{4}{4.8}\selectfont] at (axis cs:6,10) {3.2};
\node[font=\fontsize{4}{4.8}\selectfont] at (axis cs:7,10) {3.7};
\node[font=\fontsize{4}{4.8}\selectfont] at (axis cs:8,10) {4.7};
\node[font=\fontsize{4}{4.8}\selectfont] at (axis cs:9,10) {0.5};
\node[font=\fontsize{4}{4.8}\selectfont] at (axis cs:10,10) {3.0};
\end{axis}
\end{tikzpicture}
            \end{minipage}%
            \hfill
            \begin{minipage}[t]{0.48\textwidth}
                \input{plots/11_appendix/few_shot/delta/seed36_rotation_permutation}
            \end{minipage}
        \end{minipage}}
            \vspace{-5pt}
        \caption{\small
            Same as Figure~\ref{fig:icl_deltas_qwen32}, for \model{seed36}.
        }
        \vspace{-8pt}
        \label{fig:few_shot_delta_aliases_seed36}
    \end{figure}
    \FloatBarrier

\end{document}